\documentclass[10pt,journal,compsoc]{IEEEtran}
\usepackage{amsmath,epsfig}
\usepackage{graphicx}
\usepackage{amssymb}
\usepackage{amsfonts}
\usepackage{subfigure}
\usepackage{algorithmic}
\usepackage{algorithm}
\usepackage{xspace}
\usepackage{bbm}
\usepackage{breqn}
\usepackage{tabularx}
\usepackage{caption}
\usepackage{hhline}
\usepackage{enumitem}
\usepackage{multirow}
\usepackage{array,longtable,calc}
\usepackage{xcolor,colortbl}
\usepackage{hyperref}
\usepackage{url}
\usepackage[nocompress]{cite}

\newcommand{\ignore}[1]{}

\newcommand{\subparagraph}{}
\usepackage{titlesec}
\renewcommand{\arraystretch}{1.2}
\titleformat{\paragraph}[runin]{\bfseries}{}{0pt}{}
\titlespacing*{\paragraph}{0pt}{5pt}{4pt}

\newcommand{\Y}{{\cal Y}}

\newcommand{\bx}{\mathbf{x}}
\newcommand{\by}{\mathbf{y}}

\newcommand{\bw}{\mathbf{w}}

\newcommand{\be}{\mathbf{e}}

\newcommand{\argmin}{\operatornamewithlimits{argmin}}

\def\be {\begin{equation}}
\def\ee {\end{equation}}
\def\beas {\begin{eqnarray*}}
\def\eeas {\end{eqnarray*}}
\def\bea {\begin{eqnarray}}
\def\eea {\end{eqnarray}}

\newcommand{\parens}[1]{\left(#1\right)}
\newcommand{\brackets}[1]{\left[#1\right]}

\begin{document}
	
\title{3D Object Proposals using Stereo Imagery for Accurate Object Class Detection}

\author{Xiaozhi~Chen$^*$, Kaustav~Kundu$^*$, Yukun~Zhu, 
	Huimin~Ma, Sanja~Fidler and Raquel~Urtasun 
	\IEEEcompsocitemizethanks{\IEEEcompsocthanksitem $^\ast$ Denotes equal contribution. %
        \IEEEcompsocthanksitem X. Chen and H. Ma are with the Department of Electronic Engineering, Tsinghua University, China. 
		\IEEEcompsocthanksitem K. Kundu, Y. Zhu, 
		S. Fidler and R. Urtasun are with the Department of Computer Science, University of Toronto, Canada.}%
	}


\IEEEtitleabstractindextext{%
	\begin{abstract}
The goal of this paper is to perform 3D object detection in the context of autonomous driving. Our method aims at generating a set of high-quality 3D object proposals by exploiting stereo imagery. We formulate the problem as minimizing an energy function that encodes object size priors, placement of objects on the  ground plane as well as several depth informed features that reason about free space, point cloud densities and distance to the ground. We then exploit a CNN on top of these proposals to perform object detection. In particular, we employ a convolutional neural net (CNN) that exploits context and depth information to jointly regress to 3D bounding box coordinates and object pose.
Our experiments show significant performance gains over existing RGB and RGB-D object proposal methods  on the challenging KITTI benchmark. When combined with the CNN, our approach outperforms all existing results in object detection and orientation estimation tasks for all three KITTI object classes. Furthermore, we experiment also with the setting where LIDAR information is available, and show that using both LIDAR and stereo leads to the best result.

\end{abstract}

}

\maketitle

\IEEEraisesectionheading{\section{Introduction}\label{sec:introduction}}

\IEEEPARstart{A}{utonomous} driving is receiving a lot of attention from both industry and the research community.
Most self-driving cars build their perception systems on expensive sensors, such as LIDAR, radar and high-precision GPS.
Cameras are an appealing alternative as they provide richer sensing at a much lower cost. 
This paper aims at high-performance 2D and 3D object detection in the context of autonomous driving by exploiting stereo imagery.

With impressive advances in deep learning in the past few years, recent efforts in object detection exploit object proposals to facilitate classifiers with powerful, hierarchical visual representation~\cite{alexnet, verydeep}.
Compared with traditional sliding window based methods~\cite{dpm}, the pipeline of generating object proposals that are combined with convolutional neural networks has lead to more than 20\% absolute performance gains~\cite{girshick2013rich, ZhuSegDeepM15} on the PASCAL VOC dataset~\cite{pascal-voc-2010}.

Object proposal methods aim at generating a moderate number of candidate regions that cover most of the ground truth objects in the image.
One typical approach is to perform region grouping based on superpixels using a variety of similarity measures~\cite{van2011segmentation, ArbelaezCVPR14}. 
Low-level cues such as color contrast, saliency~\cite{AlexePAMI12}, gradient~\cite{BingObj2014} and contour information~\cite{zitnick2014edge} have also been exploited in order to select promising object boxes from densely sampled windows.
There has also been some recent work on learning to generate a diverse set of region candidates with ensembles of binary segmentation models~\cite{krahnbuhl15}, parametric energies~\cite{TLeeICCV15} or CNN-based cascaded classifiers~\cite{deepproposal}.

The object proposal methods have proven effective on the PASCAL VOC benchmark. However, they have very low achievable recall on the autonomous driving benchmark KITTI~\cite{kitti}, which presents the bottleneck for the state-of-the-art object detector R-CNN~\cite{girshick2013rich, girshick15fastrcnn} on this benchmark.
On one hand, the PASCAL VOC dataset uses a loose overlap criteria for localization measure, i.e., a predicted box is considered to be correct if its overlap with the ground-truth box exceeds 50\%.
For self-driving cars, however, object detection requires a stricter overlap criteria to enable correct estimates of the distance of vehicles from the ego-car. 
Moreover, objects in KITTI images are typically small and many of them are heavily occluded or truncated. 
These challenging conditions limit the performance of most existing bottom-up proposals that rely on intensity and texture for superpixel merging and window scoring.

\begin{figure*}[t!]
\begin{center}
\addtolength{\tabcolsep}{-4.5pt}
\begin{tabular}{cccc}
 Image & \small Depth from Stereo & \small depth-Feat & \small Prior\\[-0.5mm]
\includegraphics[height=0.098\linewidth,trim = 5mm 0mm 0mm 0mm, clip]{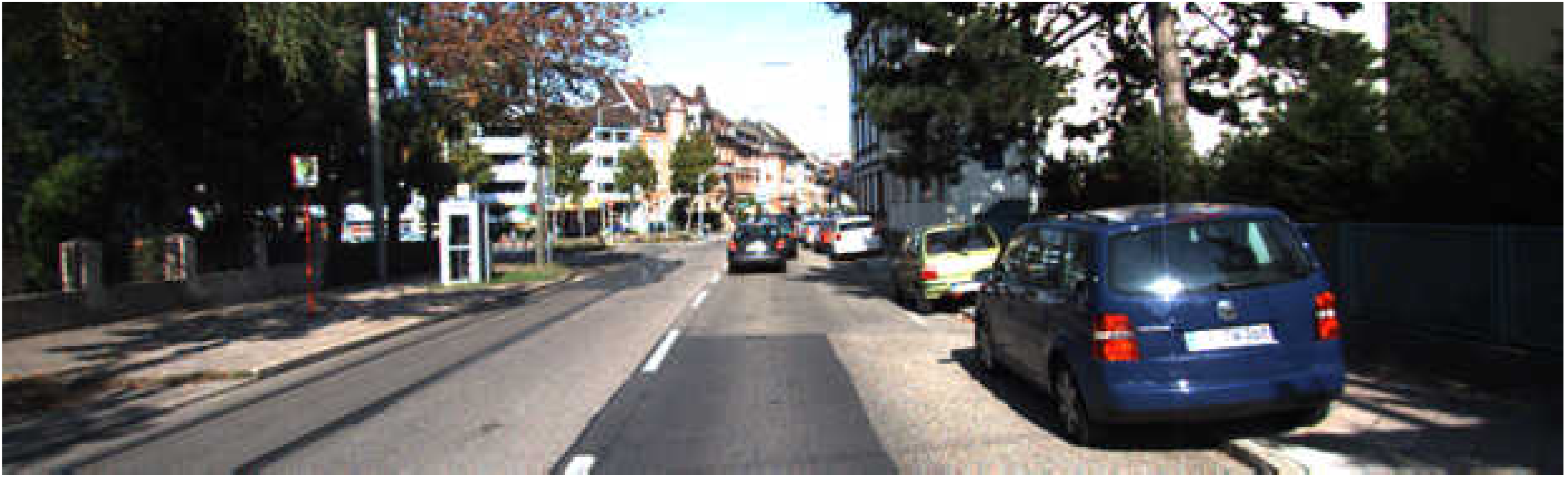} & 
\includegraphics[height=0.098\linewidth,trim = 0mm 12mm 33mm 0mm, clip]{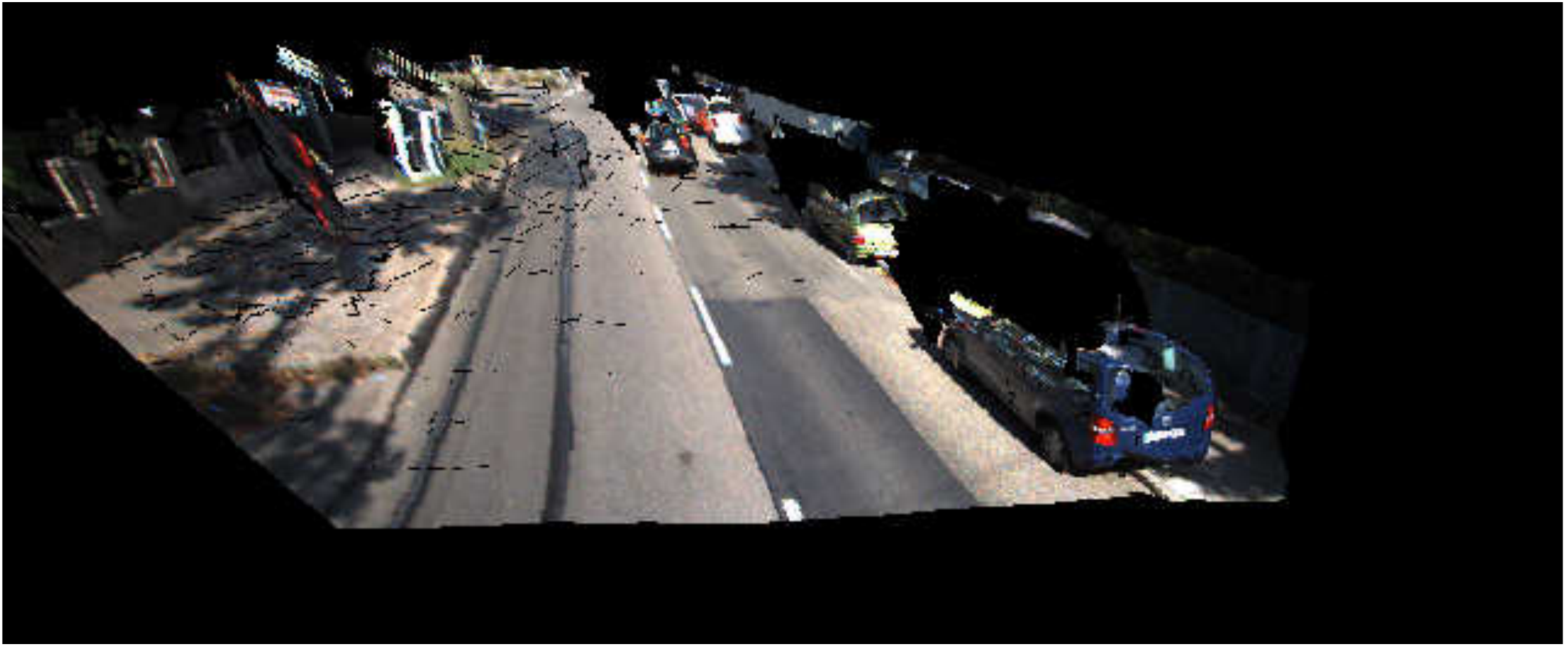} &
\includegraphics[height=0.098\linewidth,trim = 15mm 0mm 0mm 0mm, clip]{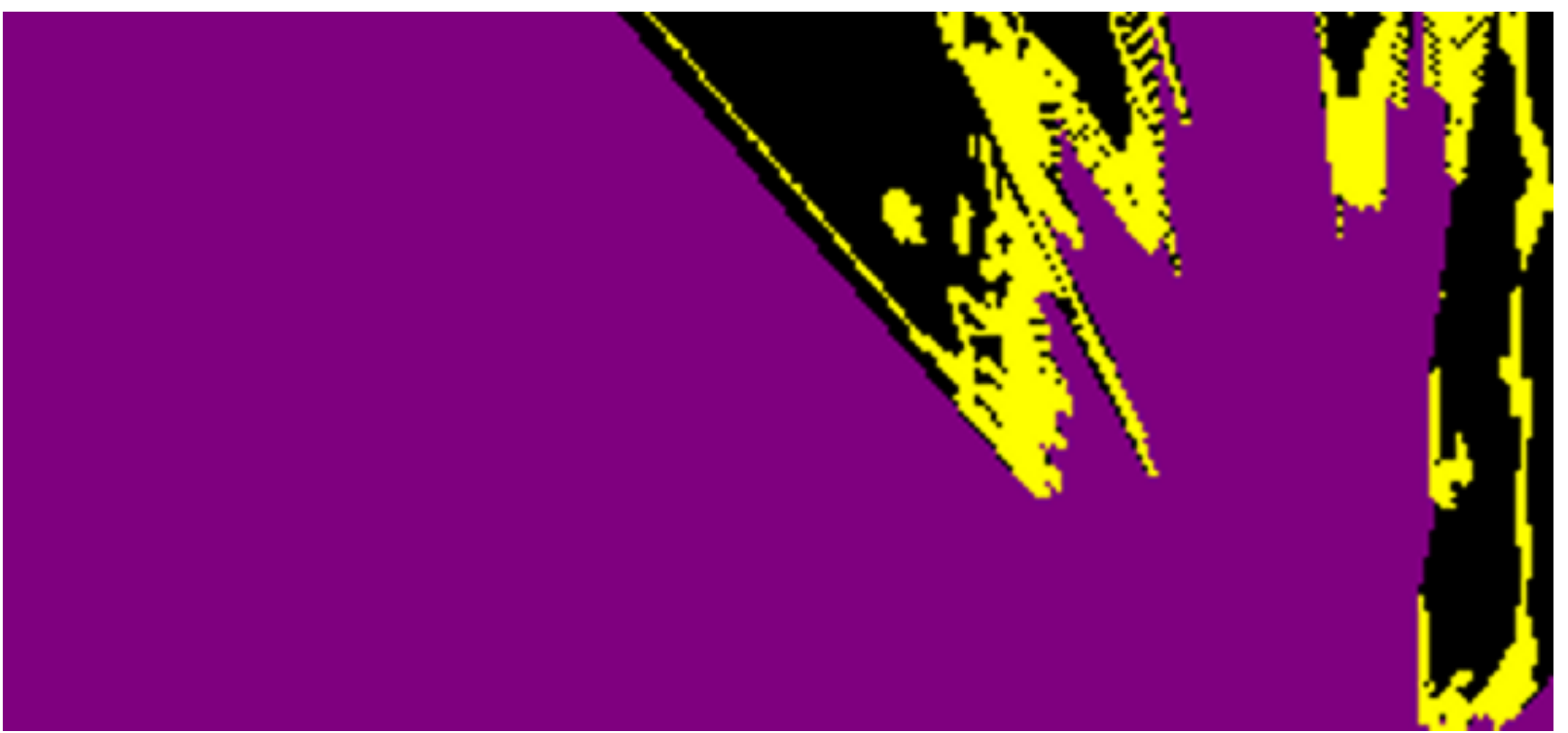} &
\includegraphics[height=0.098\linewidth,trim = 0mm 12mm 33mm 0mm, clip]{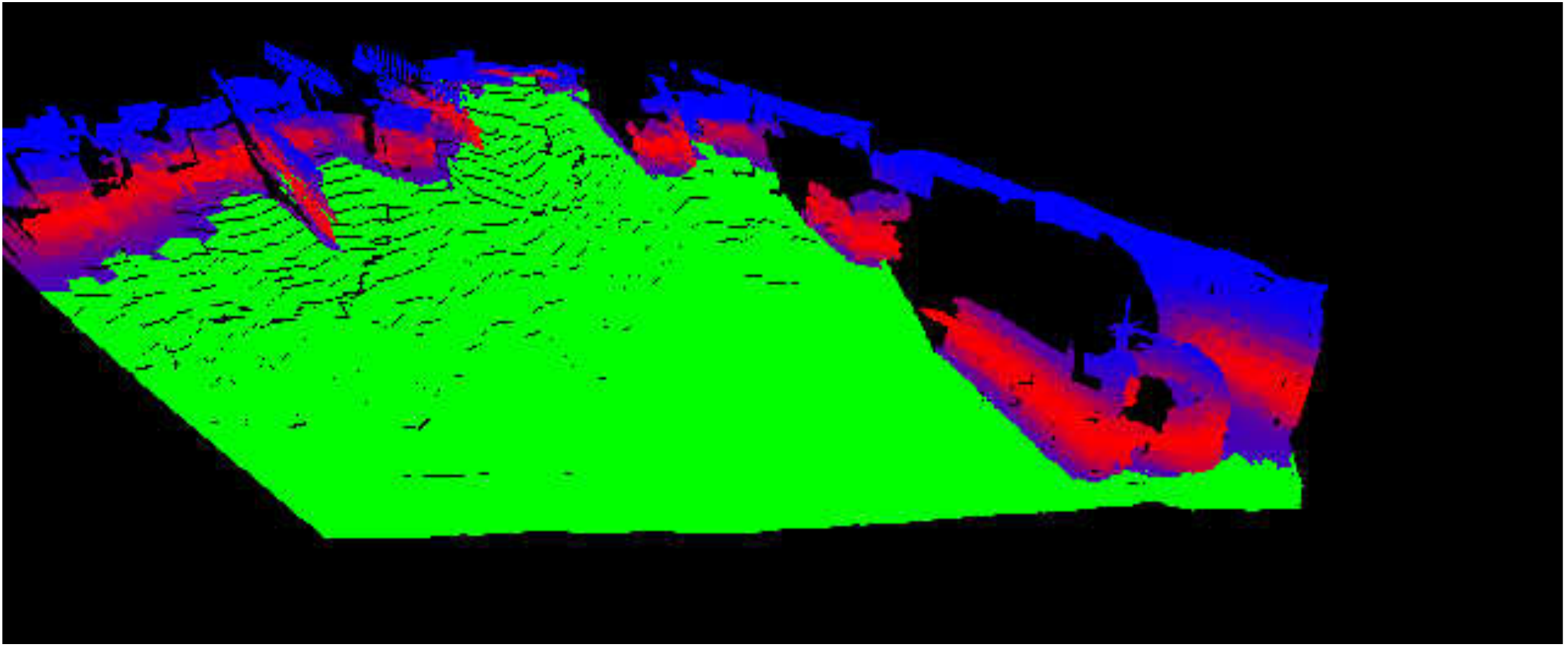}
\end{tabular}
\vspace{-5mm}
\end{center}
\caption{\small {\bf Features in our model} (from left to right): left camera image, stereo 3D reconstruction, depth-based features and our prior. In the third image, occupancy is marked with yellow ($P$ in Eq.~\eqref{eq:s}) and purple denotes free space ($F$ in Eq.~\eqref{eq:f}). In the prior, the ground plane is green and blue to red indicates increasing prior value of object height.}
\label{fig:feature_viz}
\vspace{-3mm}
\end{figure*}

In this paper, we propose a novel 3D object detection approach that exploits stereo imagery and contextual information specific to the domain of autonomous driving.
We propose a 3D object proposal method that goes beyond 2D bounding boxes and is capable of generating high-quality 3D bounding box proposals.
We make use of the 3D information estimated from a stereo camera pair by placing 3D candidate boxes on the ground plane and scoring them via 3D point cloud features. 
In particular, our scoring function encodes several depth informed features such as point densities inside a candidate box, free space, visibility, as well as object size priors and height above the ground plane.
The inference process is very efficient as all the features can be computed in constant time via 3D integral images.
Learning can be done using structured SVM~\cite{ssvm} to obtain class-specific weights for these features.
We also present a 3D object detection neural network that takes 3D object proposals as input and predict accurate 3D bounding boxes. The neural net exploits contextual information and uses a multi-task loss to jointly regress to bounding box coordinates and object orientation.

We evaluate our approach on the challenging KITTI detection benchmark~\cite{kitti}. Extensive experiments show that:
1) The proposed 3D object proposals achieve significantly higher recall than the state-of-the-art across all overlap thresholds under various occlusion and truncation levels. In particular, compared with the state-of-the-art RGB-D method MCG-D~\cite{guptaECCV14}, we obtain $25\%$ higher recall with 2K proposals.
2) Our 3D object detection network combined with 3D object proposals outperforms all published results on object detection and orientation estimation for \emph{Car}, \emph{Cyclist} and \emph{Pedestrian}.
3) Our approach is capable of producing accurate 3D bounding box detections, which allows us to locate objects in 3D and infer the distance and pose of objects from the ego-car.
4) We also apply our approach to LIDAR point clouds with more precise, but sparser, depth estimation. When combining stereo and LIDAR data, we obtain the highest 3D object detection accuracy.

A preliminary version of this work was presented in~\cite{XiaozhiNIPS15}. In this manuscript, we make extensions in the following aspects:
1) A more detailed description of the inference process of proposal generation. 
2) The 3D object proposal model is extended with a class-independent variant.
3) The detection neural network is extended to a two-stream network to leverage both appearance and depth features.
4) We further apply our model to point clouds obtained via LIDAR, and provide  comparison of the stereo, LIDAR and the hybrid settings.
5) We extensively evaluate the 3D bounding box recall and 3D object detection performance.
6) Our manuscript includes ablation studies of network design, depth features, as well as ground plane estimation.

\vspace{-2mm}
\section{Related Work}

Our work is closely related to object proposal generation and 3D object detection. We briefly review the literature with a focus on the domain of autonomous driving.

\vspace{-1mm}
\paragraph{Object Proposal Generation.}

Object proposal generation has become an important technique in object detection.
It speeds up region searching and enables object detectors to leverage the great power of deep neural networks~\cite{alexnet,verydeep}.
Numerous works have been proposed for different modalities, i.e., RGB~\cite{van2011segmentation,zitnick2014edge,CarreiraCpmcPAMI2012,ArbelaezCVPR14,BingObj2014,TLeeICCV15}, RGB-D~\cite{Banica13,guptaECCV14,Lin13,KarpathyICRA13}, and video~\cite{OneataECCV14,moving2015CVPR}.

In RGB, one typical paradigm is to generate candidate segments by grouping superpixels or multiple figure-ground segmentations with diverse seeds. 
Grouping-based methods~\cite{van2011segmentation,ArbelaezCVPR14, randPrim13} build on multiple oversegmentations and merge superpixels based on complementary cues such as color, texture and shape.
Geodesic proposals~\cite{krahenbuhl14} learn to place diverse seeds and identify promising regions by computing geodesic distance transforms. 
CPMC~\cite{CarreiraCpmcPAMI2012} solves a sequence of binary parametric min-cut problems with different seeds and unary terms. The resulting regions are then ranked using Gestalt-like features and diversified using maximum marginal relevance measures. This approach is widely used in recognition tasks~\cite{s2p,segdpm,ZhuSegDeepM15}.
Some recent approaches also follow this pipeline by learning an ensemble of local and global CRFs~\cite{krahnbuhl15} or minimizing parametric energies that encode mid-level cues such as symmetry and closure~\cite{TLeeICCV15}.
Another paradigm generates bounding box proposals by scoring exhaustively sampled windows.
In~\cite{AlexePAMI12}, a large pool of windows are scored with a diverse set of features such as color contrast, edges, location and size.
BING~\cite{BingObj2014} scores windows using simple gradient features which serve as an object closure measure and can be computed extremely fast.
BING++~\cite{bing++} further improves its localization quality using edge and superpixel based box refinement~\cite{ChenCVPR15}.
EdgeBoxes~\cite{zitnick2014edge} design an effective scoring function by computing the number of contours that exist in or straddle the bounding box.
\cite{deepproposal} computes integral image features from inverse cascading layers of CNN for candidate box scoring and refinement.
A detailed comparison of existing proposal methods has been carried out in~\cite{Hosang2015PAMI}.
While most of these approaches achieve more than 90\% recall with 2K proposals on the PASCAL VOC benchmark~\cite{pascal-voc-2010}, they have significant lower recall on the KITTI dataset.

In RGB-D, \cite{Banica13,Lin13} extend CPMC~\cite{CarreiraCpmcPAMI2012} with depth cues and fit 3D cubes around candidate regions to generate cuboid object proposals.
\cite{guptaECCV14} extends MCG~\cite{ArbelaezCVPR14} with RGB-D contours as well as depth features to generate 2.5D proposals. They obtain significantly better performance compared with purely RGB approaches.
In~\cite{KarpathyICRA13}, candidate objects are proposed from 3D meshes by oversegmentation and several intrinsic shape measures. 
Our work is also relevant to Sliding Shapes~\cite{SongECCV14}, which densely evaluates 3D windows with exemplar-SVM classifiers in 3D point clouds. However, they train exemplar classifiers on CAD models with hundreds of rendered views and complex shape features, resulting in very inefficient training and inference.
In our work, we advance over past work by exploiting the physical sizes of objects, the ground plane, as well as depth features and contextual information in 3D.


\begin{figure*}[t!]
	\begin{center}
		\includegraphics[width=0.88\linewidth,trim = 0mm 12mm 0mm 12mm, clip]{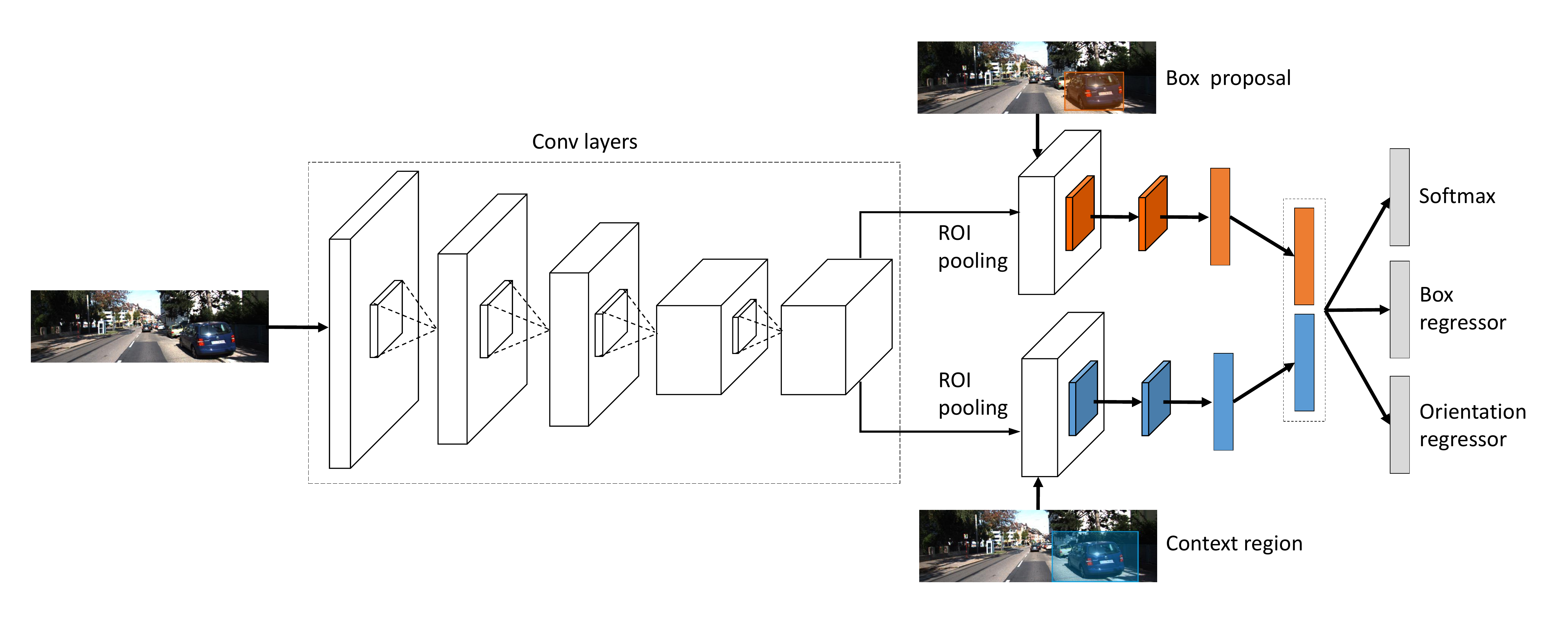}
		\vspace{-1mm}
		\caption{{\bf The single-stream network for 3D object detection:} Input can be an RGB image or a 6-channel RGB-HHA image.}
		\label{fig:net}
		\vspace{-3mm}
	\end{center}
\end{figure*}

\vspace{-1mm}
\paragraph{3D Object Detection.}
In the domain of autonomous driving, accurate 3D localization and pose estimation of objects beyond 2D boxes are desired.
In~\cite{PepikPAMI15}, the Deformable Part-based Model~\cite{dpm} is extended to 3D by adding viewpoint information and 3D part geometry. The potentials are parameterized in 3D object coordinates instead of the image plane.
Zia et al.~\cite{ZiaIJCV15} initialize a set of candidate objects using a variant of poselets detectors and model part-level occlusion and configuration with 3D deformable wireframes.
\cite{BarTITS15} trains an ensemble of subcategory models by clustering object instances with appearance and geometry features.
In~\cite{LongACCV14}, a top-down bounding box re-localization scheme is proposed to refine Selective Search proposals with Regionlets features.
\cite{WangCVPR15} combines cartographic map priors and DPM detectors into a holistic model to re-reason object locations.
\cite{CarAOG_ECCV2014} uses And-Or models to learn car-to-car context and occlusion patterns.
\cite{spLBP} learns AdaBoost classifier with dense local features within subcategories.
The recently proposed 3DVP~\cite{xiangcvpr15} employs ACF detectors~\cite{DollarPAMI14} and learns occlusion patterns with 3D voxels.

With the shift of low-level features to multi-layer visual representation, most of recent approaches exploit CNNs for object detection also in the context of autonomous driving.
In~\cite{HosangCVPR15}, R-CNN is applied on pedestrian detection with proposals generated by SquaresChnFtrs detector, achieving moderate performance.
\cite{DeepParts2015} learns part detectors with convolutional features to handle occlusion in pedestrian detection.
\cite{CompACT2015} designs a complexity-aware cascade pedestrian detector with convolutional features.
Parallel to our work, Faster R-CNN~\cite{renNIPS15fasterrcnn} improves upon their prior R-CNN~\cite{girshick2013rich} pipeline by integrating proposal generation and R-CNN into an end-to-end trainable network.
However, these methods only produce 2D detections, whereas our work aims at 3D object detection in order to infer both, accurate object pose as well as the distance from the ego-car.

\vspace{-1mm}
\section{3D Object Proposals}
\label{sec:approach}

Our approach aims at generating a diverse set of 3D object proposals in the context of autonomous driving. 3D reasoning is crucial in this domain as it eases problems such as occlusion and large scale variation. 
The input to our method is a stereo image pair. We compute depth using the method by Yamaguchi et al.~\cite{Koichiro14}, yielding a point cloud $\bx$. 
We place object proposals in 3D space in the form of 3D bounding boxes. 
Note that only depth information (no appearance) is used in our proposal generation process. Next we describe our parameterization and the framework.

\vspace{-1mm}
\subsection{Proposal Generation as Energy Minimization}
\label{sec:grid}

We use a 3D bounding box to represent each object proposal $\by$,
which is parametrized by a tuple, $\parens{x, y, z, \theta, c, t}$, where $\parens{x, y, z}$ is the 3D box center and $\theta$ denotes the azimuth angle. 
Here, $c \in C$ is the object class  and $t \in \{1, \ldots, T_{c}\}$ indexes a set of 3D box templates, which are learnt from training data to represent the typical physical size of each class $c$ (details in Sec.~\ref{sec:templates}). 
We discretize the 3D space into voxels for candidate box sampling and thus each box $\by$ is represented in discretized form (details in Sec.~\ref{sec:inference}).

We generate proposals by minimizing an energy function which encodes several depth-informed potentials. We encode the fact that the object should live in a space occupied with high density by the point cloud.
Furthermore, the box $ \by $ should have minimal overlap with the free space in the scene.
We also encode the height prior of objects, and the fact that the point cloud in the box's immediate vicinity should have lower prior values of object height than the box.
The energy function is formulated as:
\begin{align}
E(\bx, \by) =&\ \bw^{\top}_{c, pcd} \phi_{pcd}(\bx, \by) +  \bw^{\top}_{c, fs} \phi_{fs}(\bx, \by) \notag\\ 
& + \bw^{\top}_{c, ht} \phi_{ht}(\bx, \by) +   \bw^{\top}_{c, ht-contr} \phi_{ht-contr}(\bx, \by)\notag. 
\label{eq:total}
\end{align}
The weights of the energy terms are learnt via structured SVM~\cite{TsochantaridisICML2004} (details in Sec.~\ref{sec:learning}).
Note that the above formulation encodes dependency of weights on the object class, thus weights are learnt specific to each class. However, we can also learn a single set of weights for all classes (details in Sec.~\ref{sec:class-ind}).
We next explain each potential in more detail.

\vspace{-1mm}
\paragraph{Point Cloud Density:} This potential encodes the point cloud density within the box:
\begin{equation}
\label{eq:s}
\phi_{pcd}(\bx, \by) = \frac{\sum_{v \in \Omega(\by)} P(v)}{|\Omega(\by)|}
\end{equation}
where $P(v)\in\{0,1\}$ indicates whether voxel $v$ contains point cloud points or not, and $\Omega(\by)$ denotes the set of voxels within box $\by$. The feature $P$ is visualized in Fig.~\ref{fig:feature_viz}. This potential is simply computed as the fraction of occupied voxels within the box. By using integral accumulators (integral images in 3D), the potential can be computed efficiently in constant time.

\begin{figure}[t!]
	\begin{center}
		\includegraphics[width=0.95\linewidth,trim=10mm 10mm 15mm 15mm, clip]{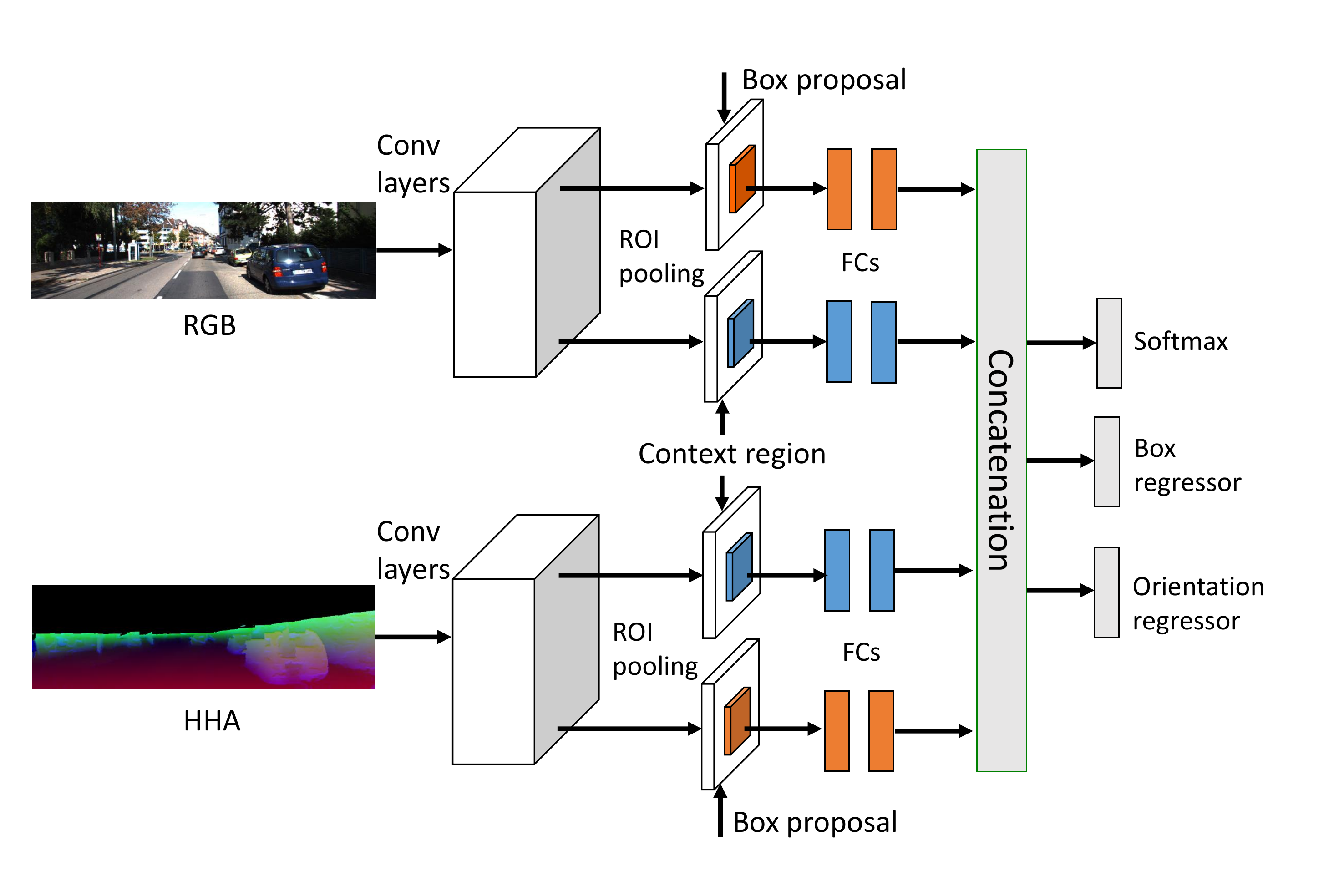}
		\caption{\textbf{Two-stream network for 3D object detection:} The convnet learns from RGB (top) and HHA~\cite{guptaECCV14} (bottom) images as input, and concatenates features from {\it fc7} layers for multi-task prediction. The model is trained end-to-end.}
		\label{fig:rgbd-nets}
        \vspace{-3mm}
	\end{center}
\end{figure}

\vspace{-1mm}
\paragraph{Free Space:} Free space is defined as the space that lies on the rays between the point cloud and the camera.
This potential encodes the fact that the box should not contain a significant amount of free space (since it is occupied by the object).
We define $F$ as a binary valued grid, where $F(v) = 1$ means that the ray from the camera to voxel $ v $ is not intercepted by any occupied voxel, i.e., voxel $v$ belongs to the free space. The potential is defined as follows:
\begin{equation}
\label{eq:f}
\phi_{fs}(\bx, \by) = \frac{\sum_{v \in \Omega(\by)}  (1-F(v))}{|\Omega(\by)|}
\end{equation}
It encourages less free space within the box, and can be efficiently computed using integral accumulators.

\vspace{-1mm}
\paragraph{Height Prior:} This potential encourages the height of the point cloud within the box w.r.t. the road plane to be close to the mean height $\mu_{c, ht}$ of the object class $c$.
We encode it as follows:
\begin{equation}
\phi_{ht}(\bx, \by) = \dfrac{1}{|\Omega(\by)|}\sum_{v \in \Omega(\by)} H_{c}(v) 
\end{equation}
with
\begin{align}
H_{c}(v) = \left\lbrace \begin{array}{ll}
\text{exp} \brackets{-\dfrac{1}{2}\parens{\dfrac{d_v - \mu_{c, ht}}{\sigma_{c, ht}}}^2}, & \text{if} \,\,\, P(v) = 1 \\
0, & \text{o.w.}
\end{array} \right.
\label{eq:ht_prior}
\end{align}
Here, $ d_v $ is the distance between the center of the voxel $ v $ and the road plane, along the direction of the gravity vector.
By assuming a Gaussian distribution of the data, we compute $\mu_{c, ht}, \sigma_{c, ht}$ as the MLE estimates of  mean height and standard deviation. The feature is shown in Fig.~\ref{fig:feature_viz}. It can be efficiently computed via integral accumulators.

\vspace{-1mm}
\paragraph{Height Contrast:} This potential encodes the fact that the point cloud surrounding the box should have lower values of the height prior relative to the box.
    We first compute a surrounding region $\by^+$ of box $\by$ by extending $\by$ by 0.6m\footnote{The value as well as other hyper parameters (e.g., voxel size) are determined based on the performance on the validation set.} in the direction of each face. We formulate the contrast of height priors between box $\by$ and surrounding box $\by^+$ as:
\begin{equation}
\phi_{ht-contr}(\bx, \by) = \dfrac{\phi_{ht}(\bx, \by)}{\phi_{ht}(\bx, \by^+) - \phi_{ht}(\bx, \by)}
\end{equation}

\vspace{-1mm}
\subsection{Inference}
\label{sec:inference}


We compute the point cloud $\bx$ from a stereo image pair using the approach by Yamaguchi et al.~\cite{Koichiro14}. Then we discretize the 3D space and estimate the road plane for 3D candidate box sampling. We perform exhaustive scoring of each candidate using our energy function, and use non-maximal suppression (NMS) to obtain top $K$ diverse 3D proposals. 
In particular, we use a greedy algorithm, where at each iteration we select the next proposal that has the lowest energy and its IoU overlap with the previously selected proposals does not exceed a threshold $\delta$.
Specifically, the $m^{th}$ proposal $\by^m$ is obtained by solving the following problem:
\begin{equation}
\begin{aligned}
    \by^m = & \argmin_{\by \in \Y} E(\bx,\by) \\
    \text{s.t.}\quad & \text{IoU}(\by, \by^i) < \delta, \quad \forall i \in \{0, \dots, m-1\},
\end{aligned}
\end{equation}

\vspace{-1mm}
\paragraph{Discretization and Accumulators:}
The point cloud is defined in a left-handed coordinate system, where the Y-axis goes in the direction of gravity and the positive Z-axis is along the camera's  viewing direction. We discretize the 3D continuous space such that the each voxel has length of $0.2$m in each dimension. We compute the point cloud occupancy, free space and height prior grids in this voxel space, as well as their 3D integral accumulators. 

\vspace{-1mm}
\paragraph{Ground Plane Estimation:}
We estimate the ground plane by classifying superpixels~\cite{Koichiro14} using a very small neural network, and fitting a plane to the estimated ground pixels using RANSAC. We use the following features on the superpixels as input to the network: mean RGB values, average 2D and 3D position, pitch and roll angles relative to the camera of the plane fit to the superpixel, a flag as to whether the average 2D position was above the horizon line, and standard deviation of both the color values and 3D position. This results in a 22-dimensional feature vector. The neural network consists of only a single hidden layer which also has 22 units. We use \emph{tanh} as the activation function and cross-entropy as the loss function. We train the network on the KITTI's road benchmark~\cite{kitti}.

\vspace{-1mm}
\paragraph{Bounding Boxes Sampling and Scoring:}
For 3D candidate box sampling, we use three size templates per class and two orientations $\theta \in \{0,90\}$. As all the features can be efficiently computed via integral accumulators, it takes constant time to evaluate each configuration $\by$. Despite that, evaluating exhaustively in the entire space would be slow. We reduce the search space by skipping empty boxes that do not contain any points.
With ground plane estimation, we further reduce the search space along the vertical dimension by only placing candidate boxes on the ground plane. 

However, to alleviate the noise of stereo depth at large distances, we sample additional candidate boxes at distances larger than $20$m from the camera. In particular, let $y_{road}$ denote the height of the ground plane. We deviate this height along the vertical dimension to compute two additional planes that have heights $y = y_{road} \pm \sigma_{road}$. Here $\sigma_{road}$ denotes the MLE estimate of the standard deviation of a Gaussian distribution modeling the distance of objects from the ground plane. We then sample additional boxes on these planes. With our sampling strategy, scoring all configurations can be done in a fraction of a second.

Note that the energy function is computed independently with respect to each candidate box. We rank all boxes according to the values of $E(\bx,\by)$, and perform greedy inference with non-maxima suppression (NMS). In practice, we perform NMS in 2D as it achieves similar recall as NMS in 3D while being much faster.
The IoU threshold $\delta$ is set to 0.75.
The entire feature computation and inference process takes 1.2s per image on average for 2K proposals.

\vspace{-1mm}
\subsection{Learning}
\label{sec:learning}

We next explain how we obtain the 3D bounding box templates, and how we learn the weights in our model.

\vspace{-1mm}
\subsubsection{3D Bounding Box Templates}
\label{sec:templates}

The size templates are obtained by clustering the ground truth 3D bounding boxes on the training set. In particular, we first compute a histogram for the object sizes, and choose a cluster of boxes that have IoU overlaps with the mode of the histogram above 0.6, then remove those boxes and iterate. The representative size templates are computed by averaging the box sizes in each cluster.

\vspace{-1mm}
\subsubsection{Learning the Weights in Our Model}

We use structured SVM~\cite{TsochantaridisICML2004} to learn the model's weights $\{w_{c, pcd}, w_{c, fs}, w_{c, ht}, w_{c, ht-contr}\}$. Given $N$ input-output training pairs, $\{\bx^{(i)},\by^{(i)}\}_{i=1, \cdots, N}$, we obtain the parameters by solving the following optimization problem:
\begin{align}
&\min_{\mathbf{w}\in\mathbb{R}^D} 
\frac{1}{2} ||\mathbf{w}||^2 
+ \frac{C}{N} \sum_{i=1}^N \xi_i \nonumber \\
\text{s.t.:}\quad   & \bw^T (\phi(\bx^{(i)},\by) - \phi(\bx^{(i)},\by^{(i)})) \\ 
& \geq \Delta (\by^{(i)},\by)-\xi_i,  \forall \by  \setminus \by^{(i)}\nonumber 
\end{align}
We use the parallel cutting plane implementation of~\cite{box} to solve this minimization problem. 
As the task loss $\Delta (\by^{(i)},\by)$, we use the  strict 3D Intersection-over-Union (IoU) to encourage accurate placement of the 3D proposals.
In particular, 3D IoU is computed as the volume of intersection of two 3D bounding boxes divided by the volume of their union.

\vspace{-1mm}
\subsubsection{Class-Independent 3D Proposals}
\label{sec:class-ind}
The method described above learns separate weights for each category, thus generating class-dependent object proposals. However, the approach can be easily modified to generate class-independent proposals by learning only a single scoring model for all categories. In particular, we learn object templates for all classes jointly rather than for each specific class. Therefore,  the weights in this energy are class-independent (we have only a single set of weights). We compare these two approaches in the experiments.

\vspace{-1mm}
\section{3D Object Detection Networks}
In this section, we describe how we score the top-ranked 3D object proposals via convolutional networks. We design a network architecture for two tasks: joint 2D object detection and orientation estimation, and 3D object detection.

\vspace{-1mm}
\subsection{Joint 2D Object Detection and Pose Estimation}

The architecture of our network for joint 2D object detection and orientation estimation is shown in Fig.~\ref{fig:net}. The network is built upon Fast R-CNN~\cite{girshick15fastrcnn}, which share the convolutional features across all proposals and use a ROI pooling layer to compute proposal-specific features. We extend this basic network by adding a context branch after the last convolutional layer (i.e., {\it conv5}), and an orientation regression loss to jointly learn object location and orientation.
Specifically, the first branch encodes features from the original candidate regions while the second branch is specific to context regions, which are computed by enlarging the candidate boxes by a factor of 1.5, following the segDeepM approach~\cite{ZhuSegDeepM15}.
Both branches consist of a ROI pooling layer and two fully connected layers. ROIs are obtained by projecting the 3D proposals onto the image plane and then onto the {\it conv5} feature maps. We concatenate the features from $\textit{fc}_7$ layers and feed them to the prediction layers.

We predict the class labels, bounding box coordinate offsets, and object orientation jointly using a multi-task loss. We define the category loss as cross entropy, the orientation loss and bounding box offset loss as a smooth $\ell_1$ loss. We parameterize the bounding box coordinates as in~\cite{girshick2013rich}. Each loss is weighted equally and only the category label loss is employed for the background boxes.

\vspace{-1mm}
\subsection{3D Object Detection}
For 3D object detection, we want to output full 3D bounding boxes for objects.
We use the same network as in Fig.~\ref{fig:net}, except that 2D bounding box regressors are replaced by 3D bounding box regressors. Similarly to 2D box regression, we parametrize the centers of 3D boxes with size normalization for scale-invariant translation, and the 3D box sizes with log-space shift.
In particular, we denote a 3D box proposal as $P=(P_x,P_y,P_z,P_x^s,P_y^s,P_z^s)$, and its corresponding ground truth 3D box as $G=(G_x,G_y,G_z,G_x^s,G_y^s,G_z^s)$, which specify the box center and the box size in each dimension. The regression targets for the box center $T_c(P)$ and the box size $T_c^s(P)$ are parametrized as follows:
\begin{align}
T_c(P)=\frac{G_c - P_c}{P_c^s}, \, \, \, \,
T_c^s(P)=log\frac{G_c^s}{P_c^s}, \, \, \, \,
\forall c \in \{x,y,z\}
\end{align}
Given the 3D box coordinates and the estimated orientation, we then compute the azimuth angle $\theta$ of the box.

\begin{figure*}[t!]
\begin{center}
\begin{tabular}{p{3mm}ccc}
\hspace{-5mm}\rotatebox{90}{\hspace{3mm}Car} &\hspace{-1cm}
\raisebox{-0.38\height}{\includegraphics[width=0.265\linewidth,trim = 5mm 0mm 0mm 0mm, clip]{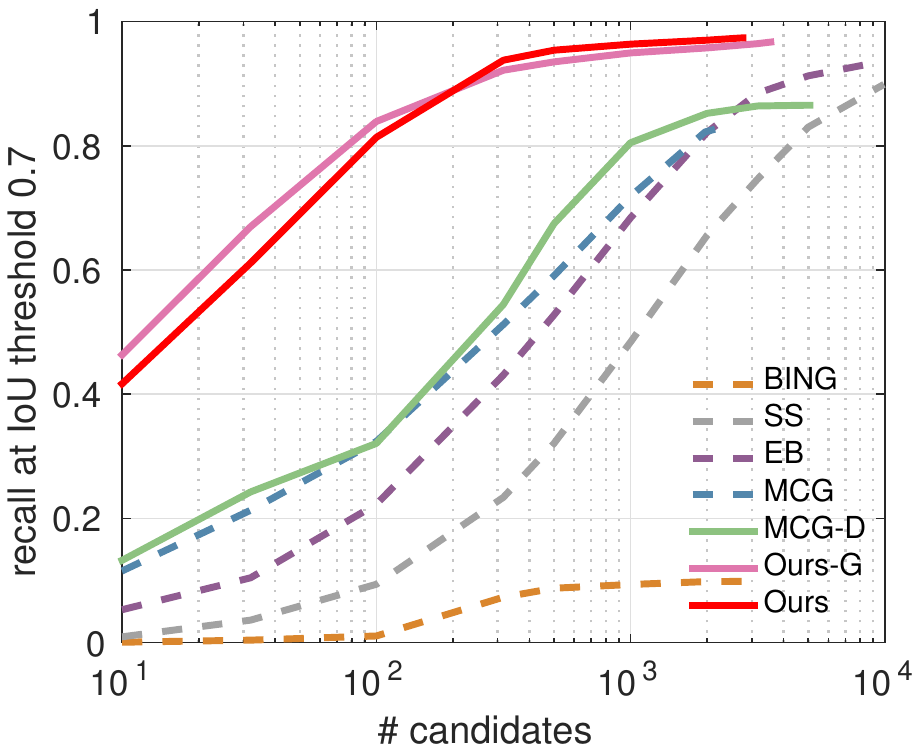}}&
\raisebox{-0.38\height}{\includegraphics[width=0.265\linewidth,trim = 5mm 0mm 0mm 0mm, clip]{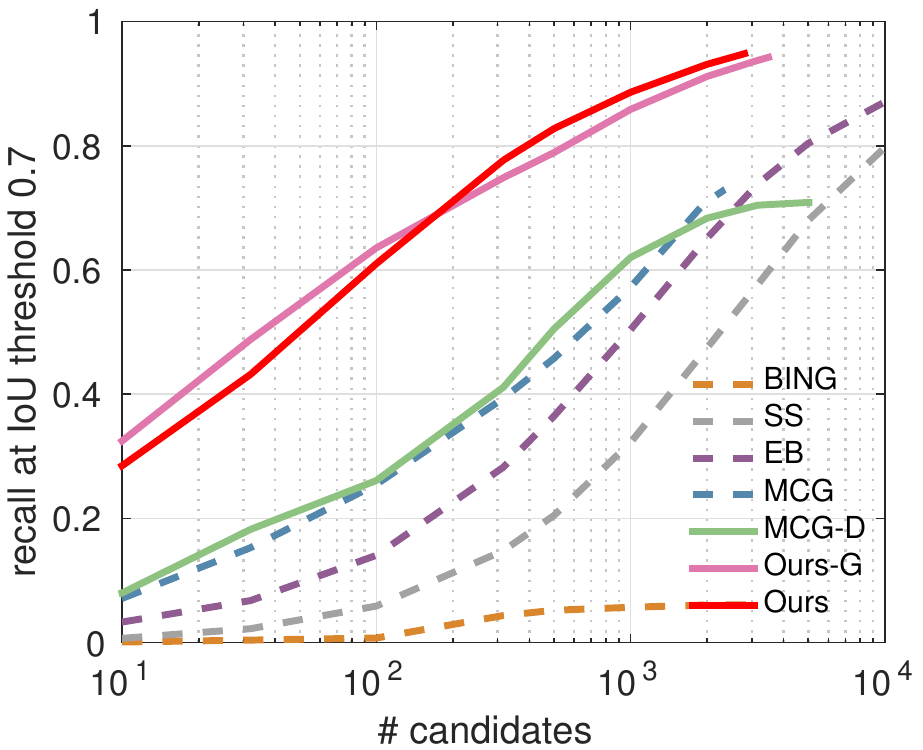}}&
\raisebox{-0.38\height}{\includegraphics[width=0.265\linewidth,trim = 5mm 0mm 0mm 0mm, clip]{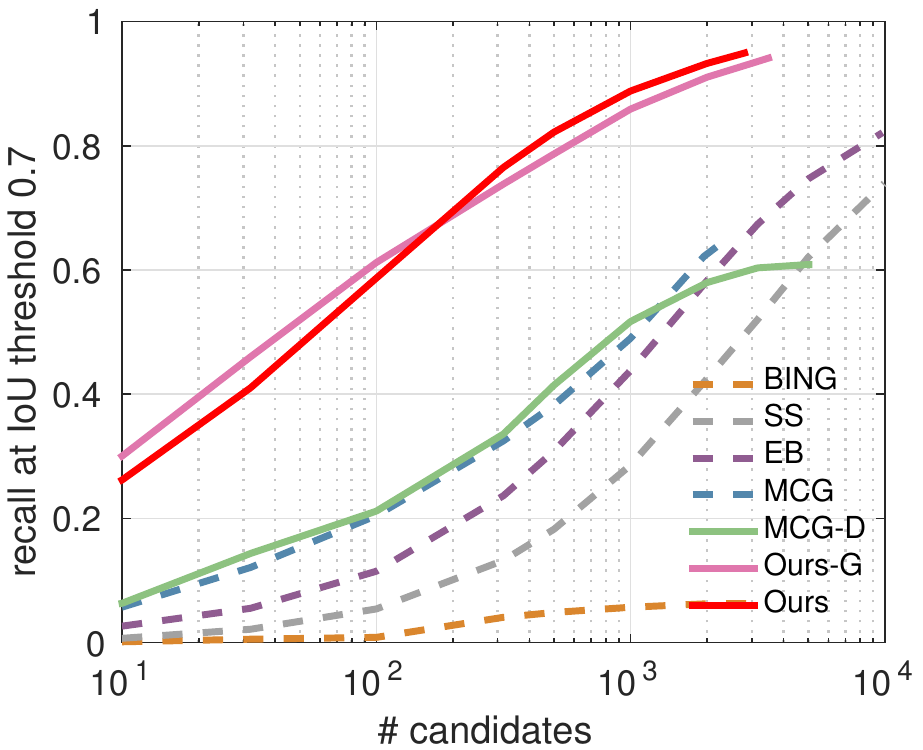}}\\
\hspace{-5mm}\rotatebox{90}{\hspace{3mm}Pedestrian} &\hspace{-1cm}
\raisebox{-0.38\height}{\includegraphics[width=0.265\linewidth,trim = 5mm 0mm 0mm 0mm, clip]{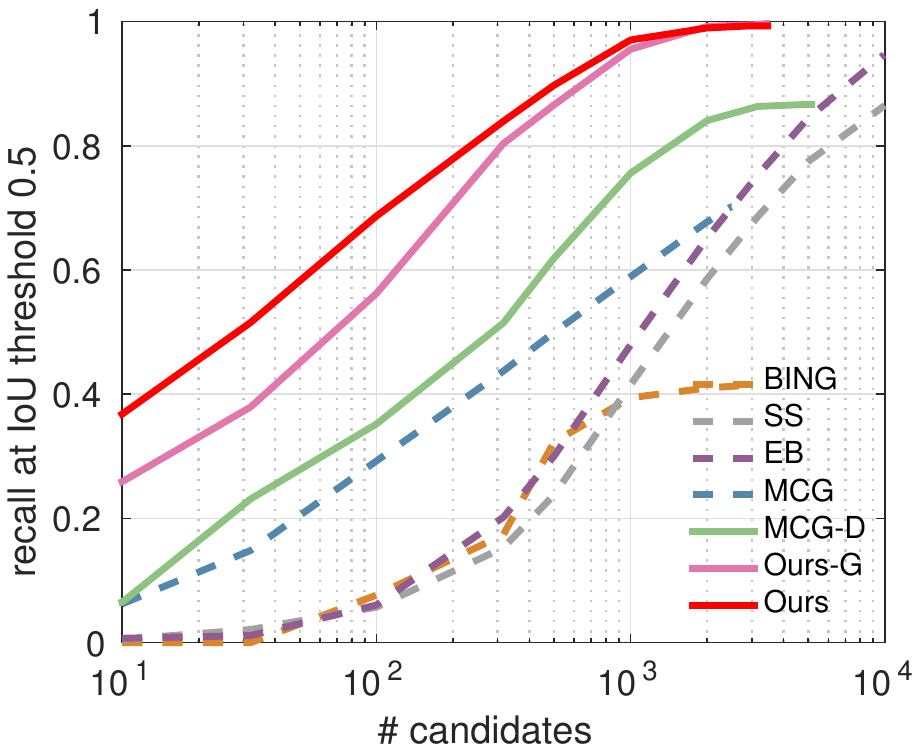}}&
\raisebox{-0.38\height}{\includegraphics[width=0.265\linewidth,trim = 5mm 0mm 0mm 0mm, clip]{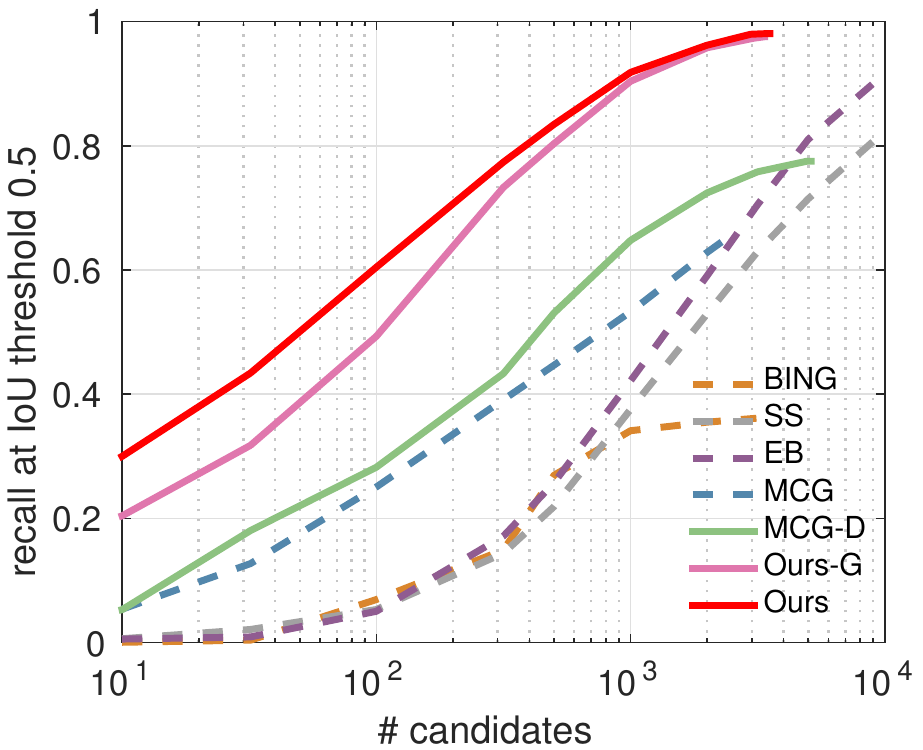}}&
\raisebox{-0.38\height}{\includegraphics[width=0.265\linewidth,trim = 5mm 0mm 0mm 0mm, clip]{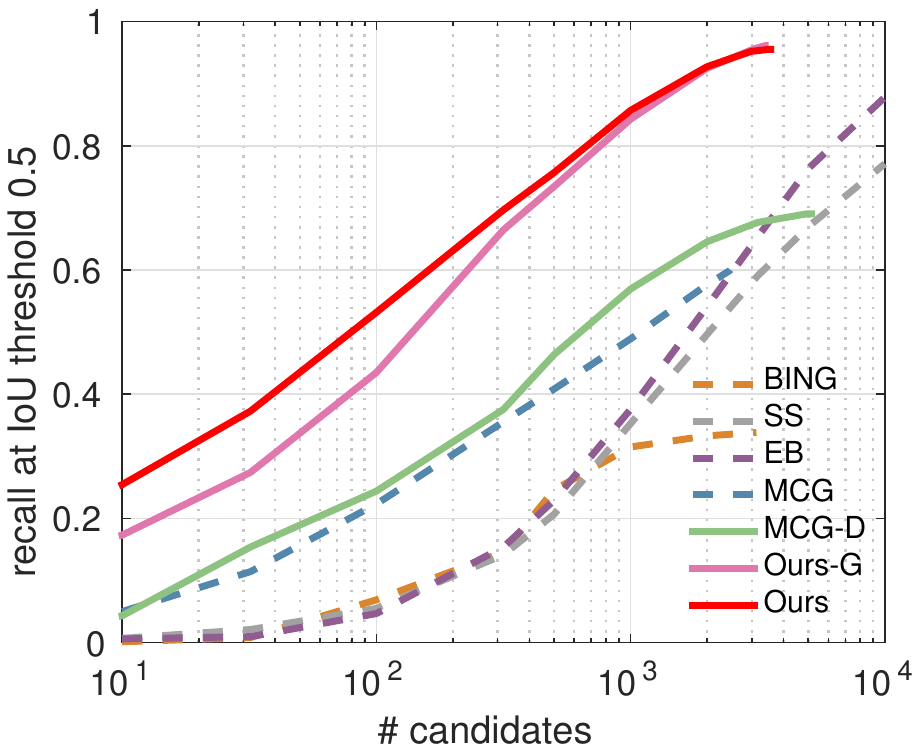}}\\
\hspace{-5mm}\rotatebox{90}{\hspace{3mm}Cyclist} &\hspace{-1cm}
\raisebox{-0.38\height}{\includegraphics[width=0.265\linewidth,trim = 5mm 0mm 0mm 0mm, clip]{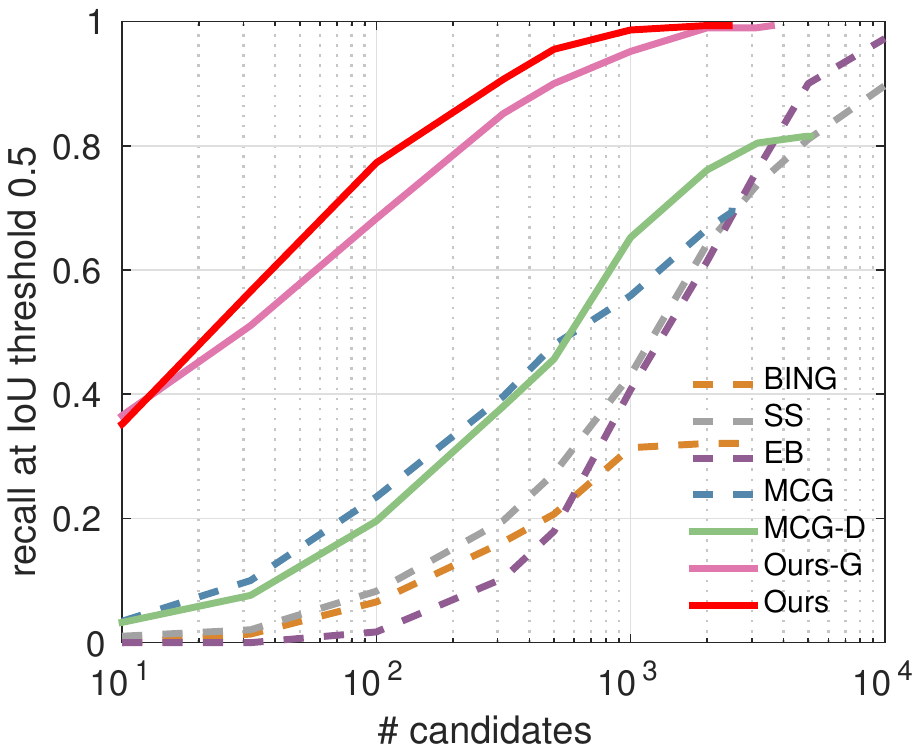}}&
\raisebox{-0.38\height}{\includegraphics[width=0.265\linewidth,trim = 5mm 0mm 0mm 0mm, clip]{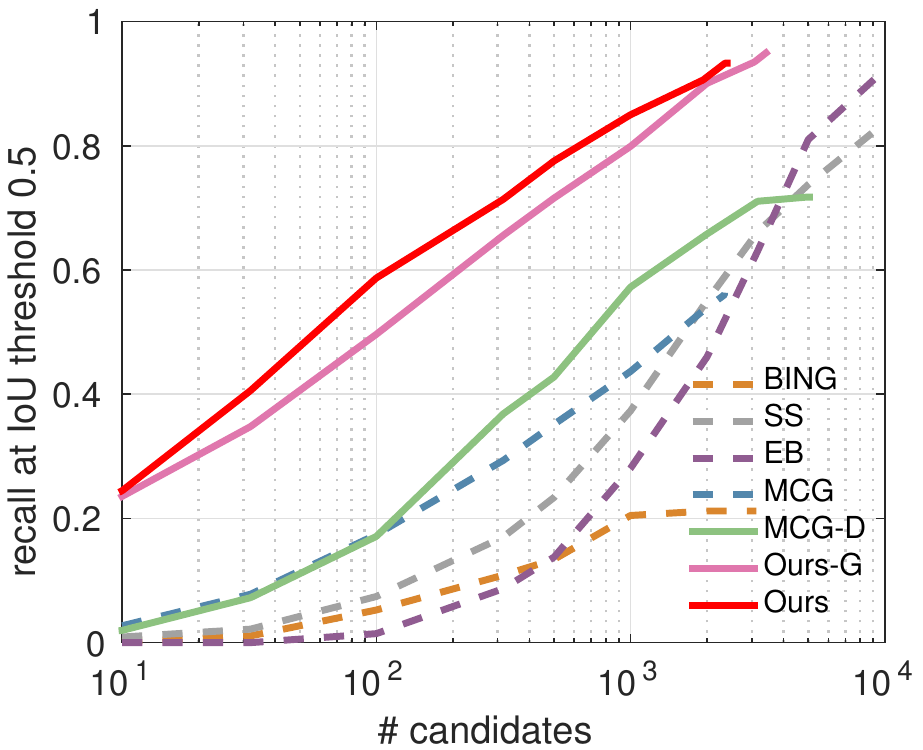}}&
\raisebox{-0.38\height}{\includegraphics[width=0.265\linewidth,trim = 5mm 0mm 0mm 0mm, clip]{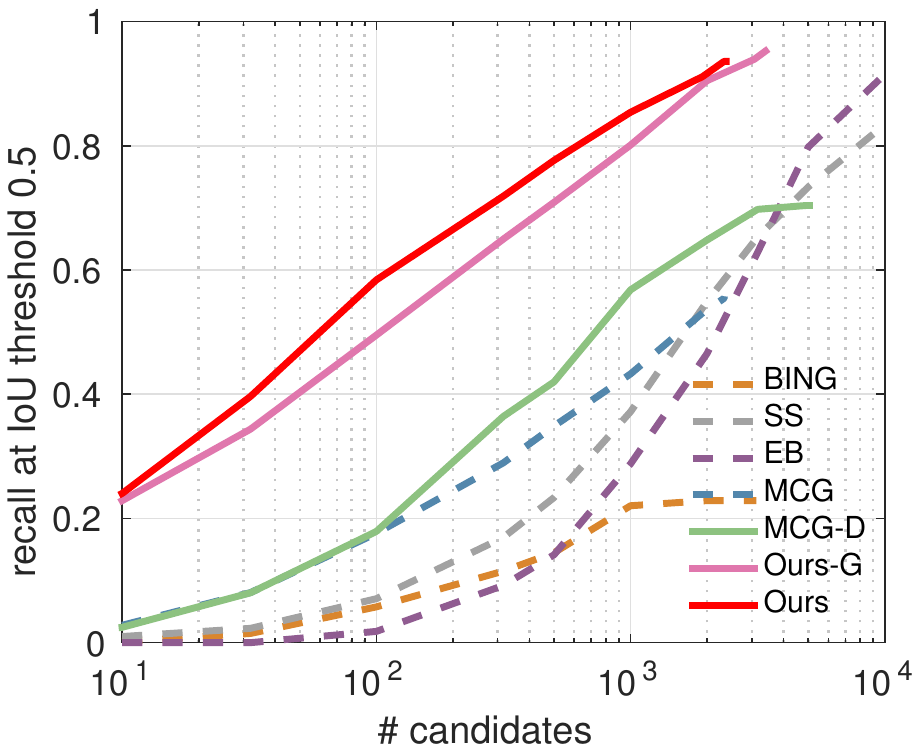}}\\
& (a) Easy & (b) Moderate & (c) Hard
\end{tabular}
\vspace{-3mm}
\caption{\textbf{2D bounding box Recall vs  number of Candidates}. {\bf ``Ours-G"}: class-independent proposals. {\bf ``Ours"}: class-dependent proposals. We use an overlap threshold of  0.7  for {\it Car}, and 0.5  for {\it Pedestrian} and {\it Cyclist}, following the KITTI evaluation protocol~\cite{kitti}. From left to right are for \emph{Easy}, \emph{Moderate}, and \emph{Hard} evaluation regimes, respectively.}
\label{fig:2d-recall-vs-cand}
\vspace{-3mm}
\end{center}
\end{figure*}

\begin{figure*}[t!]
\begin{center}
\begin{tabular}{p{3mm}ccc}
\hspace{-5mm}\rotatebox{90}{\hspace{3mm}Car} &\hspace{-1cm}
\raisebox{-0.38\height}{\includegraphics[width=0.27\linewidth,trim = 5mm 0mm 0mm 0mm, clip]{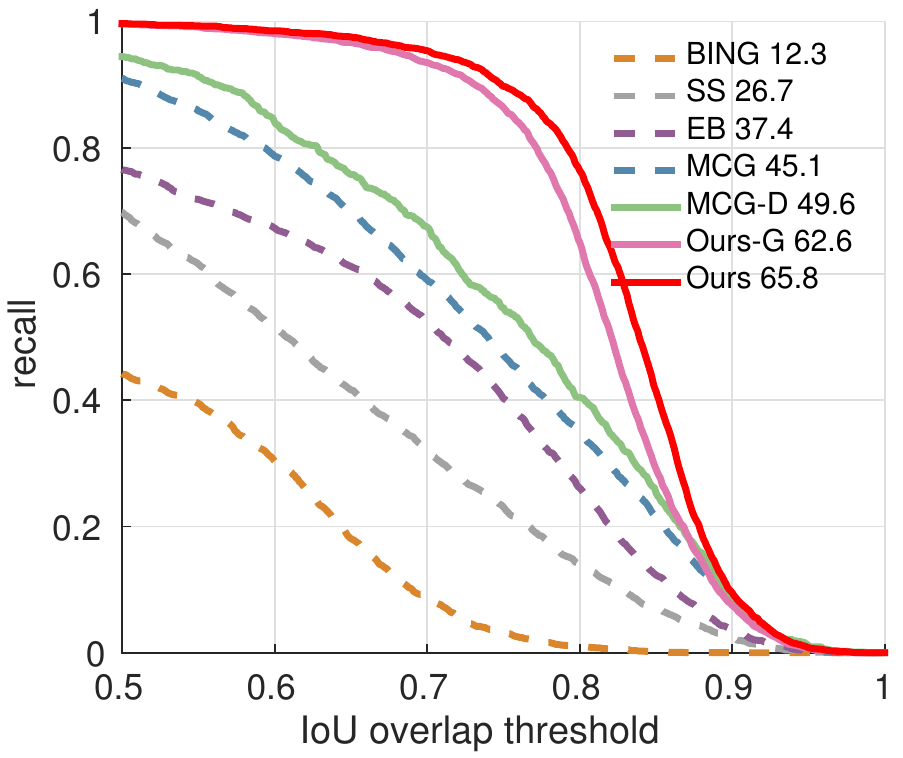}}&
\raisebox{-0.38\height}{\includegraphics[width=0.27\linewidth,trim = 5mm 0mm 0mm 0mm, clip]{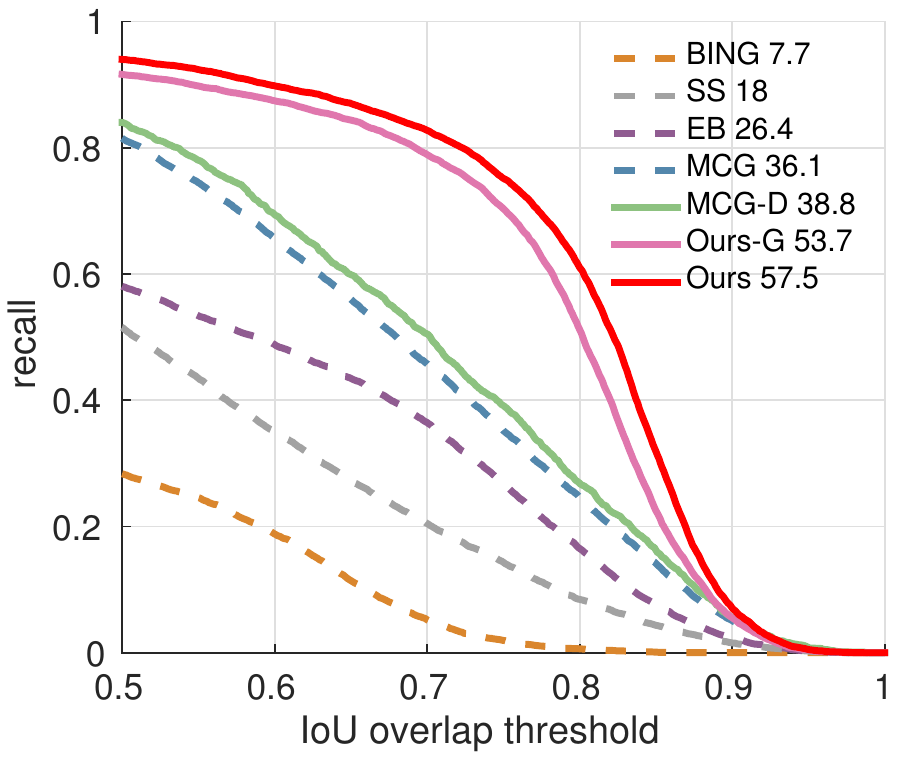}}&
\raisebox{-0.38\height}{\includegraphics[width=0.27\linewidth,trim = 5mm 0mm 0mm 0mm, clip]{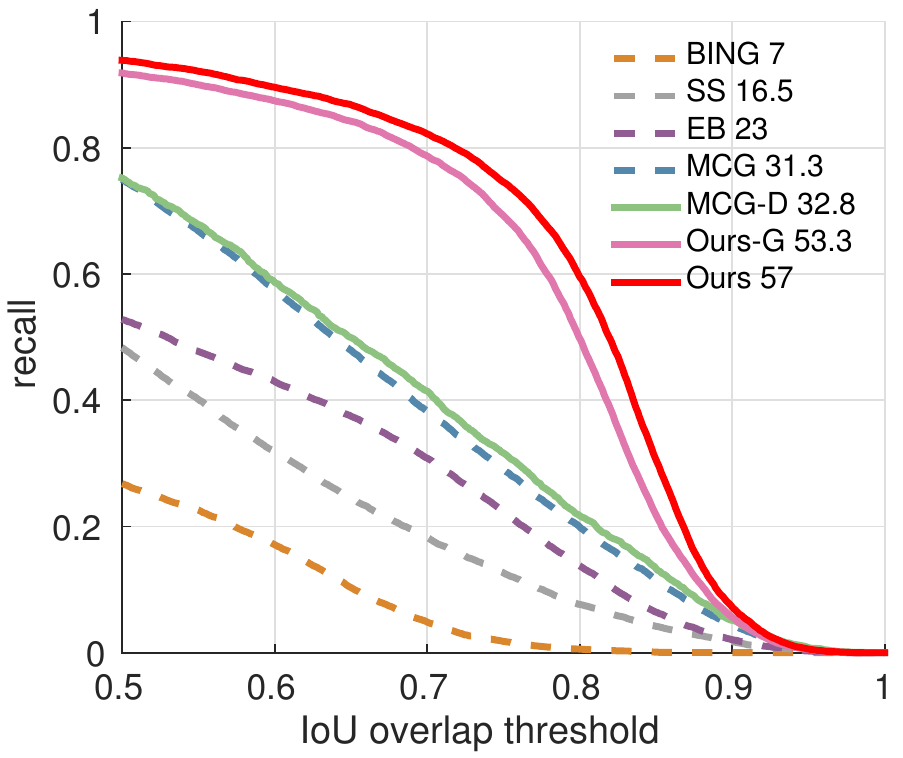}}\\
\hspace{-5mm}\rotatebox{90}{\hspace{3mm}Pedestrian} &\hspace{-1cm}
\raisebox{-0.38\height}{\includegraphics[width=0.27\linewidth,trim = 5mm 0mm 0mm 0mm, clip]{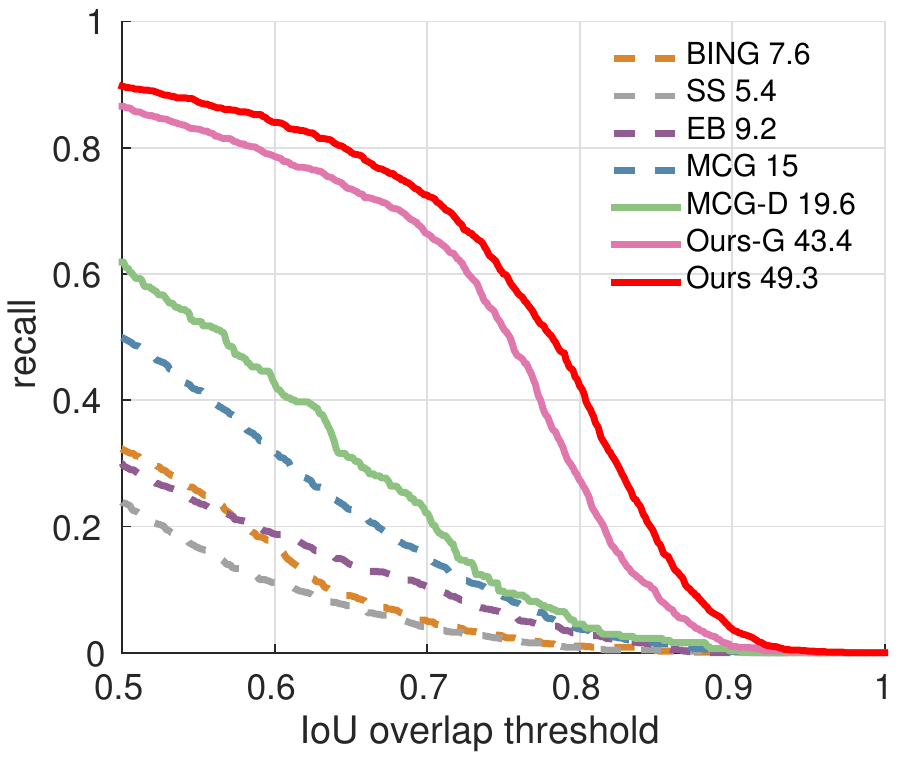}}&
\raisebox{-0.38\height}{\includegraphics[width=0.27\linewidth,trim = 5mm 0mm 0mm 0mm, clip]{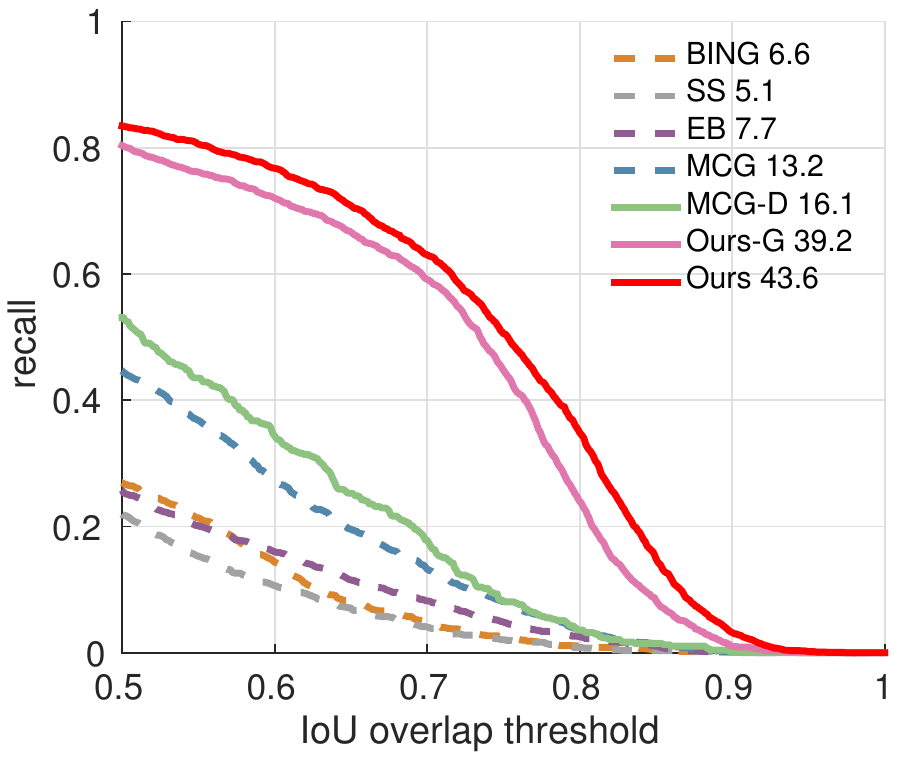}}&
\raisebox{-0.38\height}{\includegraphics[width=0.27\linewidth,trim = 5mm 0mm 0mm 0mm, clip]{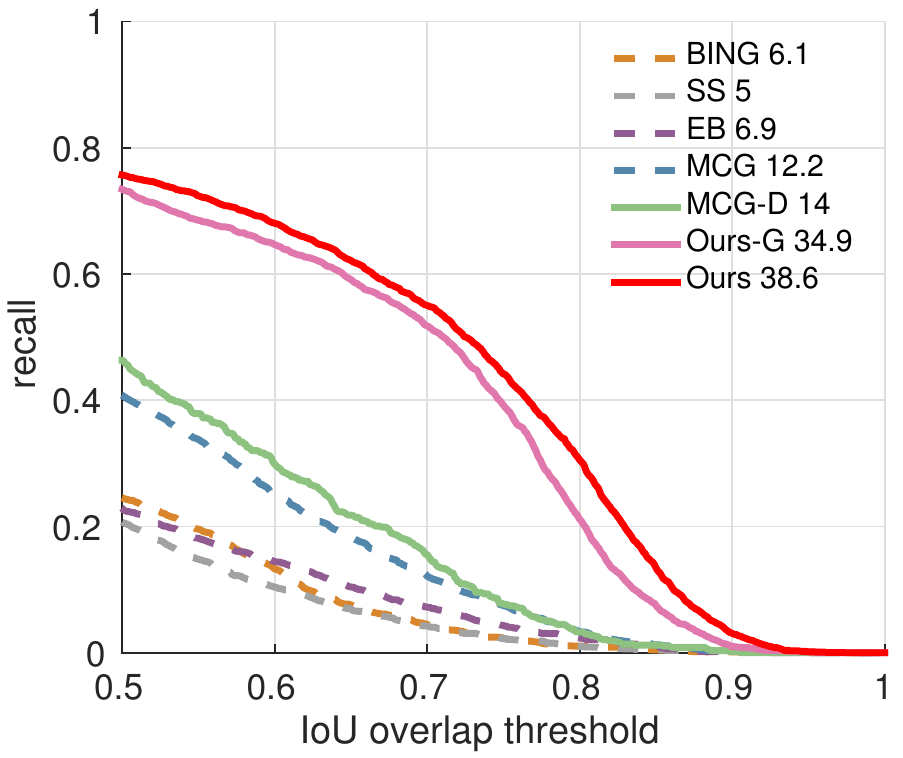}}\\
\hspace{-5mm}\rotatebox{90}{\hspace{3mm}Cyclist} &\hspace{-1cm}
\raisebox{-0.38\height}{\includegraphics[width=0.27\linewidth,trim = 5mm 0mm 0mm 0mm, clip]{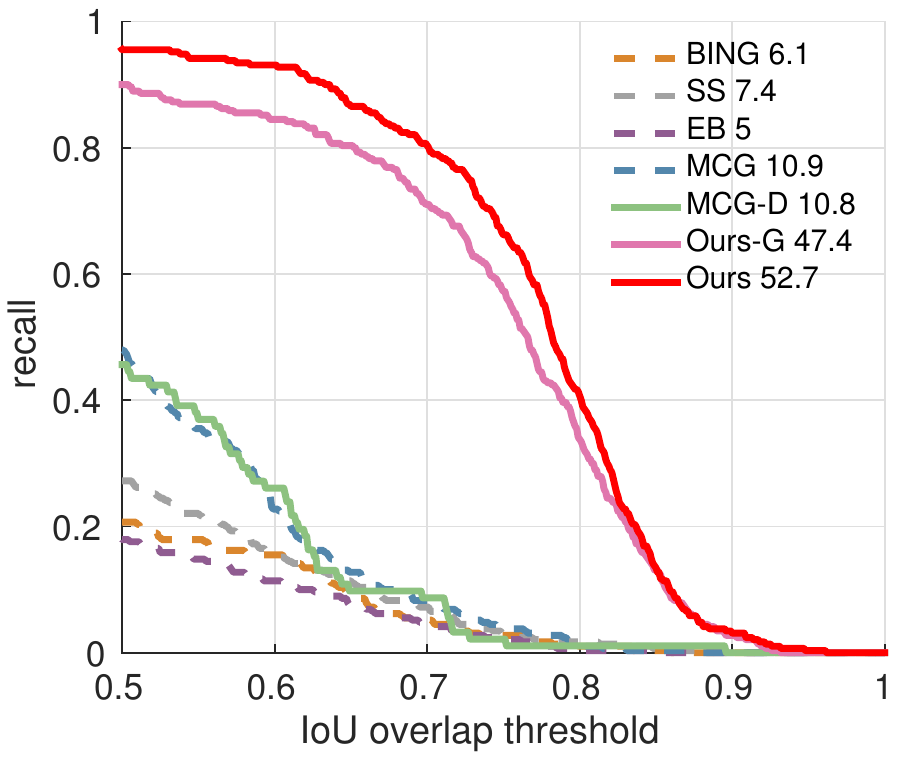}}&
\raisebox{-0.38\height}{\includegraphics[width=0.27\linewidth,trim = 5mm 0mm 0mm 0mm, clip]{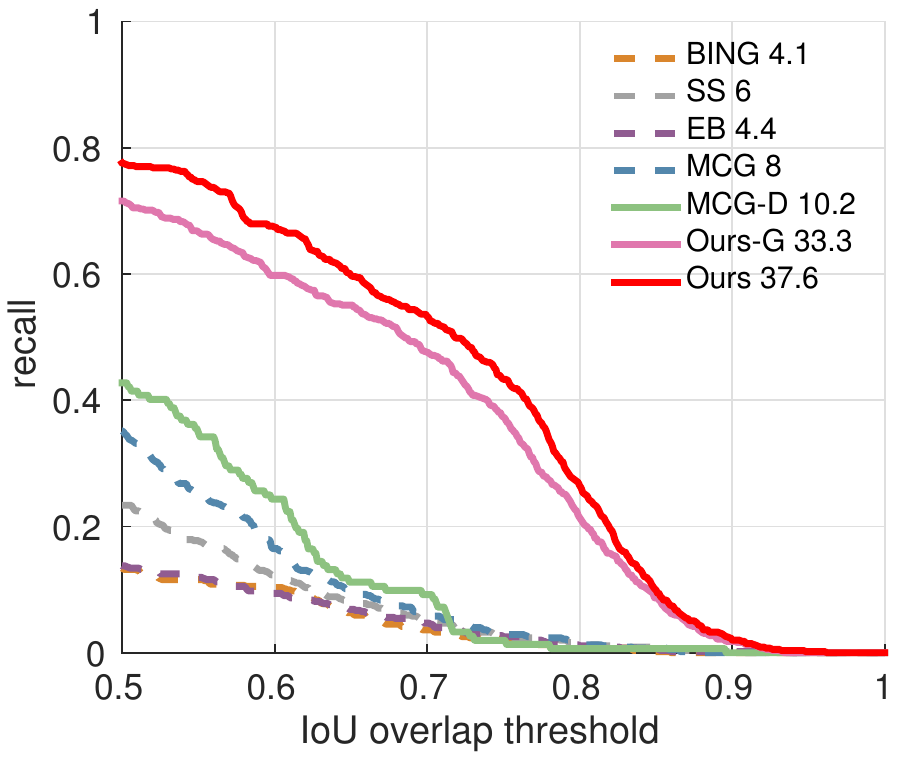}}&
\raisebox{-0.38\height}{\includegraphics[width=0.27\linewidth,trim = 5mm 0mm 0mm 0mm, clip]{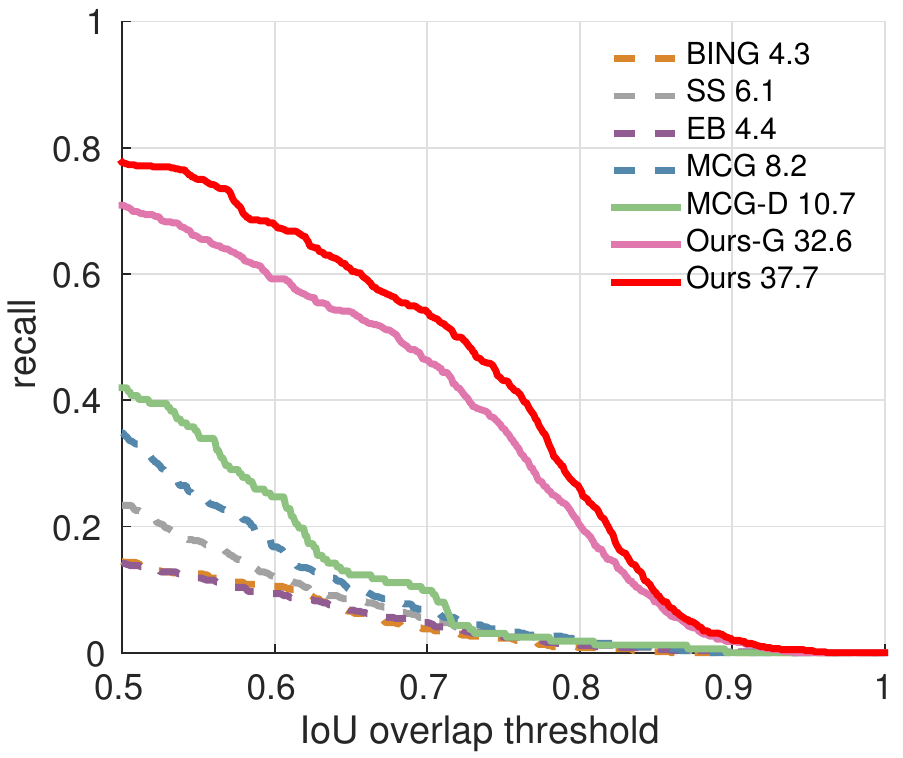}}\\
& (a) Easy & (b) Moderate & (c) Hard
\end{tabular}
\vspace{-3mm}
\caption{\textbf{2D bounding box Recall vs IoU for 500 proposals}. The number next to the label indicates the average recall (AR).}
\label{fig:2d-recall-vs-iou-500}
\vspace{-3mm}
\end{center}
\end{figure*}

\begin{figure*}[t!]
	\begin{center}
		\begin{tabular}{ccc}
			\raisebox{-0.38\height}{\includegraphics[width=0.27\linewidth,trim = 5mm 0mm 0mm 0mm, clip]{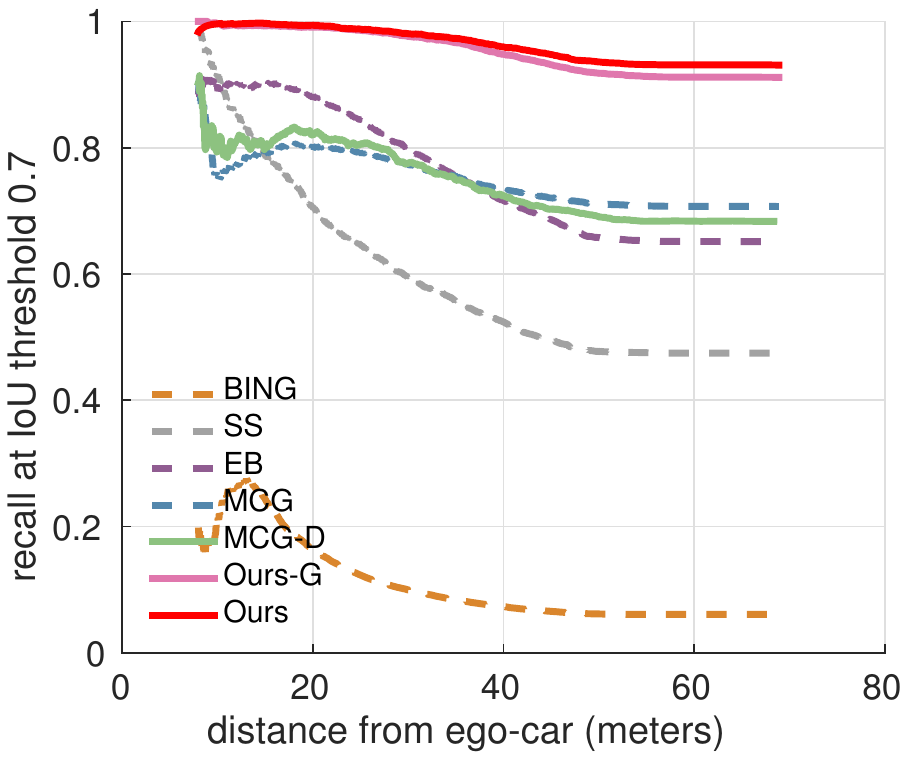}}&
			\raisebox{-0.38\height}{\includegraphics[width=0.27\linewidth,trim = 5mm 0mm 0mm 0mm, clip]{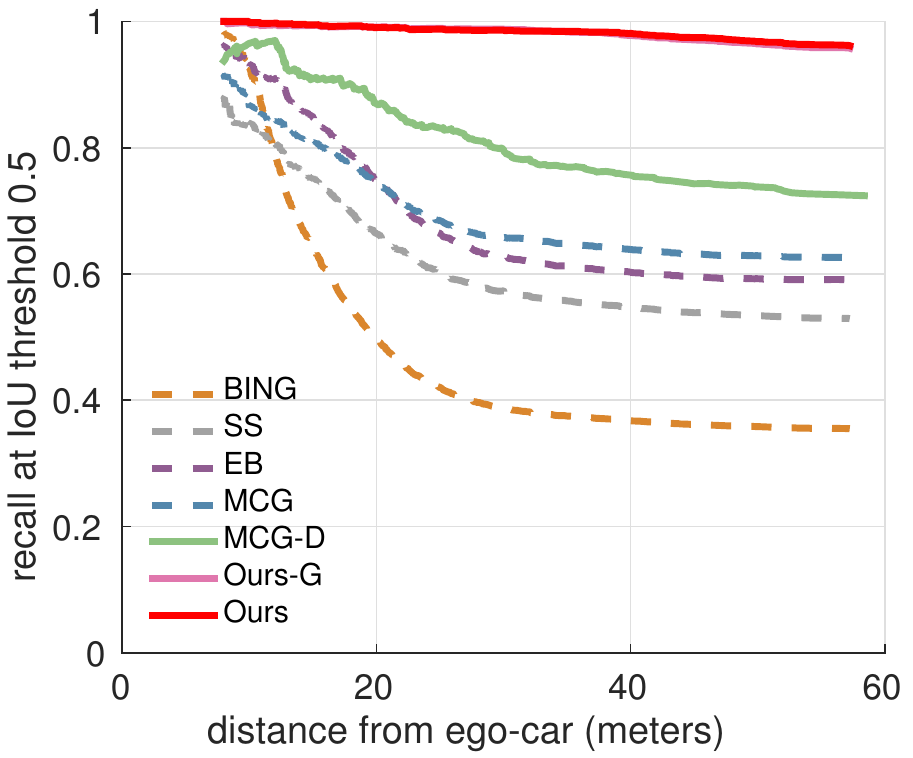}}&
            \raisebox{-0.38\height}{\includegraphics[width=0.27\linewidth,trim = 5mm 0mm 0mm 0mm, clip]{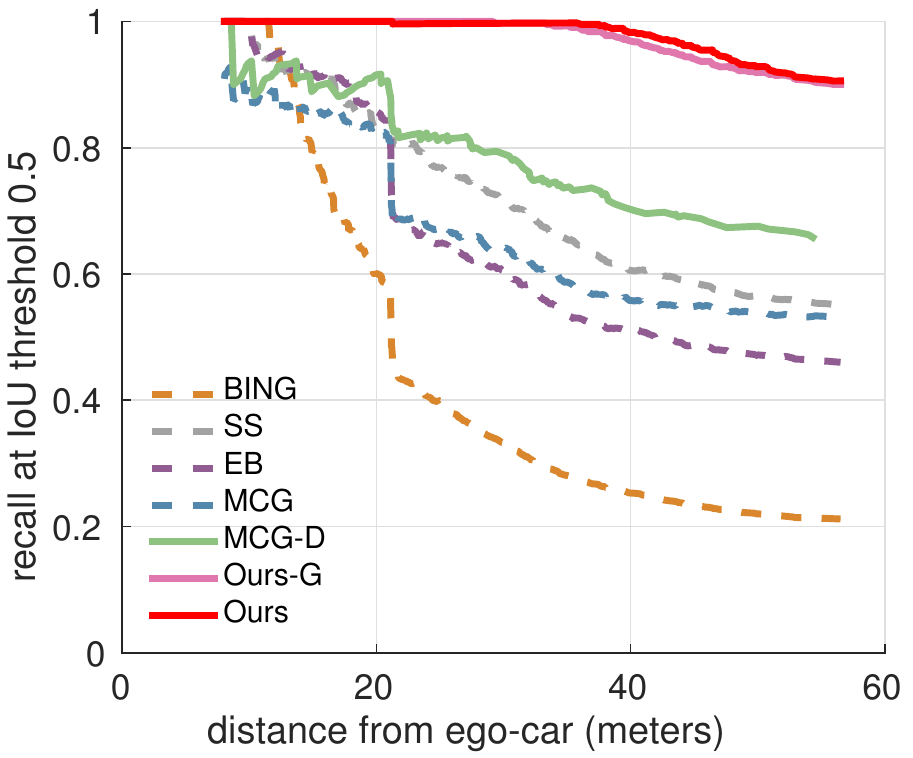}} \\
			Car & Pedestrian & Cyclist \\
		\end{tabular}
	    \vspace{-3mm}
        \caption{\textbf{2D bounding box Recall vs Distance with 2000 proposals on \emph{moderate} data}. We use overlap threshold of  0.7  for {\it Car}, and 0.5  for {\it Pedestrian}, {\it Cyclist}.}
		\label{fig:2d-recall-dist}
	\end{center}
	\vspace{-4mm}
\end{figure*}

\begin{figure*}[t!]
\begin{center}
	\begin{tabular}{ccc}
		\raisebox{-0.38\height}{\includegraphics[width=0.27\linewidth,trim = 5mm 0mm 0mm 0mm, clip]{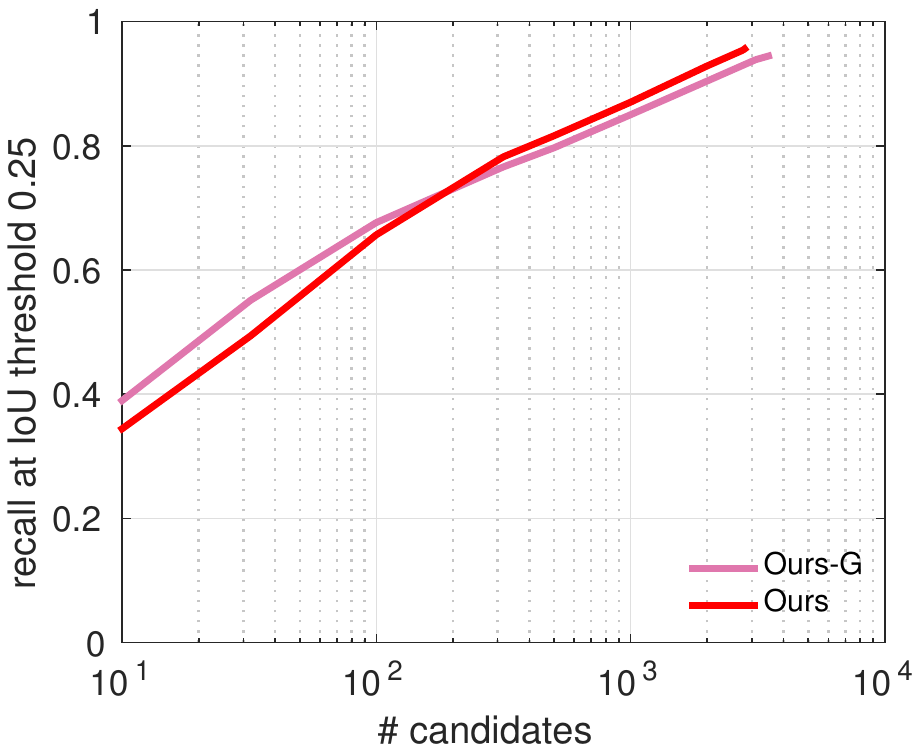}}&
		\raisebox{-0.38\height}{\includegraphics[width=0.27\linewidth,trim = 5mm 0mm 0mm 0mm, clip]{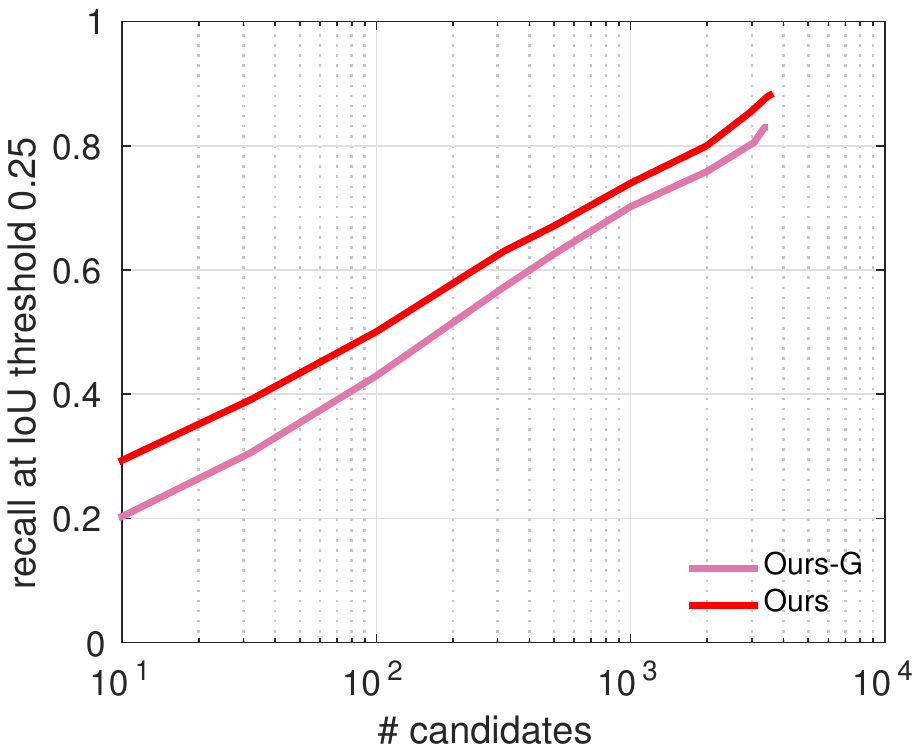}}&
		\raisebox{-0.38\height}{\includegraphics[width=0.27\linewidth,trim = 5mm 0mm 0mm 0mm, clip]{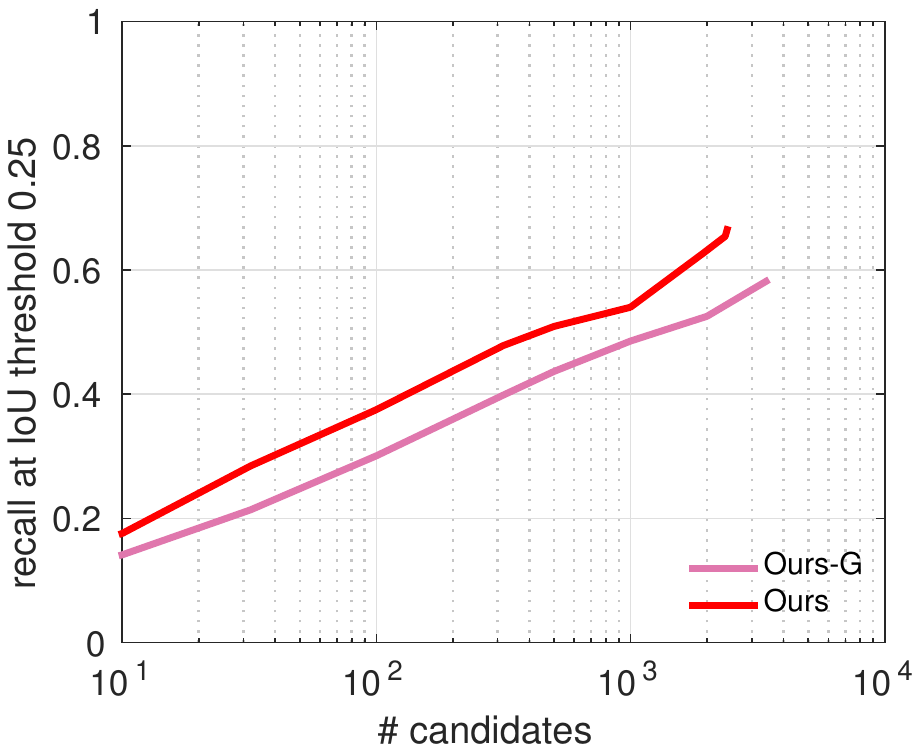}}\\
		Car & Pedestrian & Cyclist \\
	\end{tabular}
    \vspace{-3mm}
	\caption{\textbf{3D bounding box Recall vs \#Candidates on \emph{moderate} data}. 3D IoU threshold is set to 0.25.}
	\label{fig:3d-recall-vs-cand}
	\vspace{-3mm}
\end{center}
\end{figure*}

\vspace{-1mm}
\subsection{CNN Scoring with Depth Features}
Despite only using appearance features (i.e. RGB image), the basic network described above already performs very well in practice. To take advantage of depth information in CNN scoring process, we further compute a depth image encoded with HHA features~\cite{guptaECCV14}.
HHA has three channels which represent the disparity map, height above the ground, and the angle of the normal at each pixel with respect to the gravity direction.
We explore two approaches to learn feature representation with both RGB and depth images as input. 
The first approach is a single-stream network, which directly combines RGB channels and HHA channels to form a 6-channel image, and feed it to the network. This architecture is exactly the same as the basic model in Fig.~\ref{fig:net}, except that its input is a 6-channel image.
The second approach is a two-stream network which learns features from RGB and HHA images respectively, as shown in Fig.~\ref{fig:rgbd-nets}. Note that the two-stream network has almost double the parameters of the single-stream model, and thus requires more GPU memory in training.

\vspace{-1mm}
\subsection{Implementation Details}
In our object detection experiments, we use class-specific weights for proposal generation.
For network training, we choose training samples based on the IoU overlap threshold of the 2D bounding box proposals and ground truth boxes.
Since KITTI uses different overlap criteria for {\it Car} and {\it Pedestrian}/{\it Cyclist}, we set the threshold  for {\it Car} to 0.7  and 0.5 for the {\it Pedestrian}/{\it Cyclist} classes.
By default, we use the VGG-16 network~\cite{verydeep} trained on ImageNet to initialize our networks.
We initialize the context branch by copying the weights of the fully-connected layers from the pre-trained model.
For the two-stream RGB-HHA model, which requires more GPU memory, we use the 7-layer VGG\_CNN\_M\_1024 network~\cite{chatfield2014return}.
The weights for the HHA channels/branch are copied from the corresponding RGB channels/branch.
For the one-stream model, we fine-tune all the layers starting from \emph{conv1}.
For the two-stream model, all layers are fine-tuned for the HHA branch, and only layers above \emph{conv2} are fine-tuned for the RGB branch.
We use bilinear interpolation to upscale the input image by a factor of 3.5, which is crucial to achieve very good performance since objects in KITTI imagery are typically small. We use a single scale for input images in both training and testing. We run SGD and set the initial learning rate to 0.001. After 30K iterations we reduce it to 0.0001 and run another 10K iterations. Training proposals are sampled in a image-centric manner with a batch size of 1 for images and 128 for proposals. At test time, the network takes around 2s to evaluate one image with 2K proposals on a Titan X GPU.

\section{Experimental Evaluation}
\vspace{-1mm}

We evaluate our approach on the challenging KITTI detection benchmark~\cite{kitti}, which has 7,481 training and 7,518 test images. The benchmark contains three object classes: {\it Car},  {\it Pedestrian}, and {\it Cyclist}. Evaluation is done for each class in three regimes: \emph{Easy}, \emph{Moderate} and \emph{Hard}, which contain objects of different occlusion and truncation levels. We split the 7,481 training images into a \emph{training} set (3,712 images) and a \emph{validation} set (3,769 images). We ensure that the training and validation set do not contain images from the same video sequences, and evaluate the performance of our proposals on the validation set.

\vspace{-1mm}
\paragraph{Metrics:}
To evaluate proposals, we use the oracle recall as the metric, following ~\cite{van2011segmentation,Hosang2015PAMI}. A ground truth object is said to be recalled if at least one proposal overlaps with it with IoU above a certain threshold. We set the IoU threshold to 70\% for {\it Car}, and 50\% for {\it Pedestrian} and {\it Cyclist}, following the standard KITTI's setup. The oracle recall is then computed as the percentage of recalled ground truth objects. We also report average recall (AR)~\cite{Hosang2015PAMI}, which has been shown to be highly correlated with the object detection performance. 

We also evaluate the whole pipeline of our 3D object detection model on KITTI's two tasks: 2D object detection, and joint 2D object detection and orientation estimation. Following the standard KITTI setup, we use the Average Precision (AP$_{\text{2D}}$) metric for 2D object detection task, and Average Orientation Similarity (AOS)~\cite{kitti} for joint 2D object detection and orientation estimation task.

For the task of 3D object detection, we evaluate the performance using two metrics: Average Precision (AP$_{\text{3D}}$) using 3D bounding box overlap measure, and Average Localization Precision (ALP).
Similar to the setting in~\cite{SongECCV14}, we use 25\% overlap criteria for the 3D bounding box overlap measure.
Average Localization Precision is computed similarly to AP, except that the bounding box overlap is replaced by 3D localization precision.
We consider a predicted 3D location to be correct if its distance to the ground truth 3D location is smaller than certain threshold.
Note that this 3D localization precision measure is used when computing both precision and recall.

\begin{table}[t!]
\caption{Running time of different proposal methods.}
\vspace{-3mm}
\label{tab:running-time}
\begin{center}
\renewcommand{\arraystretch}{1.3}
\addtolength{\tabcolsep}{15pt}
\begin{tabular}{c||c}
\hline 
Method & Time (sec.) \\
\hline\hline
BING~\cite{BingObj2014}   & 0.01 \\
Selective Search (SS)~\cite{van2011segmentation} & 15 \\
EdgeBoxes (EB)~\cite{zitnick2014edge}  & 1.5 \\
MCG~\cite{ArbelaezCVPR14}  & 100 \\ 
MCG-D~\cite{guptaECCV14}  & 160  \\
\hline \hline
Ours & 1.2 \\
\hline
\end{tabular}
\vspace{-3mm}
\end{center}
\end{table}

\vspace{-1mm}
\paragraph{Baselines:}
We compare our proposal method with several top-performing approaches on the validation set:  MCG-D~\cite{guptaECCV14}, MCG~\cite{ArbelaezCVPR14}, Selective Search (SS)~\cite{van2011segmentation}, BING~\cite{BingObj2014}, and EdgeBoxes (EB)~\cite{zitnick2014edge}. 

\vspace{-2mm}
\subsection{Proposal Recall}
We evaluate recall of the two variants of our approach: class-dependent and class-independent proposals. We denote the class-independent variant as `Ours-G'. 

\vspace{-1mm}
\paragraph{2D Bounding Box Recall:}  
Fig.~\ref{fig:2d-recall-vs-cand} shows recall as a function of the number of candidates. 
We can see that in general our class-specific proposals perform slightly better than the class-independent variant. This suggests the advantage of exploiting size priors tailored to each class.
By using 1000 proposals, our approach achieves almost 90\% recall for {\it Car} in the \emph{Moderate} and \emph{Hard} regimes, while for \emph{Easy} we need only 200 proposals to reach the same recall. In contrast, other methods saturate or require orders of magnitude more proposals to reach 90\% recall. For {\it Pedestrian} and {\it Cyclist} our approach achieves similar improvements over the baselines. Note that while our approach uses depth-based features, MCG-D combines depth and appearance features, and all other methods use only appearance features. This suggests the importance of 3D reasoning in the domain of autonomous driving. In Fig.~\ref{fig:2d-recall-vs-iou-500}, we show recall as a function of the IoU overlap for 500 proposals. We obtain significantly higher recall over the baselines across all IoU overlap levels, particularly for {\it Cyclist}.

\begin{figure}[t!]
\begin{center}
\includegraphics[width=0.7\linewidth,trim = 0mm 0mm 0mm 0mm, clip]{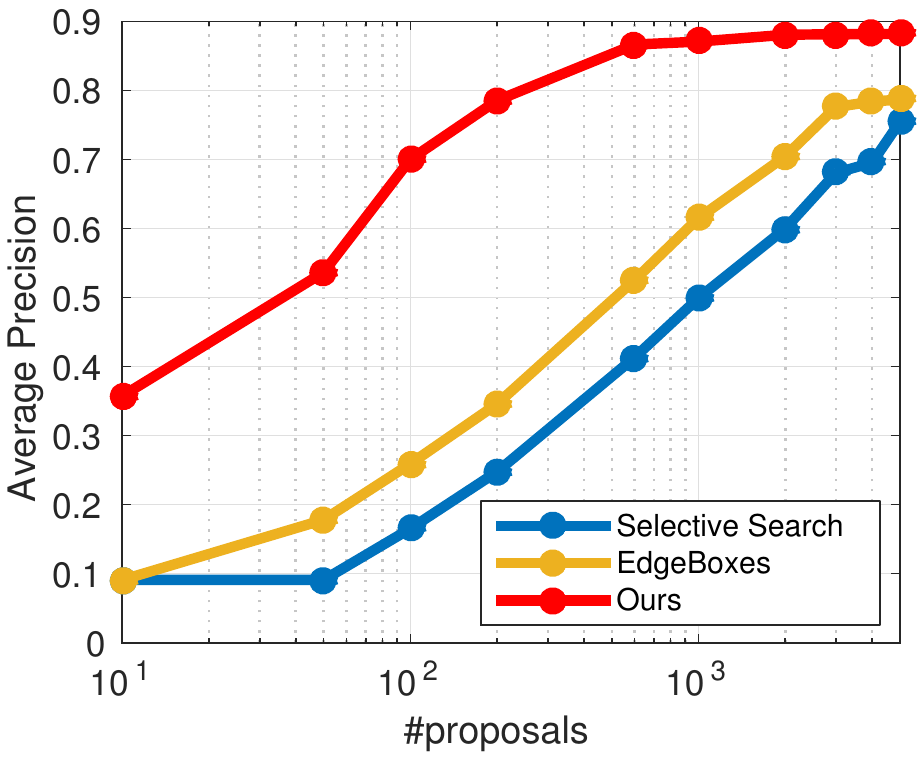}		
\vspace{-3mm}
\caption{\small AP$_{\text{2D}}$ vs \#proposals on {\it Car} for the \emph{Moderate} setting.}
\label{fig:ap_proposals}
\vspace{-3mm}
\end{center}
\end{figure}

\vspace{-1mm}
\paragraph{Recall vs Distance:} We also plot recall as a function of the object's distance from the ego-car in Fig.~\ref{fig:2d-recall-dist}. We can see that our approach remains at a very high recall even at large distances ($>$40m), while the performance of all other proposal methods drops significantly with the increasing distance. This shows the advantage of our approach in the autonomous driving scenario.

\vspace{-1mm}
\paragraph{3D Bounding Box Recall:} Since the unique benefit of our approach is the output of 3D bounding boxes, we also evaluate recall in 3D, i.e., using 3D bounding box overlap. We set IoU overlap threshold computed in 3D between the ground-truth 3D bounding boxes and ours to 0.25. As shown in Fig.~\ref{fig:3d-recall-vs-cand}, when using 2000 proposals, our approach achieves around 90\%, 80\% and 60\% 3D recall for {\it Car}, {\it Pedestrian} and {\it Cyclist}, respectively.

\vspace{-1mm}
\paragraph{Running Time:} The running time of different proposal methods are shown in Table~\ref{tab:running-time}. Our approach is fairly efficient and takes 1.2s on a single core. The only faster proposal method is BING, which however achieves a much lower recall than our approach (Fig.~\ref{fig:2d-recall-vs-cand}).

\begin{table*}[t!]
\caption{Average Precision (AP$_{\text{2D}}$) (in \%) on the test set of the KITTI Object Detection Benchmark. $^\dagger$: LIDAR-based methods; $^\ddagger$: using both LIDAR and image data.}
\label{tab:ap_test}
\vspace{-2mm}
\begin{small}
\addtolength{\tabcolsep}{1.8pt}
\begin{tabular}{c||c|c|c||c|c|c||c|c|c}
\hline
\multirow{2}{*}{Method} & \multicolumn{3}{c||}{Cars} & \multicolumn{3}{c||}{Pedestrians} & \multicolumn{3}{c}{Cyclists} \\
\cline{2-10}
& Easy & Moderate & Hard & Easy & Moderate & Hard & Easy & Moderate & Hard \\
\hline
mBoW$^\dagger$~\cite{behley2013iros}  & 36.02  & 23.76  & 18.44   & 44.28  & 31.37  & 30.62  & 28.00  & 21.62  & 20.93   \\
CSoR$^\dagger$~\cite{csor}  & 34.79  & 26.13  & 22.69 &- &- &- &- &- &-   \\
Vote3D$^\dagger$~\cite{vote3d}   & 56.80  & 47.99  & 42.57 & 44.48 & 35.74 &  33.72  & 41.43  & 31.24  & 28.60  \\
VeloFCN$^\dagger$~\cite{velofcn}  & 71.06 & 53.59 & 46.92 &- &- &- &- &- &-   \\
MV-RGBD-RF$^\ddagger$~\cite{AGonzalez2015} & 76.40 & 69.92 & 57.47 & 73.30  & 56.59  & 49.63  & 52.97  & 42.6  & 37.42   \\
SubCat~\cite{BarTITS15} & 84.14  & 75.46  & 59.71  &  54.67  & 42.34  & 37.95  & - & -  & -  \\
DA-DPM~\cite{adas2014} & - & - & - & 56.36  & 45.51  & 41.08  &   -   &    -   &   -     \\
Fusion-DPM~\cite{CPremebida_IROS2014} & - & - & - & 59.51  & 46.67  & 42.05  &   -  &  -    &  -   \\
R-CNN~\cite{HosangCVPR15} & - & - & - & 61.61 & 50.13  & 44.79  &  -   &   -   &  -  \\
pAUCEnsT~\cite{Paul2014Pedestrian} & - & - & - & 65.26  & 54.49  & 48.60  & 51.62  & 38.03  & 33.38  \\
FilteredICF~\cite{Zhang2015FilteredICF} & - & - & - & 67.65 & 56.75 & 51.12 & -   & -   &  -  \\
DeepParts~\cite{DeepParts2015} & - &- &- & 70.49 & 58.67 & 52.78 &- &- &-\\
CompACT-Deep~\cite{CompACT2015} &- &- &- & 70.69 & 58.74 & 52.71 &- &- &-\\
3DVP~\cite{xiangcvpr15} & 87.46 & 75.77  & 65.38  & - & - & - & - & - & - \\
AOG~\cite{CarAOG_ECCV2014} & 84.80 & 75.94 & 60.70 & - &  -   &  - &  - &  - & - \\
Regionlets~\cite{LongACCV14} & 84.75  & 76.45  & 59.70  & 73.14 & 61.15 & 55.21 & 70.41  & 58.72 &  51.83 \\
spLBP~\cite{spLBP} & 87.19 & 77.40 & 60.60 & - & - & - & - & - & -\\
Faster R-CNN~\cite{renNIPS15fasterrcnn} & 86.71 & 81.84 & 71.12 & 78.86 & 65.90 & 61.18 & 72.26 & 63.35 & 55.90 \\
\hline
Ours & {\bf 93.04}  & {\bf 88.64}  & {\bf 79.10}  & {\bf 81.78} & {\bf 67.47}  & {\bf 64.70} &  {\bf 78.39} & {\bf 68.94} & {\bf 61.37}  \\
\hline
\end{tabular}
\end{small}
\end{table*}

\begin{table*}[t!]
\caption{AOS scores (in \%) on the test set of KITTI's Object Detection and Orientation Estimation Benchmark. $^\dagger$: LIDAR-based methods.}
\label{tab:aos_test}
\vspace{-2mm}
\begin{small}
\addtolength{\tabcolsep}{1.8pt}
\begin{tabular}{c||c|c|c||c|c|c||c|c|c}
\hline
\multirow{2}{*}{Method} & \multicolumn{3}{c||}{Cars} & \multicolumn{3}{c||}{Pedestrians} & \multicolumn{3}{c}{Cyclists} \\
\cline{2-10}
& Easy & Moderate & Hard & Easy & Moderate & Hard & Easy & Moderate & Hard \\
\hline
CSoR$^\dagger$~\cite{csor}  & 33.97  & 25.38  & 21.95  &  - &  - &  -   & - &  - &  - \\
VeloFCN$^\dagger$~\cite{velofcn}  & 70.58 & 52.84 & 46.14 &  - &  - &  -   & - &  - &  - \\
LSVM-MDPM-sv~\cite{Geiger11,dpm} & 67.27  & 55.77  & 43.59   & 43.58  & 35.49  & 32.42  &  27.54  & 22.07  & 21.45  \\
DPM-VOC+VP~\cite{PepikPAMI15} & 72.28  & 61.84  & 46.54   & 53.55 & 39.83  & 35.73 & 30.52  &  23.17  & 21.58  \\
OC-DPM~\cite{bojan13cvpr} &  73.50  &  64.42  & 52.40  &  - &  - &  -   &  - &  - &  - \\
SubCat~\cite{BarTITS15} & 83.41  & 74.42  & 58.83 & 44.32  &  34.18  & 30.76  & -  &  -  &  -   \\
3DVP~\cite{xiangcvpr15} & 86.92 &  74.59  & 64.11 & - & - & - & - & - & - \\
\hline
Ours & {\bf 91.44} & {\bf 86.10 } & {\bf 76.52 }   & {\bf 72.94}  & {\bf 59.80} & {\bf 57.03} & {\bf 70.13} &  {\bf 58.68} & {\bf 52.35} \\
\hline
\end{tabular}
\end{small}
\end{table*}

\vspace{-1mm}
\paragraph{Qualitative Results:}
Figs.~\ref{fig:pcl-vis} and~\ref{fig:pcl-vis-ped} show qualitative results for cars and pedestrians in KITTI images. We show the input RGB image, top 100 proposals, the 3D ground truth boxes, as well as the best proposals that have the highest IoU with ground truth boxes (chosen from 2K proposals). We can see that our proposals are able to locate objects precisely even for the distant and occluded objects. 

\subsection{2D Object Detection and Orientation Estimation}

\vspace{-1mm}
\paragraph{Performance on KITTI Test:} We test our approach on KITTI's Test for two tasks: 2D object detection, and joint 2D object detection and orientation estimation. We choose our best model based on the validation set, and submit our results to the benchmark. We choose the class specific variant of our method for generating proposals. For the detection network, we use the single-stream model with RGB image as input.
As shown in Table~\ref{tab:ap_test} and Table~\ref{tab:aos_test}, our approach significantly outperforms all published methods, including both image-based and LIDAR-based methods.
In terms of 2D object detection, our approach outperforms Faster R-CNN~\cite{renNIPS15fasterrcnn}, which is the state-of-the-art model on ImageNet, MS COCO, and PASCAL datasets. We achieve 7.98\%, 3.52\% and 5.47\% improvement in AP$_{\text{2D}}$ for {\it Car}, {\it Pedestrian}, and {\it Cyclist}, in the \emph{Hard} regime.

For the task of joint 2D object detection and orientation estimation, our approach also outperforms all methods by a large margin. In particular, our approach obtains $\sim$12\% higher AOS scores than 3DVP~\cite{xiangcvpr15} on {\it Car} in the \emph{Moderate} and \emph{Hard} regimes. For {\it Pedestrian} and {\it Cyclist}, the improvements are even more significant as our results exceed the second best method by more than 20\%.

\begin{table*}[t!]
	\caption{{\bf Object detection (top)} and {\bf orientation estimation (bottom) results on KITTI's validation set}.  Here, ort: orientation regression loss; ctx: contextual information; cls: class-specific weights in proposal generation. All methods use 2K proposals per image. VGG-16 network is used.}
    \vspace{-1mm}
	\label{tab:val-2d}
	\begin{center}
		\begin{small}
			\addtolength{\tabcolsep}{0pt}
			\vspace{-2mm}
			\begin{tabular}{c|c|ccc||c|c|c||c|c|c||c|c|c}
				\hline
				\multirow{2}{*}{Metric} & \multirow{2}{*}{Method} & \multirow{2}{*}{ort} & \multirow{2}{*}{ctx} & \multirow{2}{*}{cls}
				& \multicolumn{3}{|c||}{Cars} & \multicolumn{3}{|c||}{Pedestrians} & \multicolumn{3}{|c}{Cyclists} \\
				\cline{6-14}
				& & & & & Easy & Moderate & Hard & Easy & Moderate & Hard & Easy & Moderate & Hard \\
				\hline
				\multirow{6}{*}{AP$_{\text{2D}}$}
				& SS~\cite{van2011segmentation} & \multicolumn{3}{|c||}{\multirow{2}{*}{-}} & 75.91 & 60.00 & 50.98 & 54.06 & 47.55 & 40.56 & 56.26 &39.16&38.83  \\
				& EB~\cite{zitnick2014edge} & & & & 86.81 & 70.47 & 61.16 & 57.79 & 49.99 & 42.19 & 55.01 & 37.87 & 35.80 \\
				\cline{2-14}
				& \multirow{4}{*}{Ours}
				&            & & \checkmark & 92.18 & 87.26 & 78.58  & 72.56 & 69.08 & 61.34 & {\bf 90.69} & 62.82 & 58.26 \\
				& & \checkmark & & \checkmark & 92.67 & 87.52 & 78.78 & 72.42 & {\bf 69.42} & 61.55 & 85.92 &62.54 & 57.71 \\
				& & \checkmark & \checkmark & & 92.76 & 87.30 & 78.61 & {\bf 73.76} & 66.26 & {\bf 63.15} & 85.91 & 62.82 & 57.05 \\
				& & \checkmark & \checkmark & \checkmark & {\bf 93.08} & {\bf 88.07} & {\bf 79.39} & 71.40 & 64.46 & 60.39 & 83.82 &  {\bf 63.47} & {\bf 60.93} \\
				\hline
				\multirow{6}{*}{AOS} 
				& SS~\cite{van2011segmentation} & \multicolumn{3}{|c||}{\multirow{2}{*}{-}} & 73.91 & 58.06 & 49.14 & 44.55 &39.05& 33.15 & 39.82 & 28.20 & 28.40  \\
				& EB~\cite{zitnick2014edge} & & & & 83.91 & 67.89 & 58.34 & 46.80& 40.22& 33.81 & 43.97 & 30.36 & 28.50 \\
				\cline{2-14}
				& \multirow{4}{*}{Ours}
				&            & & \checkmark & 39.52 & 38.24 & 34.01 & 34.15 & 33.08 & 29.27 &  63.88 & 43.85 & 40.36 \\
				& & \checkmark & & \checkmark & 91.46 & {\bf 85.80} & 76.73 & {\bf 62.25} & {\bf 59.15} &{\bf 52.24} & {\bf 77.60} & {\bf 55.75} & 51.23  \\
				& & \checkmark & \checkmark & & 91.22 & 85.12 & 75.74 & 61.62 & 55.01 & 52.14 & 74.28 & 53.96 & 49.05 \\
				& & \checkmark & \checkmark & \checkmark & {\bf 91.58} & {\bf 85.80} & {\bf 76.80} & 61.57 & 54.79 & 51.12 & 73.94 & 55.59 & {\bf 53.00} \\
				\hline
			\end{tabular}
		\end{small}
\vspace{-3mm}
	\end{center}
\end{table*}

\begin{figure*}[t!]
	\begin{center}
		\begin{tabular}{ccc}
			\raisebox{-0.38\height}{\includegraphics[width=0.3\linewidth,trim = 5mm 0mm 0mm 0mm, clip]{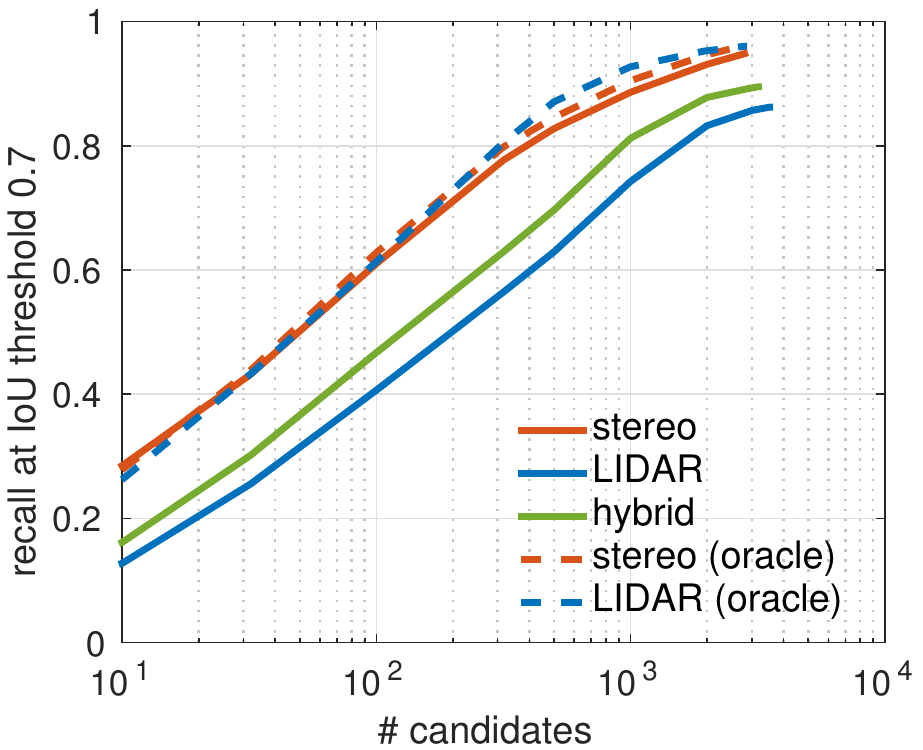}}&
			\raisebox{-0.38\height}{\includegraphics[width=0.3\linewidth,trim = 5mm 0mm 0mm 0mm, clip]{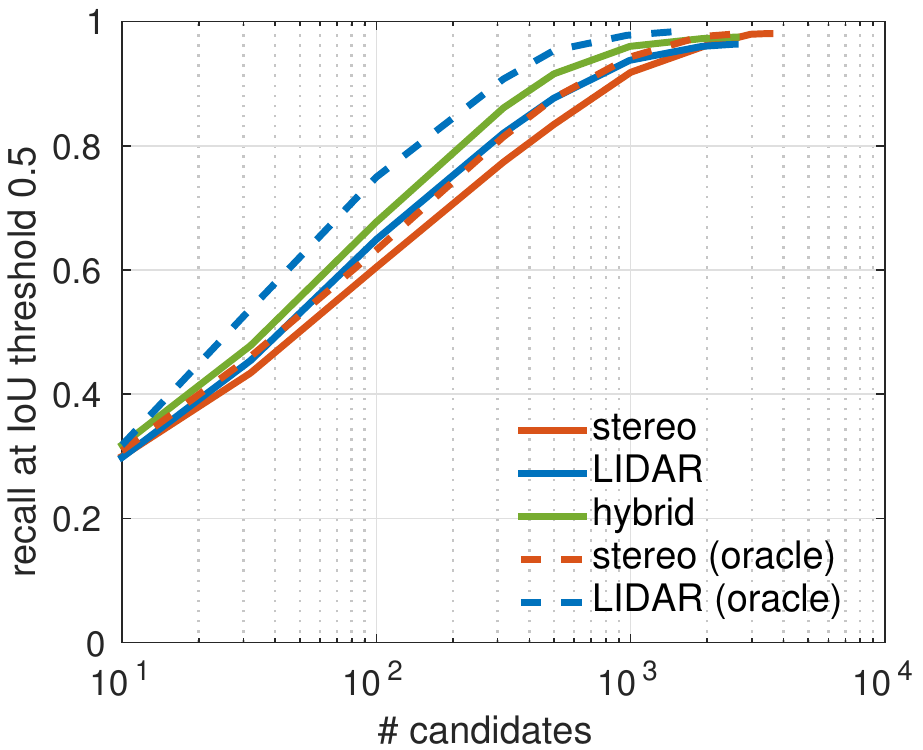}}&
			\raisebox{-0.38\height}{\includegraphics[width=0.3\linewidth,trim = 5mm 0mm 0mm 0mm, clip]{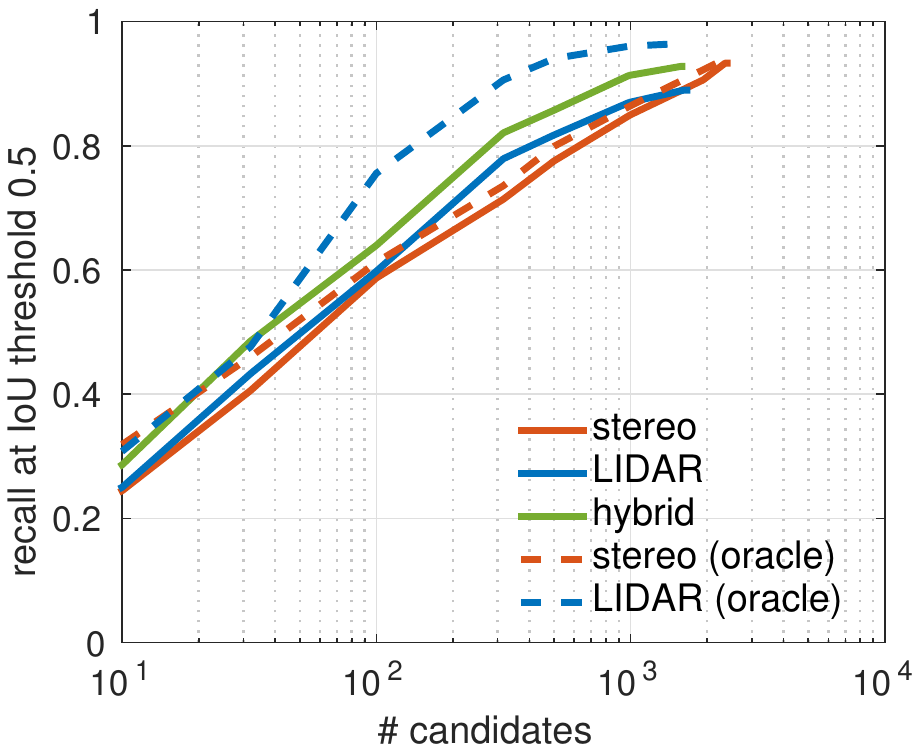}}\\
			Car & Pedestrian & Cyclist
		\end{tabular}
\vspace{-3mm}
		\caption{\textbf{Stereo vs LIDAR: 2D bounding box Recall vs number of Candidates on \emph{Moderate} data}. We use an overlap threshold of  0.7  for {\it Car}, and 0.5  for {\it Pedestrian} and {\it Cyclist}. By hybrid we mean the approach that uses both stereo and LIDAR for road plane estimation, and LIDAR for feature extraction.}
		\label{fig:2d-recall-vs-cand-lidar}
\vspace{-3mm}
	\end{center}
\end{figure*}

\begin{figure*}[t!]
	\begin{center}
		\begin{tabular}{ccc}
			\raisebox{-0.38\height}{\includegraphics[width=0.3\linewidth,trim = 5mm 0mm 0mm 0mm, clip]{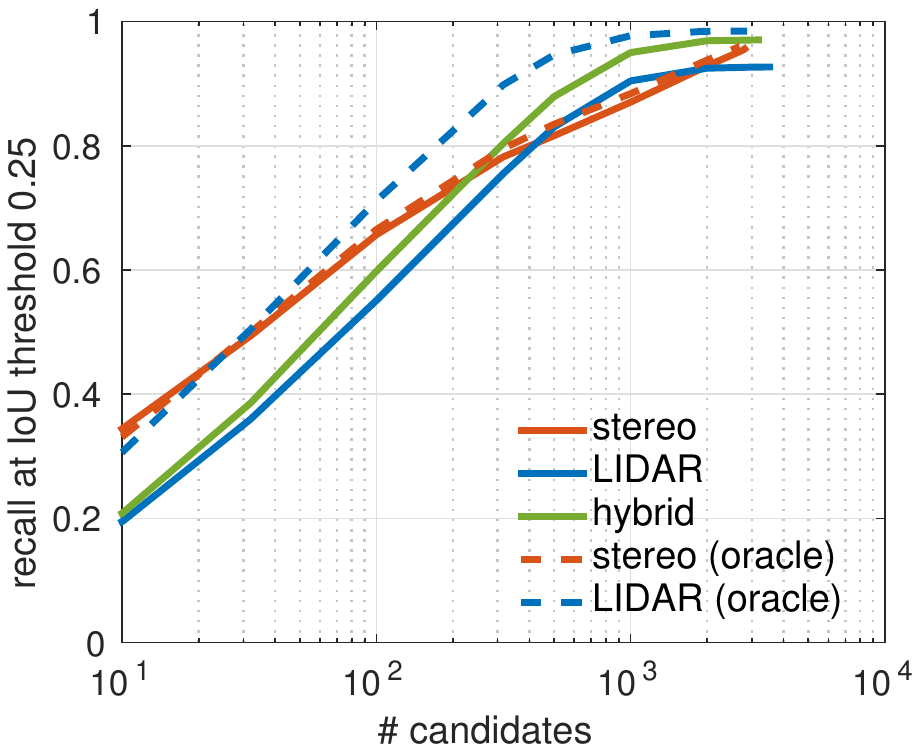}}&
			\raisebox{-0.38\height}{\includegraphics[width=0.3\linewidth,trim = 5mm 0mm 0mm 0mm, clip]{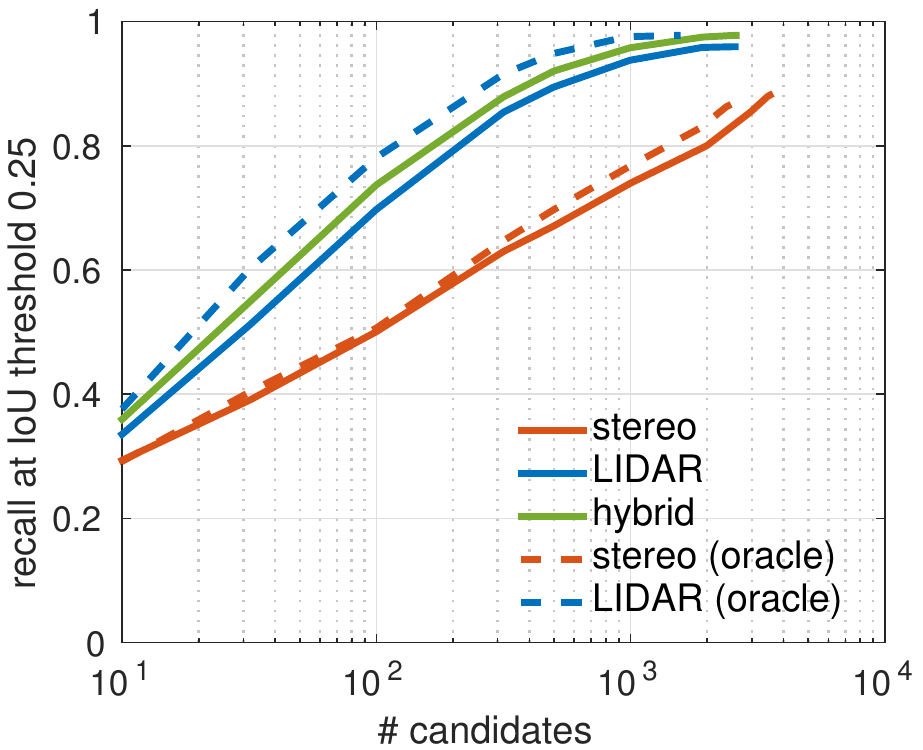}}&
			\raisebox{-0.38\height}{\includegraphics[width=0.3\linewidth,trim = 5mm 0mm 0mm 0mm, clip]{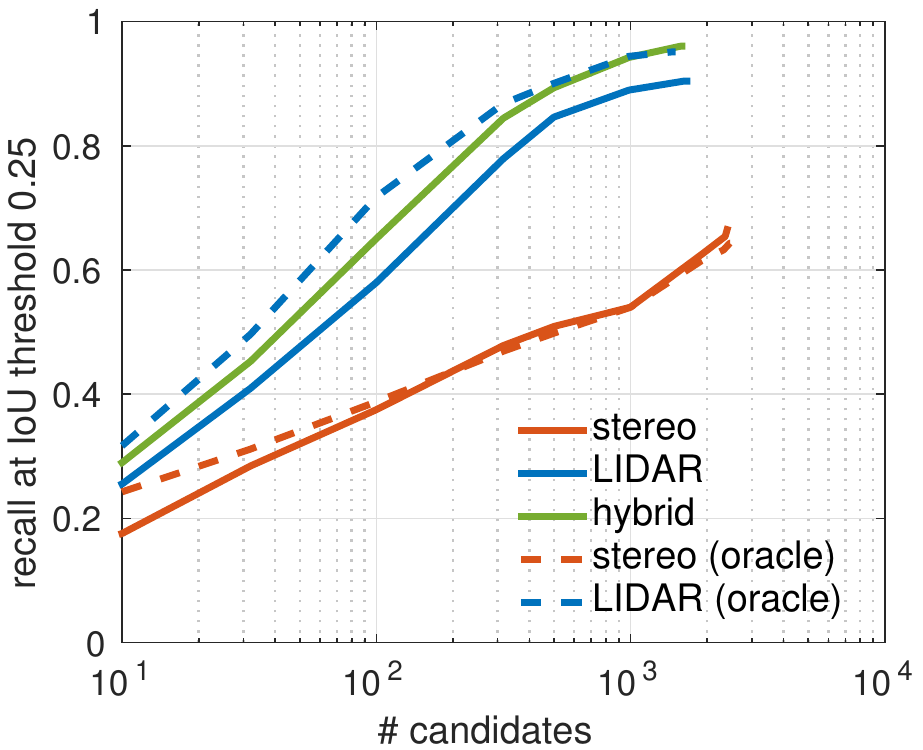}}\\
			Car & Pedestrian & Cyclist
		\end{tabular}
\vspace{-3mm}
		\caption{\textbf{Stereo vs LIDAR: 3D bbox Recall vs number of Candidates on \emph{Moderate} data}. IoU threshold is set to 0.25.}
		\label{fig:3d-recall-25-lidar}
\vspace{-3mm}
	\end{center}
\end{figure*}

\vspace{-1mm}
\paragraph{Comparison with Baselines}
We also apply our detection network on two state-of-the-art bottom-up proposals, Selective Search (SS)~\cite{van2011segmentation} and EdgeBoxes (EB)~\cite{zitnick2014edge}. We conduct the experiments on the validation set. 2K proposals per image are used for all methods. As shown in Table~\ref{tab:val-2d}, our approach outperforms Selective Search and EdgeBoxes by around 20\% in terms of 2D detection AP and orientation estimation score AOS. We also report AP$_{\text{2D}}$ as a function of the proposal budget for {\it Car} on \emph{Moderate} data, in Fig.~\ref{fig:ap_proposals}. When using only 100 proposals, our approach already achieves 70\% AP, while EdgeBoxes and Selective Search obtain only 25.9\% and 16.5\%, respectively. Furthermore, EdgeBoxes reaches its best performance (78.7\%) with 5000 proposals, while our approach needs only 200 proposals to achieve a similar AP.

\vspace{-1mm}
\subsection{3D Object Detection Performance}
As the KITTI object detection benchmark does not evaluate 3D object detection on the (held-out) test set, we evaluate 3D object detection performance on the validation set.
We use 1m and 2m as the threshold when evaluating Average Localization Precision (ALP). 
We report the results of our basic model with VGG-16 network for {\it Car} in Table~\ref{tab:stereo-vs-lidar-3d}.
Results of RGB-HHA models are presented in Table~\ref{tab:depth-3d-ap}.
We obtain the highest AP$_{\text{3D}}$ and ALP with a two-stream RGB-HHA model using proposals generated with hybrid data, i.e., combining stereo and LIDAR.
In particular, we achieve 81.21\% AP$_{\text{3D}}$ and 75.44\%/88.83\% ALP within threshold of 1m/2m in the \emph{Moderate} regime for 3D Car detection.

\vspace{-1mm}
\subsection{Stereo vs LIDAR}
We also apply our proposal method to LIDAR data. Compared with stereo, LIDAR point cloud has more precise depth estimation while being very sparse. Since road plane estimation is less accurate when using only sparse LIDAR point cloud, we also experiment with a \emph{hybrid} approach, which uses dense stereo point cloud to compute superpixel features for road region classification, and LIDAR point cloud to fit a ground plane and to compute energy potentials for inference in our model. 

\begin{table*}[t!]
	\caption{{\bf Stereo vs LIDAR on 2D Object Detection and Orientation Estimation:} AP$_{\text{2D}}$ and AOS for \emph{Car} on validation set of KITTI. VGG-16 network is used.}
	\label{tab:stereo-vs-lidar-2d}
	\begin{center}
		\begin{small}
			\addtolength{\tabcolsep}{1pt}
			\vspace{-4mm}
			\begin{tabular}{c||c|c|c||c|c|c}
				\hline
				\multirow{2}{*}{Data} & \multicolumn{3}{|c||}{AP$_{\text{2D}}$} & \multicolumn{3}{|c}{AOS} \\
				\cline{2-7}
				& Easy & Moderate & Hard & Easy & Moderate & Hard \\
				\hline
				stereo (oracle) & 92.73 & 88.30 & 79.48 & 90.98 & 86.08 & 76.90 \\
				LIDAR (oracle) & {\bf 93.21} & {\bf 88.77} & {\bf 79.70} & {\bf 91.67} & {\bf 86.61} & {\bf 77.22} \\
				\hline
				stereo & {\bf 93.08} & {\bf 88.07} & {\bf 79.39} & {\bf 91.58} & {\bf 85.80} & {\bf 76.80} \\
				LIDAR & 87.78 & 79.51 & 70.74 & 85.90 & 77.24 & 68.23 \\
				hybrid  & 92.17 & 86.52 & 78.37 & 90.62 & 84.44 & 75.91 \\
				\hline
			\end{tabular}
		\end{small}
		\vspace{-2mm}
	\end{center}
\end{table*}

\begin{table*}[t!]
	\caption{{\bf  Stereo vs LIDAR on 3D Object Detection:} AP$_{\text{3D}}$ and ALP for \emph{Car} on KITTI validation. VGG-16 network is used.}
	\label{tab:stereo-vs-lidar-3d}
	\begin{center}
		\begin{small}
			\addtolength{\tabcolsep}{1pt}
			\vspace{-4mm}
			\begin{tabular}{c||c|c|c||c|c|c||c|c|c}
				\hline
				\multirow{2}{*}{Data} & \multicolumn{3}{|c||}{AP$_{\text{3D}}$} & \multicolumn{3}{|c||}{ALP ($<$ 1m)} & \multicolumn{3}{|c}{ALP ($<$ 2m)} \\
				\cline{2-10}
				& Easy & Moderate & Hard & Easy & Moderate & Hard & Easy & Moderate & Hard \\
				\hline
				stereo (oracle) & {\bf 87.33}  & 70.46 & 63.82 & {\bf 75.98}  & 58.01  & 52.37 & {\bf 94.12} & 78.64 & 73.95 \\
				LIDAR (oracle) & 84.63  & {\bf 82.04} & {\bf 74.92} & 75.32  & {\bf 74.54}  & {\bf 68.50} & 91.88 & {\bf 89.29} & {\bf 84.06} \\
				\hline
				stereo & {\bf 88.45} & 69.52 & 62.65 & {\bf 77.90} & 58.09  & 52.17 & {\bf 94.39} & 77.57 & 72.69 \\
				LIDAR & 80.73  & 73.56 & 66.83 & 72.26  & 67.77  & 62.42 & 87.98 & 81.07 & 76.52 \\
				hybrid & 86.47  & {\bf 80.56} & {\bf 73.71} & 77.19  & {\bf 73.85}  & {\bf 67.99} & 93.00 & {\bf 87.52} & {\bf 82.51} \\
				\hline
			\end{tabular}
		\end{small}
		\vspace{-2mm}
	\end{center}
\end{table*}

\begin{table*}[t!]
\caption{{\bf Comparison of different architectures on 2D object detection and orientation estimation:} AP$_{\text{2D}}$ and AOS for {\it Car} on validation set of KITTI. 7-layer VGG\_CNN\_M\_1024~\cite{chatfield2014return} network is used.}
\label{tab:depth-2d-ap}
\begin{center}
\begin{small}
\vspace{-4mm}
\addtolength{\tabcolsep}{2pt}
\begin{tabular}{c|c||c|c|c||c|c|c}
\hline
\multirow{2}{*}{Data} & \multirow{2}{*}{Approach} & \multicolumn{3}{|c||}{AP$_{\text{2D}}$} & \multicolumn{3}{|c}{AOS} \\
\cline{3-8}
 & & Easy & Moderate & Hard & Easy & Moderate & Hard \\
\hline
\multirow{3}{*}{stereo} 
& RGB & 92.56 & 87.27 & 78.38 & 90.24 & 83.98 & 74.69 \\
& RGB-HHA, one-stream & 90.81 & 87.41 & 78.82 & {\bf 90.80} & {\bf 84.28} & {\bf 75.39} \\
& RGB-HHA, two-stream & {\bf 93.03} & {\bf 87.97} & {\bf 78.98} & 90.34 & 84.27 & 74.90 \\
\hline
\multirow{3}{*}{LIDAR}
& RGB & 87.12 & 78.64 & 69.85 & 84.30 & 75.08 & 66.13 \\
& RGB-HHA, one-stream & 87.42 & 78.98 & 70.11 & 84.88 & 75.51 & 66.53 \\
& RGB-HHA, two-stream & {\bf 88.04} & {\bf 79.39} & {\bf 70.48} & {\bf 85.05} & {\bf 75.70} & {\bf 66.63} \\
\hline
\multirow{3}{*}{hybrid} 
& RGB & 90.99 & 84.40 & 76.02 & 87.93 & 80.47 & 71.76 \\
& RGB-HHA, one-stream & 90.81 & 84.40 & {\bf 77.12} & 88.45 & {\bf 80.98} & {\bf 73.20} \\
& RGB-HHA, two-stream & {\bf 92.69} & {\bf 84.78} & 76.43 & {\bf 89.23} & 80.53 & 71.73 \\
\hline
\end{tabular}
\end{small}
\vspace{-2mm}
\end{center}
\end{table*}

\begin{table*}[t!]
\caption{{\bf Comparison of different architectures on 3D object detection:} AP$_{\text{3D}}$ and ALP for {\it Car} on validation set of KITTI. 7-layer VGG\_CNN\_M\_1024~\cite{chatfield2014return} network is used.}
\label{tab:depth-3d-ap}
\begin{center}
\begin{small}
\vspace{-4mm}
\addtolength{\tabcolsep}{0pt}
\begin{tabular}{c|c||c|c|c||c|c|c||c|c|c}
\hline
\multirow{2}{*}{Data} & \multirow{2}{*}{Approach} & \multicolumn{3}{|c||}{AP$_{\text{3D}}$} & \multicolumn{3}{|c||}{ALP ($<$ 1m)} & \multicolumn{3}{|c}{ALP ($<$ 2m)} \\
\cline{3-11}
& & Easy & Moderate & Hard & Easy & Moderate & Hard & Easy & Moderate & Hard \\
\hline
\multirow{3}{*}{stereo}
& RGB & 77.50  & 56.97 & 50.84 & 64.89  & 47.34  & 42.20 & 89.02 & 69.27 & 64.74 \\
& RGB-HHA, one-stream & 89.22  & 68.19 & {\bf 62.24} & 81.00  & 58.53  & {\bf 52.76} & 95.23 & 76.57 & 72.50 \\
& RGB-HHA, two-stream & {\bf 90.43} & {\bf 68.90} & 62.22 & {\bf 82.25}  & {\bf 58.99} & 52.71 & {\bf 95.74} & {\bf 77.66} & {\bf 73.03} \\
\hline
\multirow{3}{*}{LIDAR}
& RGB & 76.51  & 68.77 & 62.10 & 66.46  & 62.14  & 57.53 & 86.22 & 78.67 & 74.44 \\
& RGB-HHA, one-stream & {\bf 84.83} & 73.92 & 67.55 & 76.61  & 68.52  & 63.25 & {\bf 92.16} & 82.56 & 77.95 \\
& RGB-HHA, two-stream & 84.02  & {\bf 74.11} & {\bf 67.62} & {\bf 77.40} & {\bf 69.62}  & {\bf 63.92} & 91.32 & {\bf 82.72} & {\bf 78.13} \\
\hline
\multirow{3}{*}{hybrid}
& RGB & 81.83  & 75.86 & 68.66 & 71.12  & 68.46  & 62.89 & 90.59 & 85.10 & 79.86 \\
& RGB-HHA, one-stream & 87.86  & 79.61 & 72.86 & 79.54  & 74.06  & 68.58 & 94.88 & 87.94 & 83.16 \\
& RGB-HHA, two-stream & {\bf 89.49}  & {\bf 81.21} & {\bf 74.32} & {\bf 82.16}  & {\bf 75.44}  & {\bf 69.27} & {\bf 95.46} & {\bf 88.83} & {\bf 83.75} \\
\hline
\end{tabular}
\end{small}
\vspace{-2mm}
\end{center}
\end{table*}

\vspace{-1mm}
\paragraph{Proposal Recall:}
As shown in Fig.~\ref{fig:3d-recall-25-lidar}, using LIDAR point cloud significantly boosts the 3D bounding box recall, especially for small objects, i.e., pedestrians, cyclists and distant objects. We obtain the highest 3D recall with the hybrid approach, which combines stereo and LIDAR for road estimation. In terms of 2D bounding box recall shown in Fig.~\ref{fig:2d-recall-vs-cand-lidar}, stereo works slightly better.

\vspace{-1mm}
\paragraph{Detection Performance:}
We report 2D/3D detection performance in Table~\ref{tab:stereo-vs-lidar-2d} and Table~\ref{tab:stereo-vs-lidar-3d} for comparison of stereo and LIDAR approaches. In terms of 2D detection and orientation estimation, we obtain the best performance with stereo.
For 3D detection, we achieve significantly higher 3D AP and ALP using LIDAR data under the \emph{Moderate} and \emph{Hard} regimes. However, stereo still works better in the \emph{Easy} regime. This demonstrates the advantage of LIDAR point cloud in detecting small, occluded, and distant objects owing to its precise depth information, while dense point cloud from stereo works better for objects at shorter distances.  When using the hybrid approach, we get the highest 3D accuracy in the \emph{Moderate} and \emph{Hard} setting.

\vspace{-1mm}
\subsection{Ablation Studies}

\vspace{-1mm}
\paragraph{Network Design:}
We study the effect of orientation regression and the contextual branch in our detection network. As shown in Table~\ref{tab:val-2d}, by jointly doing bounding box and orientation regression we get a large boost in performance in terms of the AOS, while the 2D detection AP remains very similar. By adding the contextual branch, we achieve the highest AP$_{\text{2D}}$ and AOS on {\it Car}. We also observe different effect on {\it Pedestrian} and {\it Cyclist} in some cases, which is probably due to the fact that the contextual branch nearly doubles the size of the model and makes it more difficult to train with very few pedestrian and cyclist instances.

\vspace{-1mm}
\paragraph{RGB-D Networks:}
We study the effect of depth features in CNN scoring model on 2D detection, orientation estimation, and 3D detection.
Comparisons of different approaches are shown in Table~\ref{tab:depth-2d-ap} and Table~\ref{tab:depth-3d-ap}.
As the RGB-HHA models require more GPU memory, we use 7-layer VGG\_CNN\_M\_1024 network~\cite{chatfield2014return} for this experiment.

Overall, RGB-HHA models achieve improvement over the RGB model for both 2D and 3D detection.
The two-stream RGB-HHA model also gains in performance over the one-stream RGB-HHA model in most cases.
The improvement is significant for 3D detection while marginal ($\sim$0.5\%) for 2D detection.
In particular, the two-stream RGB-HHA model outperforms the RGB model by about 10\% and 5\% for the stereo and LIDAR/hybrid approaches respectively for 3D detection.
Note that for 3D car detection, we obtain the highest accuracy using a 7-layer two-stream RGB-HHA model with hybrid data, which even outperforms the 16-layer RGB model (see Table~\ref{tab:stereo-vs-lidar-3d}).
This suggests the importance of depth information for the 3D object detection task.

\vspace{-1mm}
\paragraph{Ground Plane:}
To study the effect of ground plane estimation on detection performance, we allow access to ``oracle" ground planes.
We approximate the oracle ground plane by fitting a plane to the footprints of ground truth 3D bounding boxes.
We show results for both, the stereo and LIDAR approaches on {\it Car} detection.
Proposal recall plots are shown in Fig.~\ref{fig:2d-recall-vs-cand-lidar} and Fig.~\ref{fig:3d-recall-25-lidar}.
2D/3D detection results are shown in Table~\ref{tab:stereo-vs-lidar-2d} and Table~\ref{tab:stereo-vs-lidar-3d}.
Giving access to oracle ground plane significantly boosts the 2D and 3D box recall of the LIDAR approach.
Similarly,  AP$_{\text{2D}}$/AOS improve by about 9\%, AP$_{\text{3D}}$/ALP by about 8\% for the LIDAR approach, while the improvement on stereo is marginal.
This suggests that exploiting a better approach for ground plane estimation using sparse LIDAR point clouds would further improve the performance of the LIDAR approach.


\begin{figure*}[t!]
	\vspace{-0mm}
	\begin{center}
		\addtolength{\tabcolsep}{-5pt}
		\begin{tabular}{p{6.5mm}ccc}
			\hspace{-0.7mm}\rotatebox{90}{\hspace{4mm}\footnotesize Images} &\hspace{-0.5cm}
			\raisebox{-0.0\height}{\includegraphics[width=0.315\linewidth,trim = 5mm 0mm 0mm 0mm, clip]{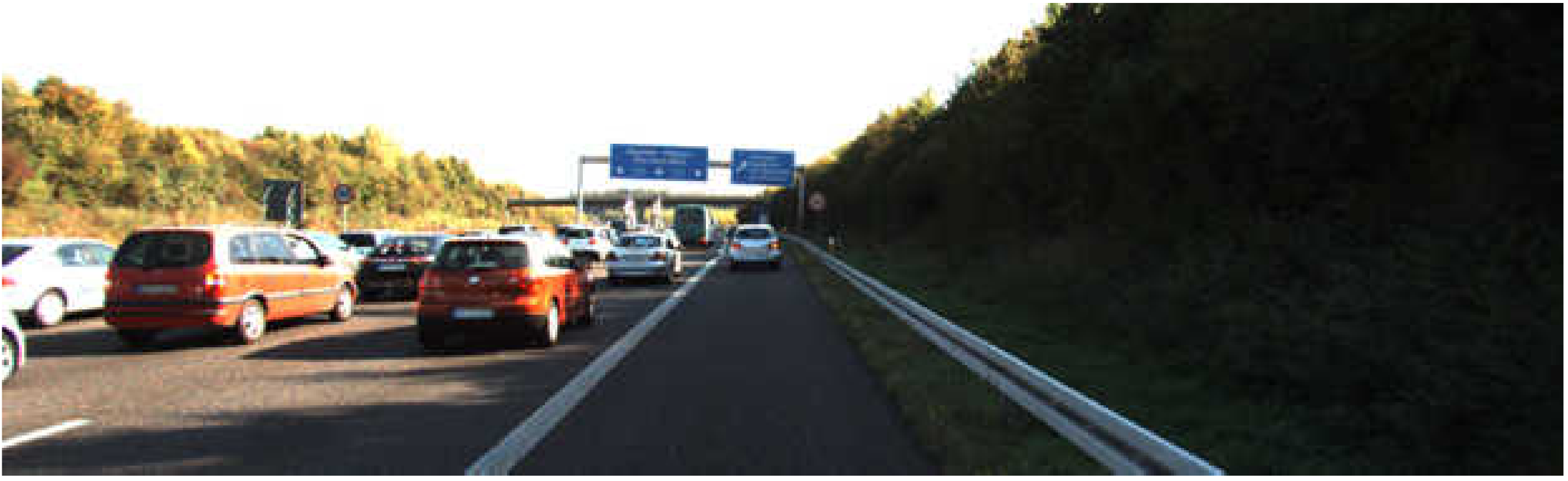}}&
			\raisebox{-0.0\height}{\includegraphics[width=0.315\linewidth,trim = 5mm 0mm 0mm 0mm, clip]{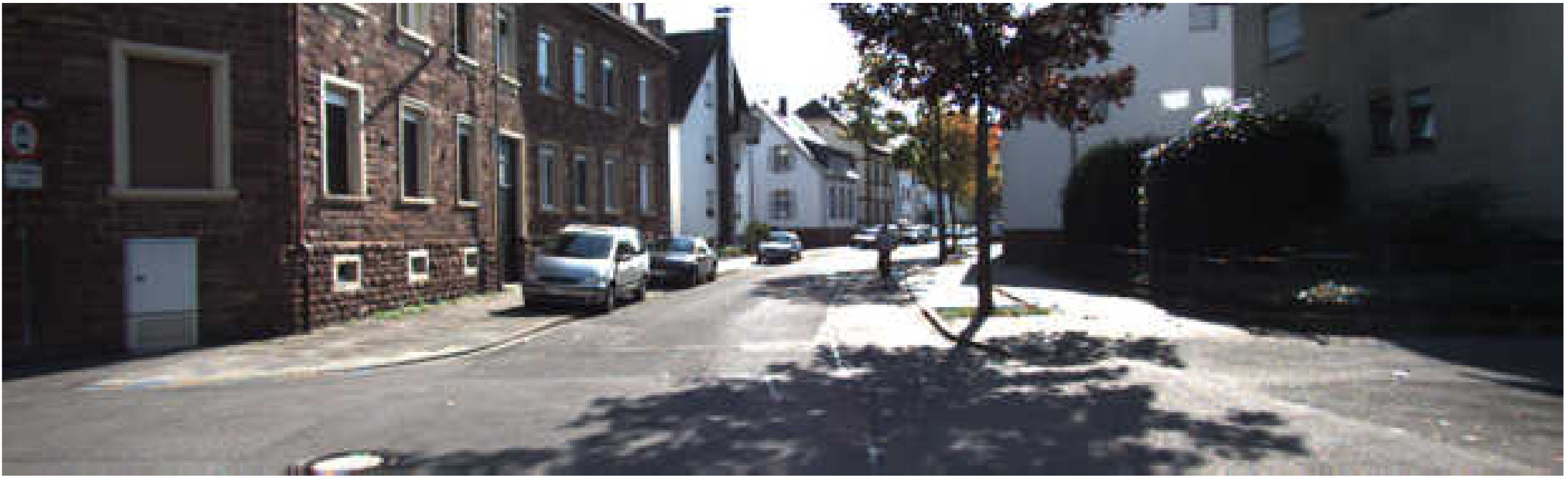}} &
			\raisebox{-0.0\height}{\includegraphics[width=0.315\linewidth,trim = 5mm 0mm 0mm 0mm, clip]{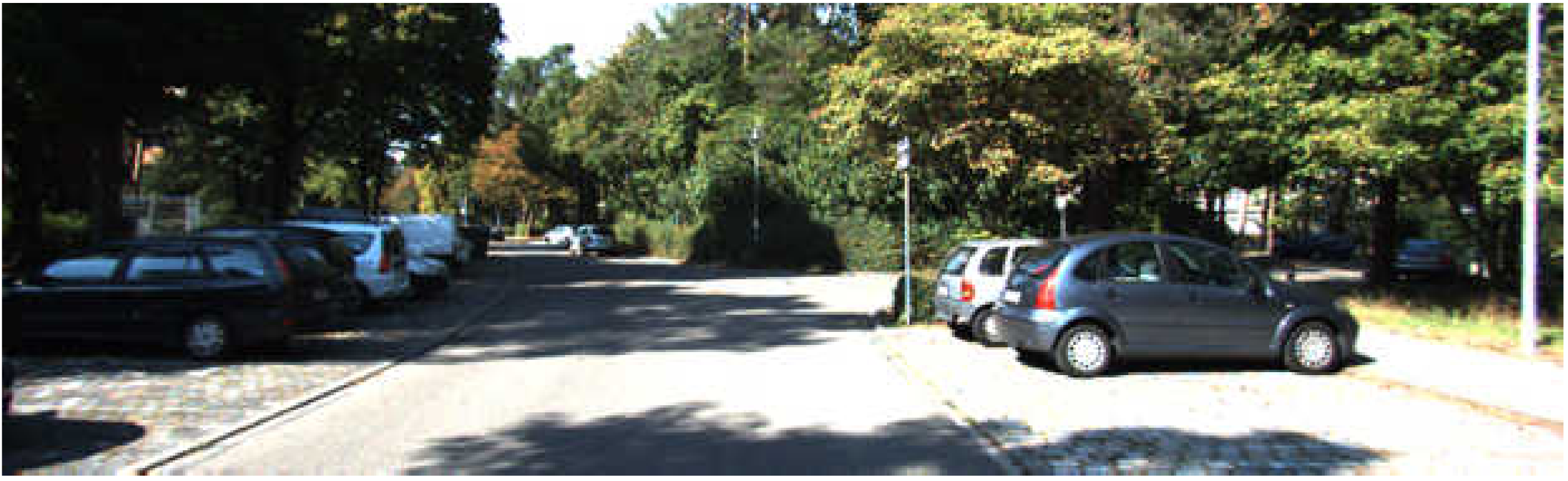}} \\[-0.6mm]
			\hspace{-0.7mm}\rotatebox{90}{\hspace{1mm}\footnotesize Top 100 prop.} &\hspace{-0.5cm}
			\raisebox{-0.0\height}{\includegraphics[width=0.315\linewidth,trim = 5mm 8mm 0mm 0mm, clip]{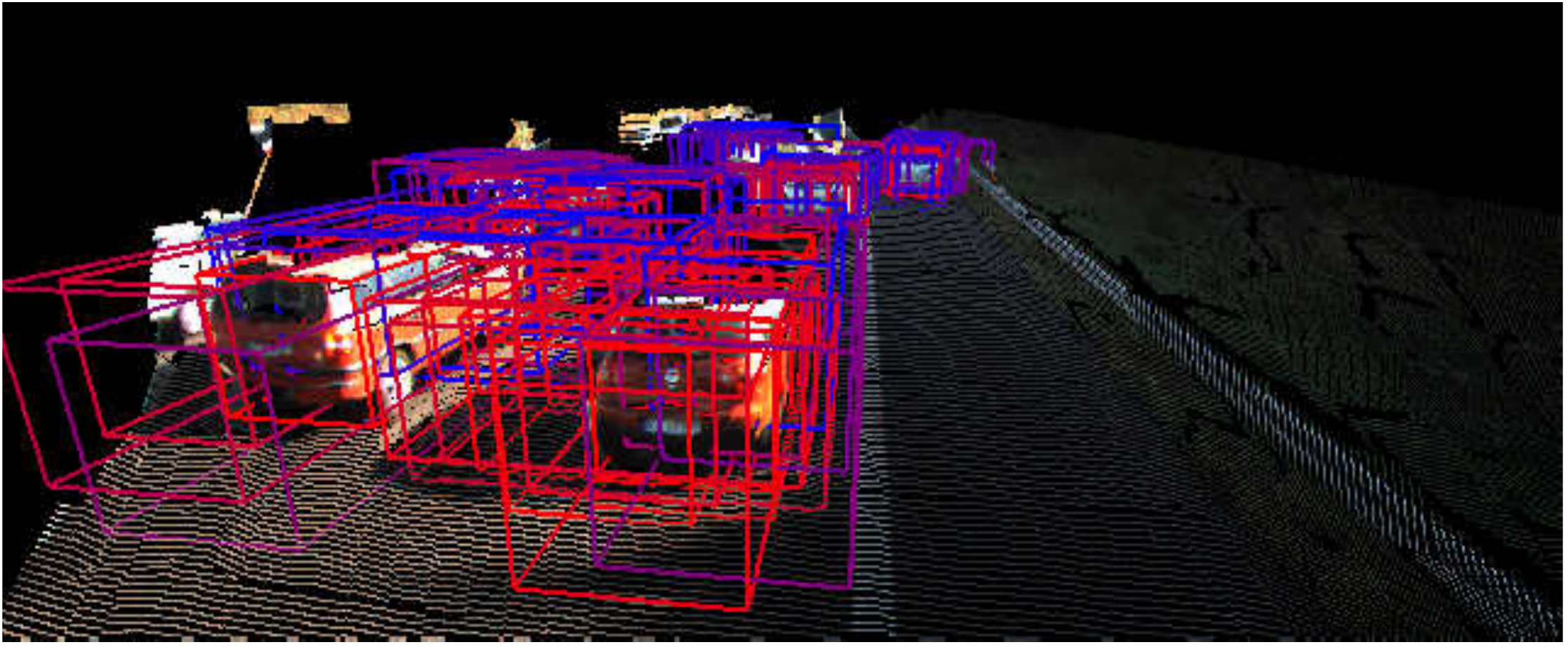}}&
			\raisebox{-0.0\height}{\includegraphics[width=0.315\linewidth,trim = 5mm 8mm 0mm 0mm, clip]{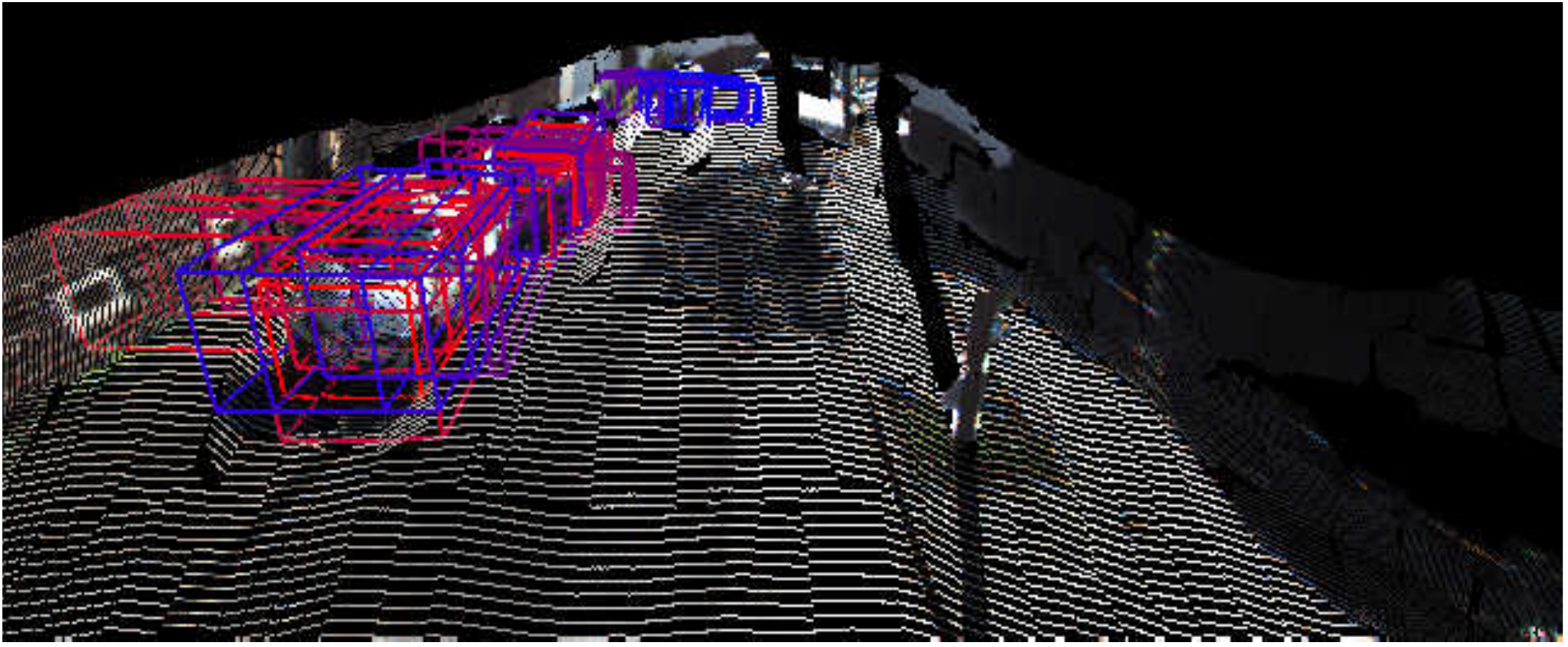}} &
			\raisebox{-0.0\height}{\includegraphics[width=0.315\linewidth,trim = 5mm 8mm 0mm 00mm, clip]{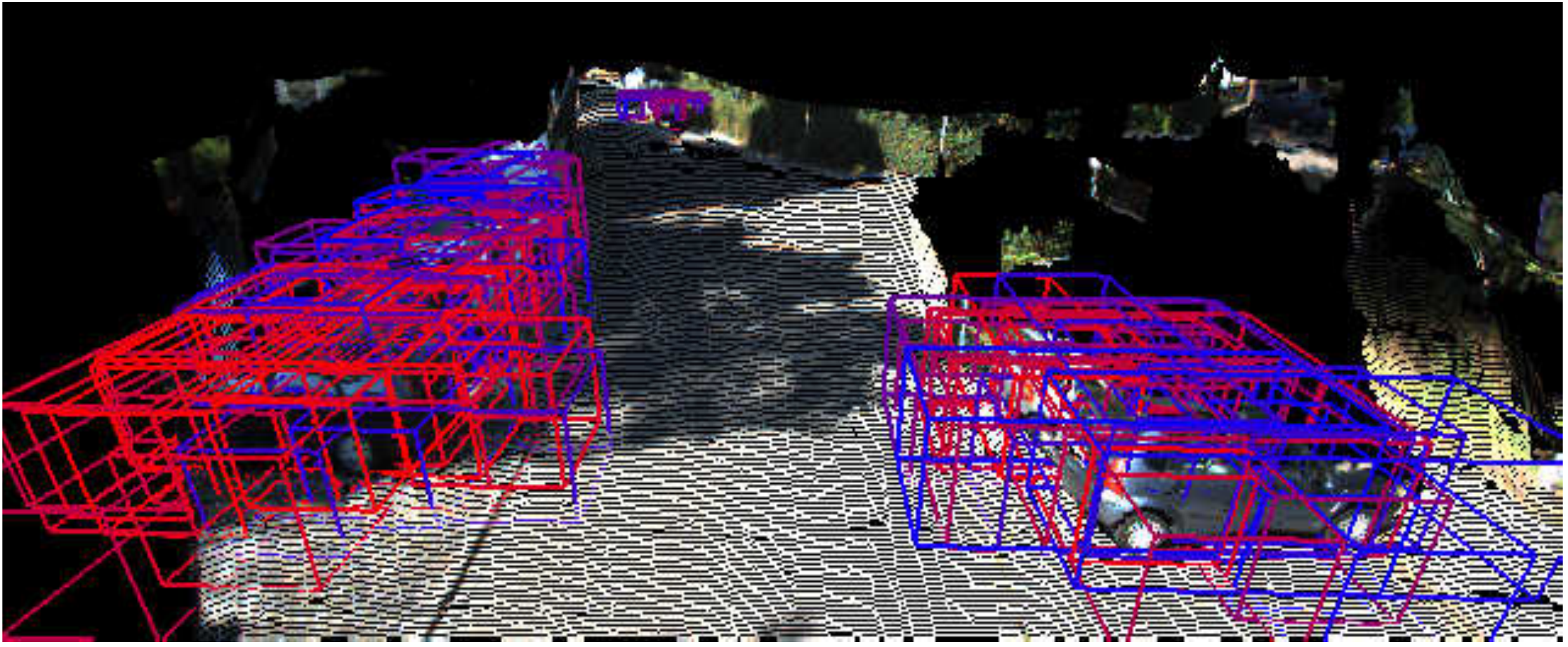}} \\[-0.6mm]
			\hspace{-0.7mm}\rotatebox{90}{\hspace{1.5mm}\footnotesize Ground truth} &\hspace{-0.5cm}
			\raisebox{-0.0\height}{\includegraphics[width=0.315\linewidth,trim = 5mm 8mm 0mm 0mm, clip]{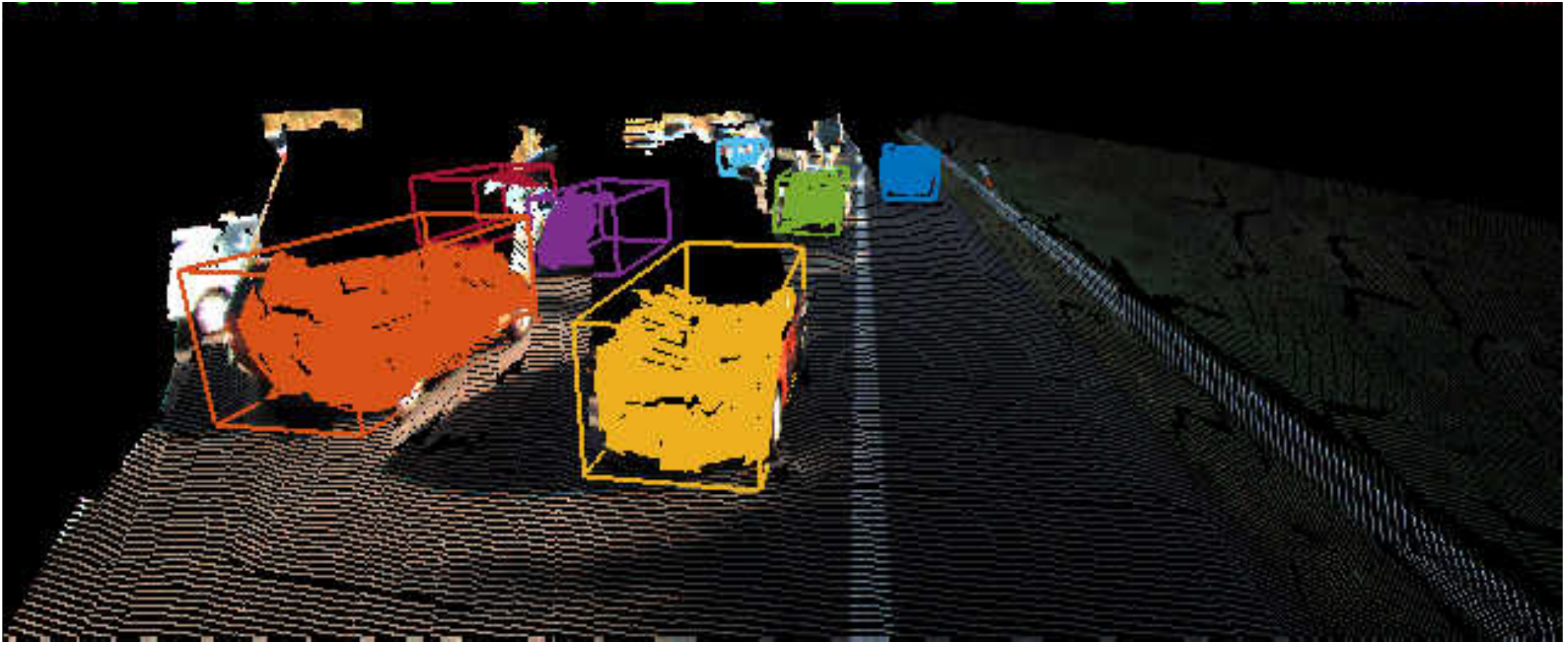}}&
			\raisebox{-0.0\height}{\includegraphics[width=0.315\linewidth,trim = 5mm 8mm 0mm 0mm, clip]{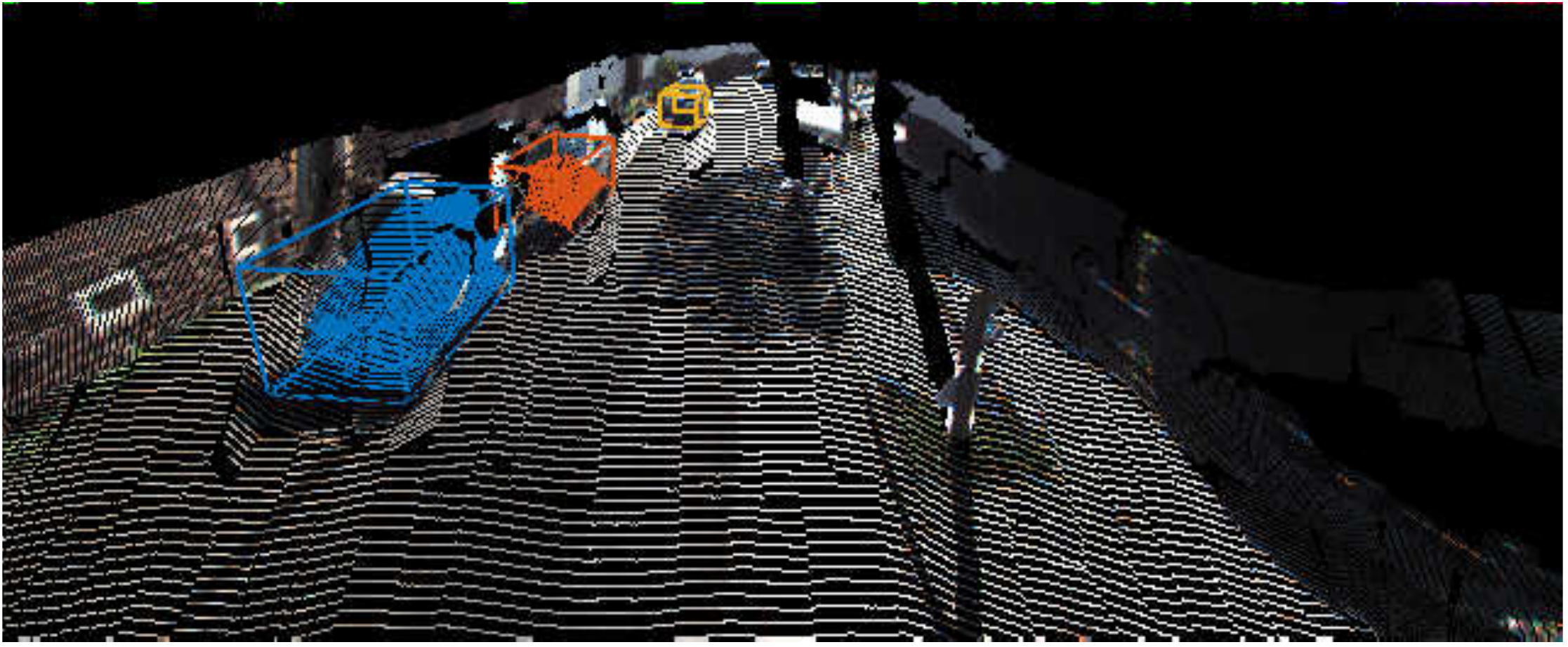}} &
			\raisebox{-0.0\height}{\includegraphics[width=0.315\linewidth,trim = 5mm 8mm 0mm 0mm, clip]{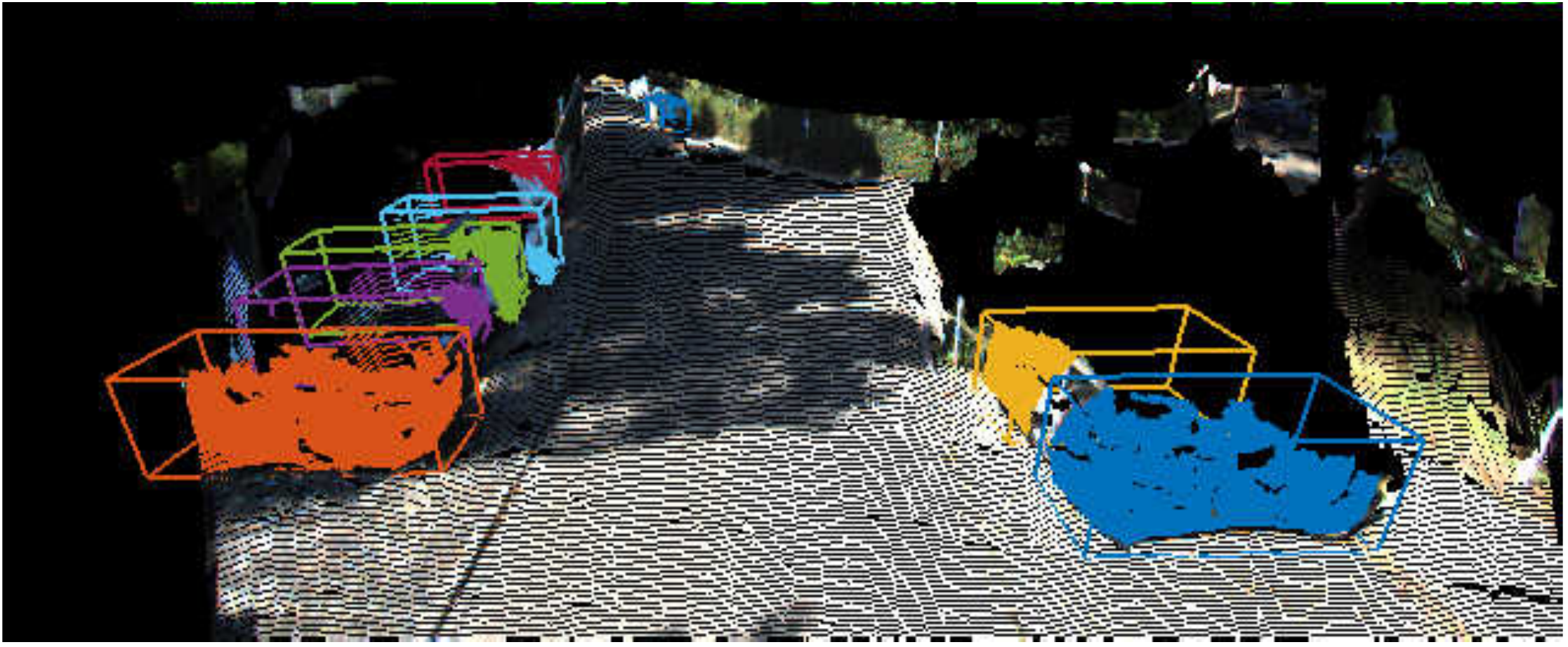}} \\[-0.6mm]
			\hspace{-0.7mm}\rotatebox{90}{\hspace{4mm}\footnotesize Best prop.} &\hspace{-0.5cm}
			\raisebox{-0.0\height}{\includegraphics[width=0.315\linewidth,trim = 5mm 8mm 0mm 0mm, clip]{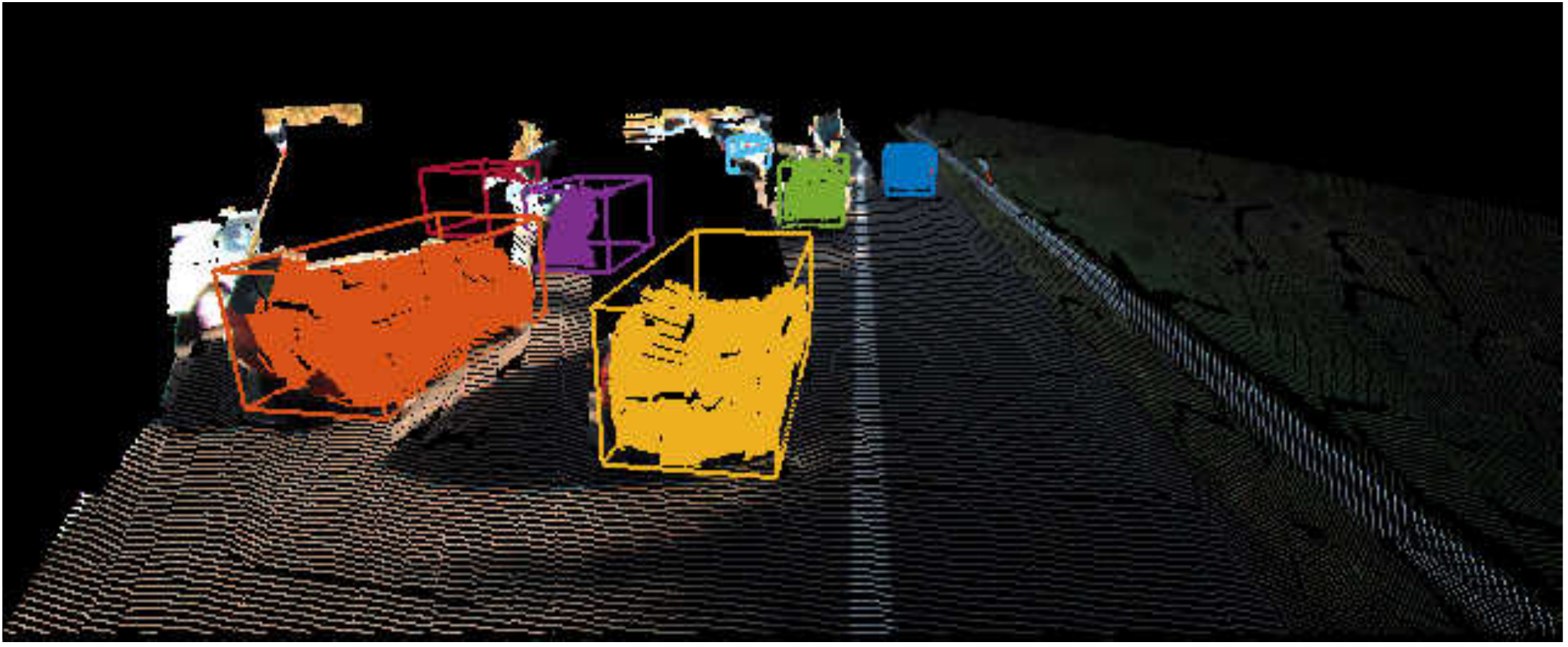}}&
			\raisebox{-0.0\height}{\includegraphics[width=0.315\linewidth,trim = 5mm 8mm 0mm 0mm, clip]{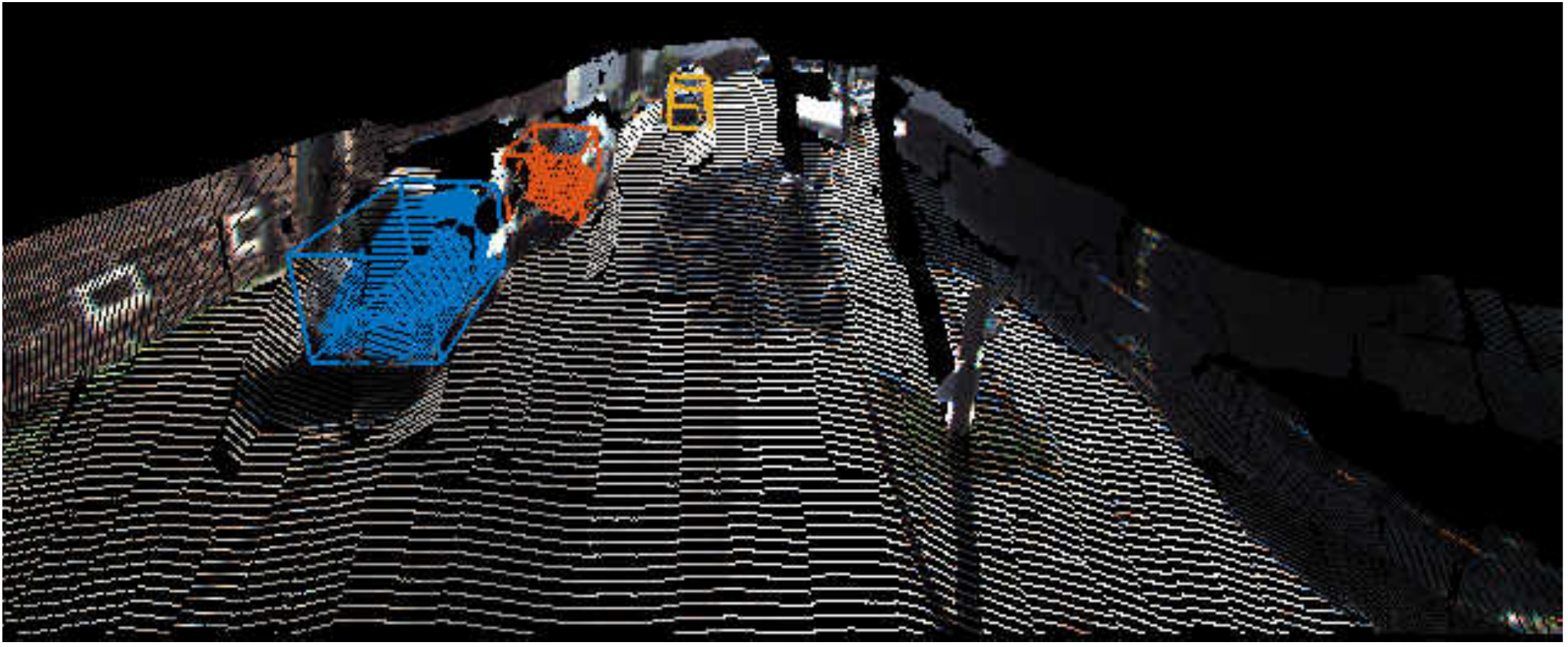}} &
			\raisebox{-0.0\height}{\includegraphics[width=0.315\linewidth,trim = 5mm 8mm 0mm 0mm, clip]{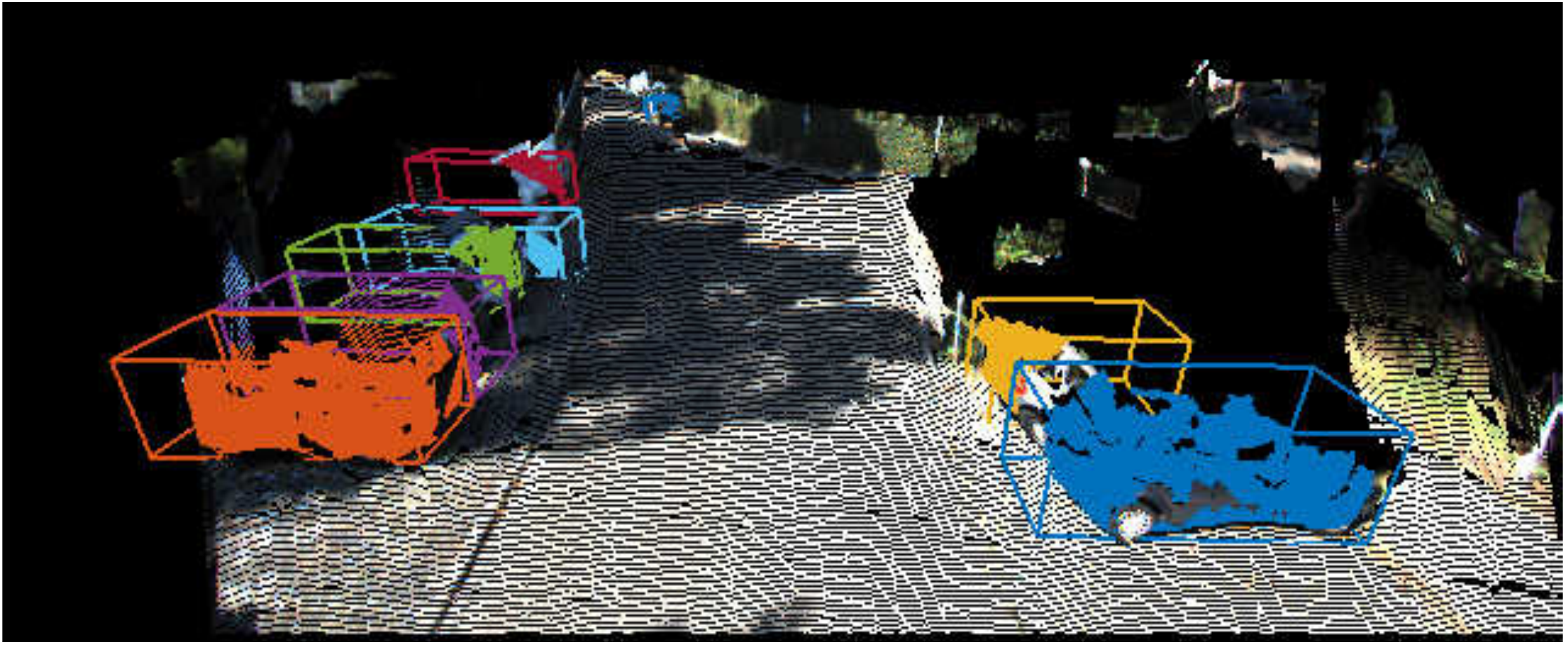}} \\[-0.6mm]
		\end{tabular}
		\vspace{-3.5mm}
		\caption{\small Qualitative results for the {\it Car} class. We show the original image, 100 top scoring proposals, ground-truth 3D boxes, and our best set of proposals that cover the ground-truth.}
		\label{fig:pcl-vis}
	\end{center}
	\vspace{-2mm}
\end{figure*}

\begin{figure*}[t!]
	\vspace{-3mm}
	\begin{center}
		\addtolength{\tabcolsep}{-5pt}
		\begin{tabular}{p{6.5mm}ccc}
			\hspace{-0.7mm}\rotatebox{90}{\hspace{4mm}\footnotesize Images} &\hspace{-0.5cm}
			\raisebox{-0.0\height}{\includegraphics[width=0.315\linewidth,trim = 5mm 0mm 0mm 0mm, clip]{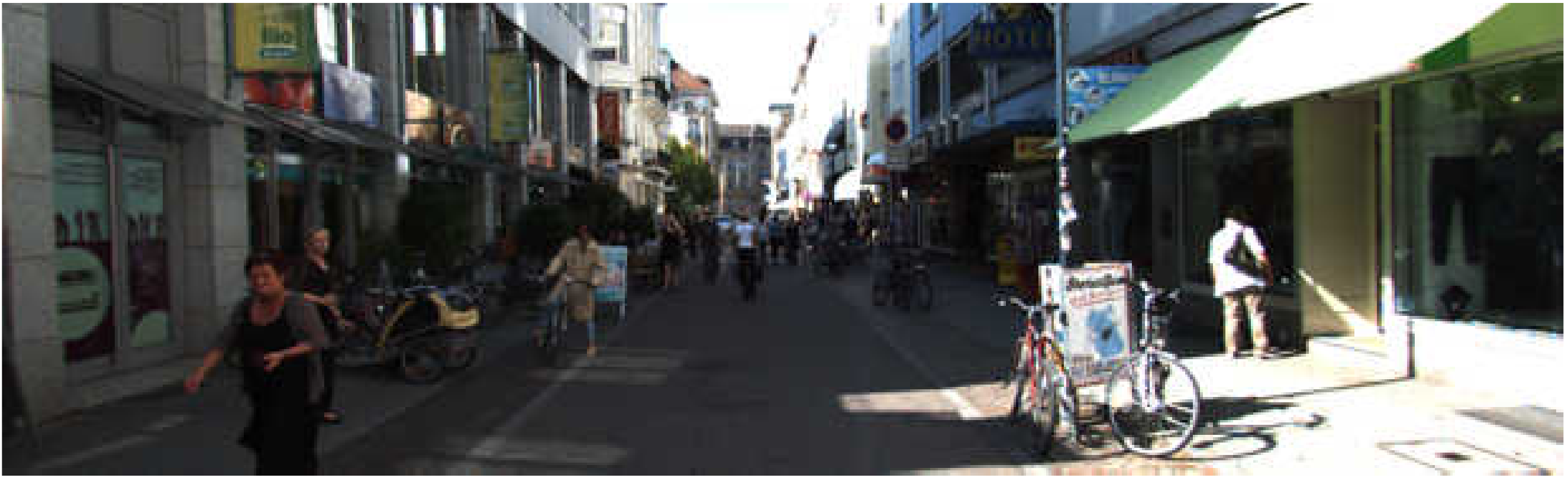}}&
			\raisebox{-0.0\height}{\includegraphics[width=0.315\linewidth,trim = 5mm 0mm 0mm 0mm, clip]{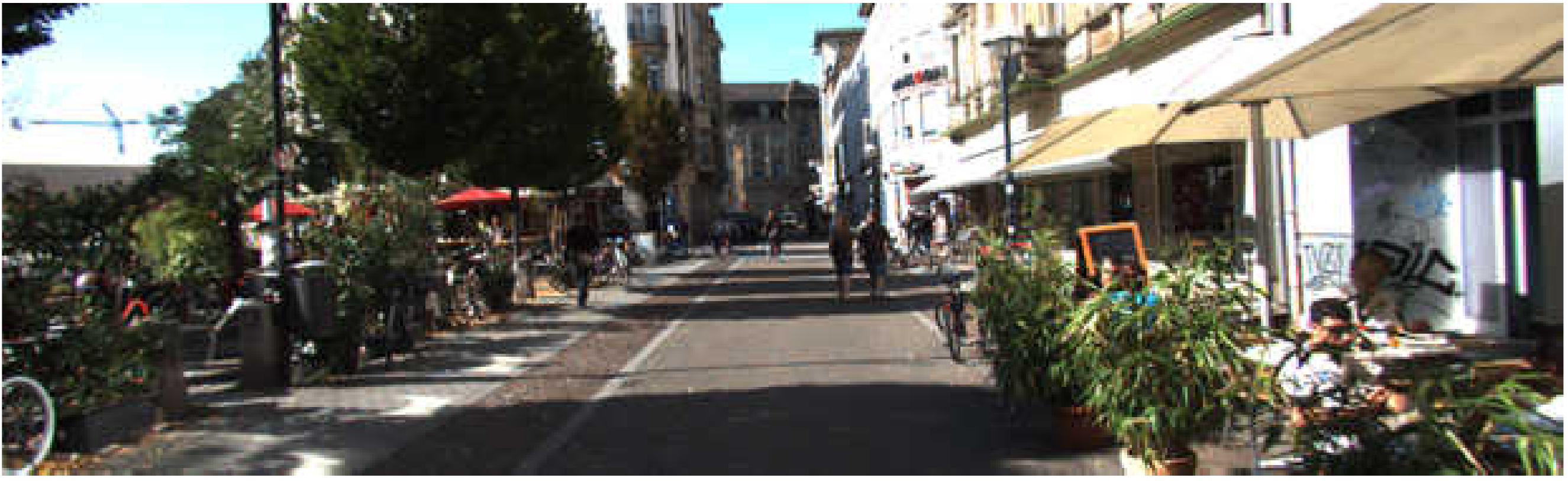}} &
			\raisebox{-0.0\height}{\includegraphics[width=0.315\linewidth,trim = 5mm 0mm 0mm 0mm, clip]{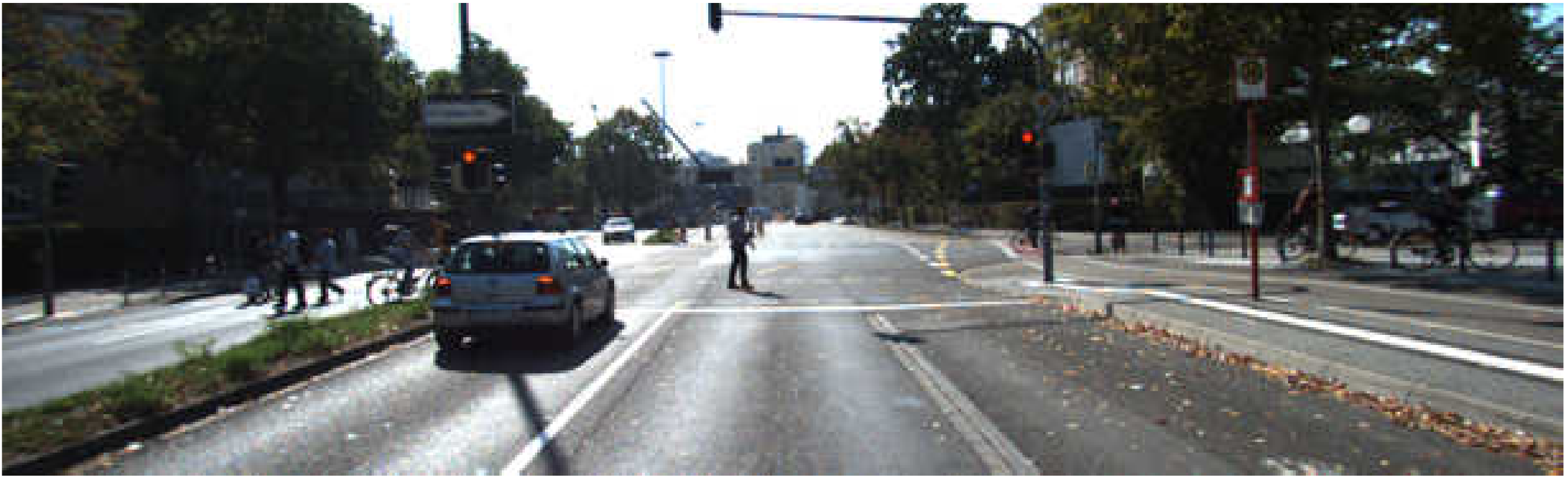}} \\[-0.7mm]
			\hspace{-0.7mm}\rotatebox{90}{\hspace{1mm}\footnotesize Top 100 prop.} &\hspace{-0.5cm}
			\raisebox{-0.0\height}{\includegraphics[width=0.315\linewidth,trim = 5mm 0mm 0mm 0mm, clip]{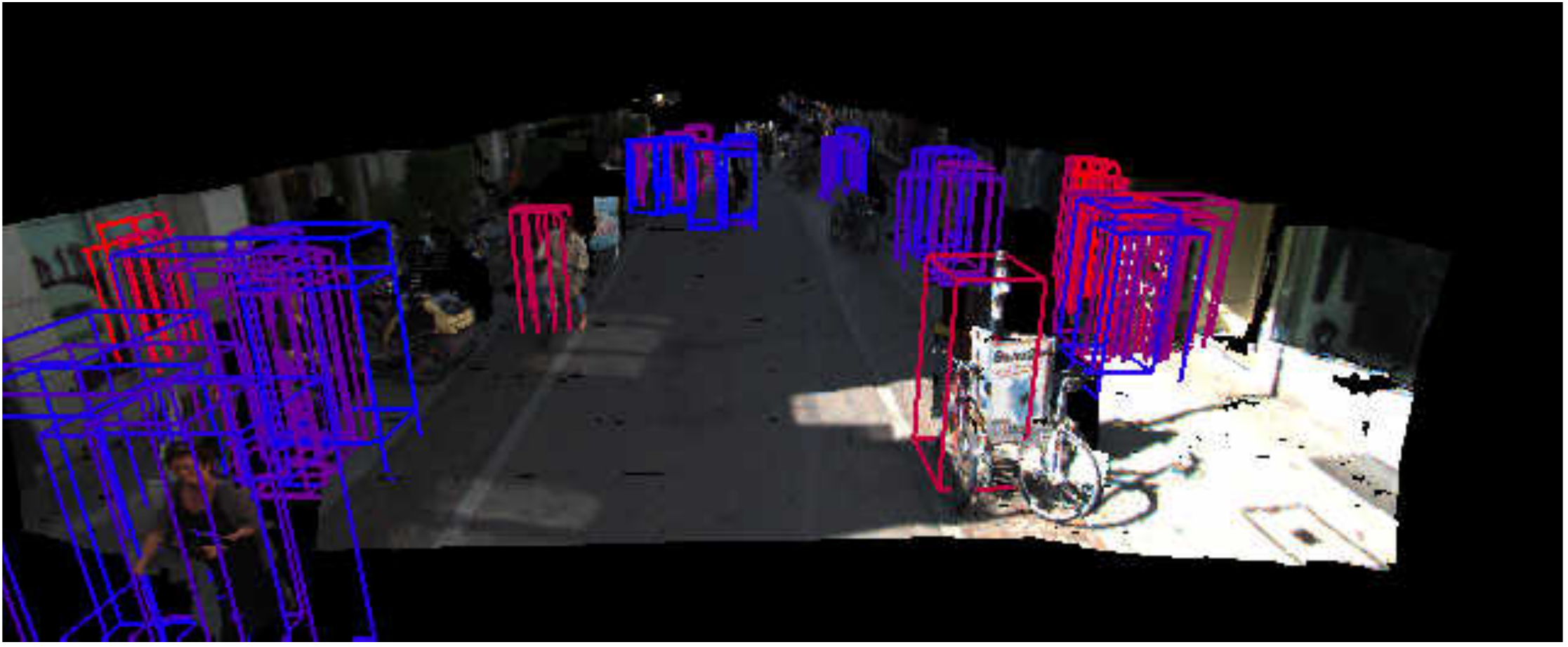}}&
			\raisebox{-0.0\height}{\includegraphics[width=0.315\linewidth,trim = 5mm 0mm 0mm 0mm, clip]{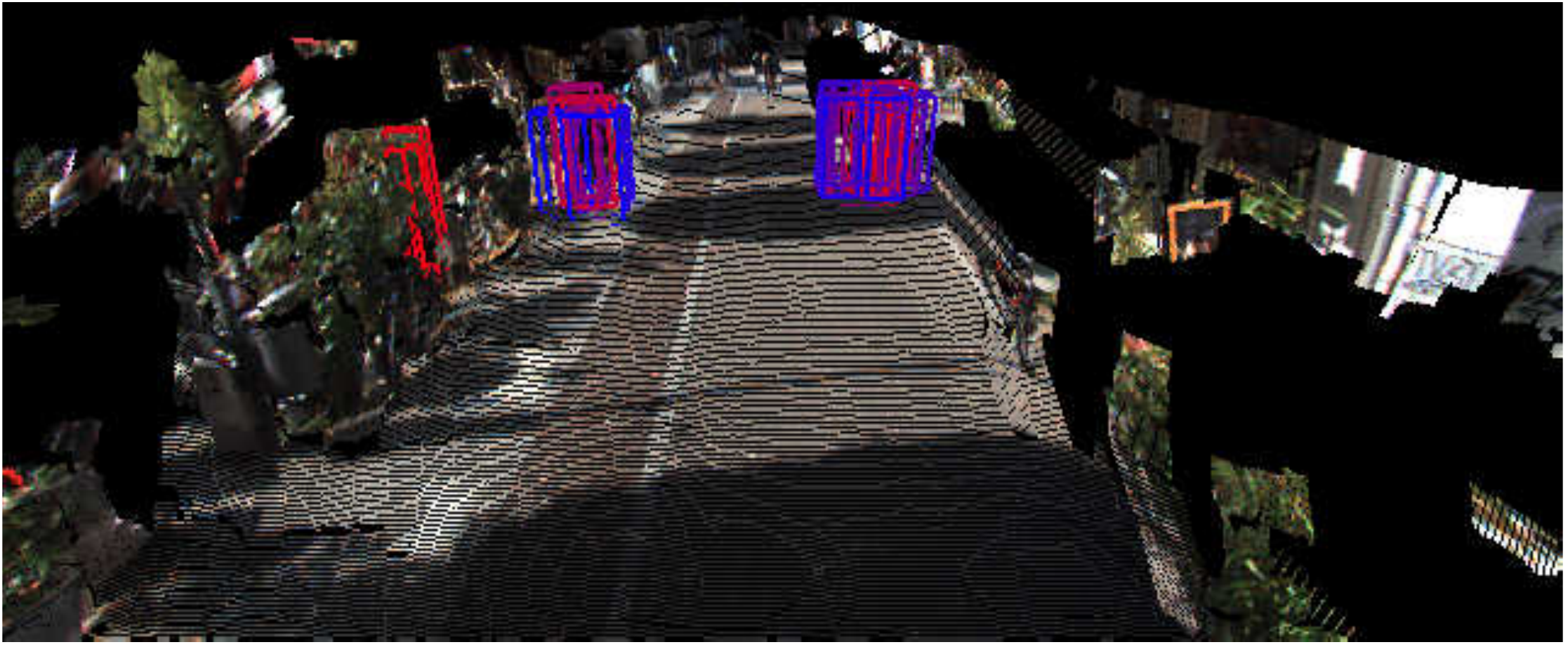}} &
			\raisebox{0.0\height}{\includegraphics[width=0.315\linewidth,trim =  5mm 0mm 0mm 0mm, clip]{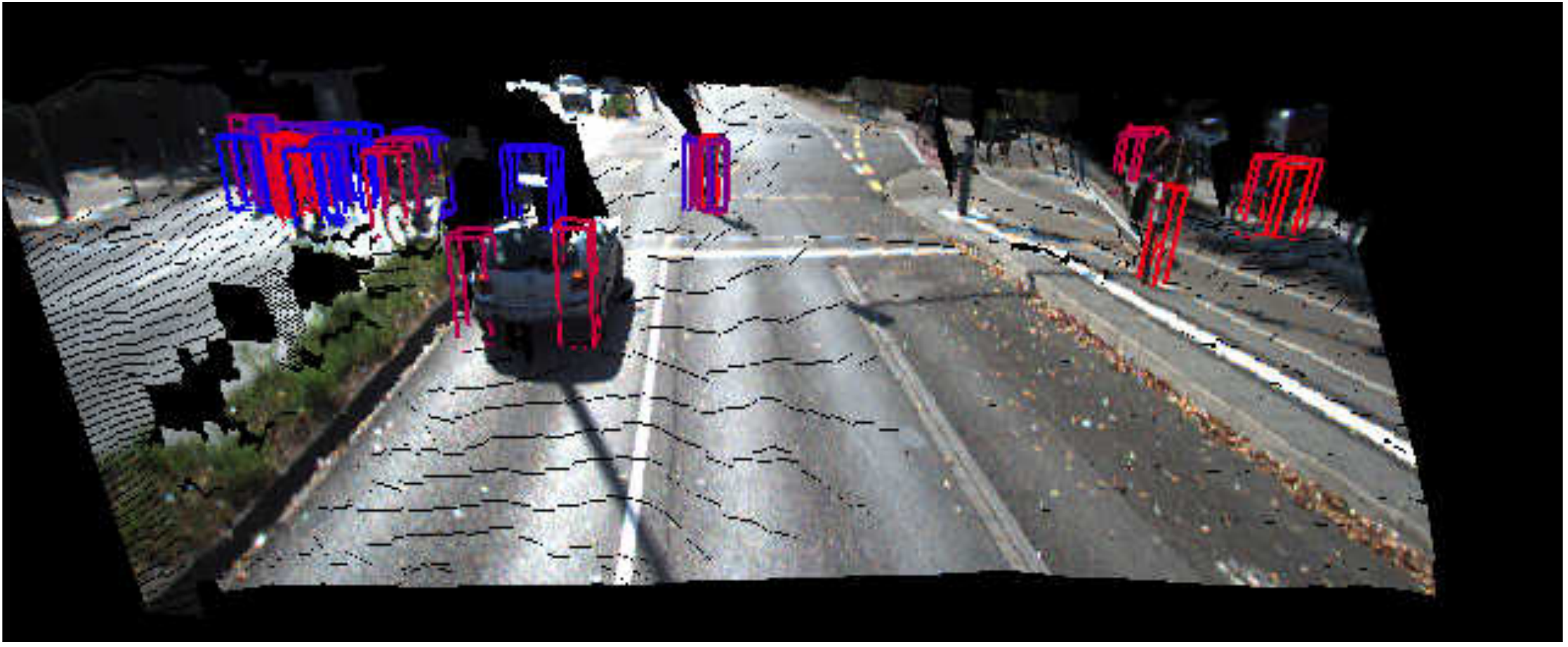}} \\[-0.7mm]
			\hspace{-0.7mm}\rotatebox{90}{\hspace{1.5mm}\footnotesize Ground truth} &\hspace{-0.5cm}
			\raisebox{-0.0\height}{\includegraphics[width=0.315\linewidth,trim = 5mm 0mm 0mm 0mm, clip]{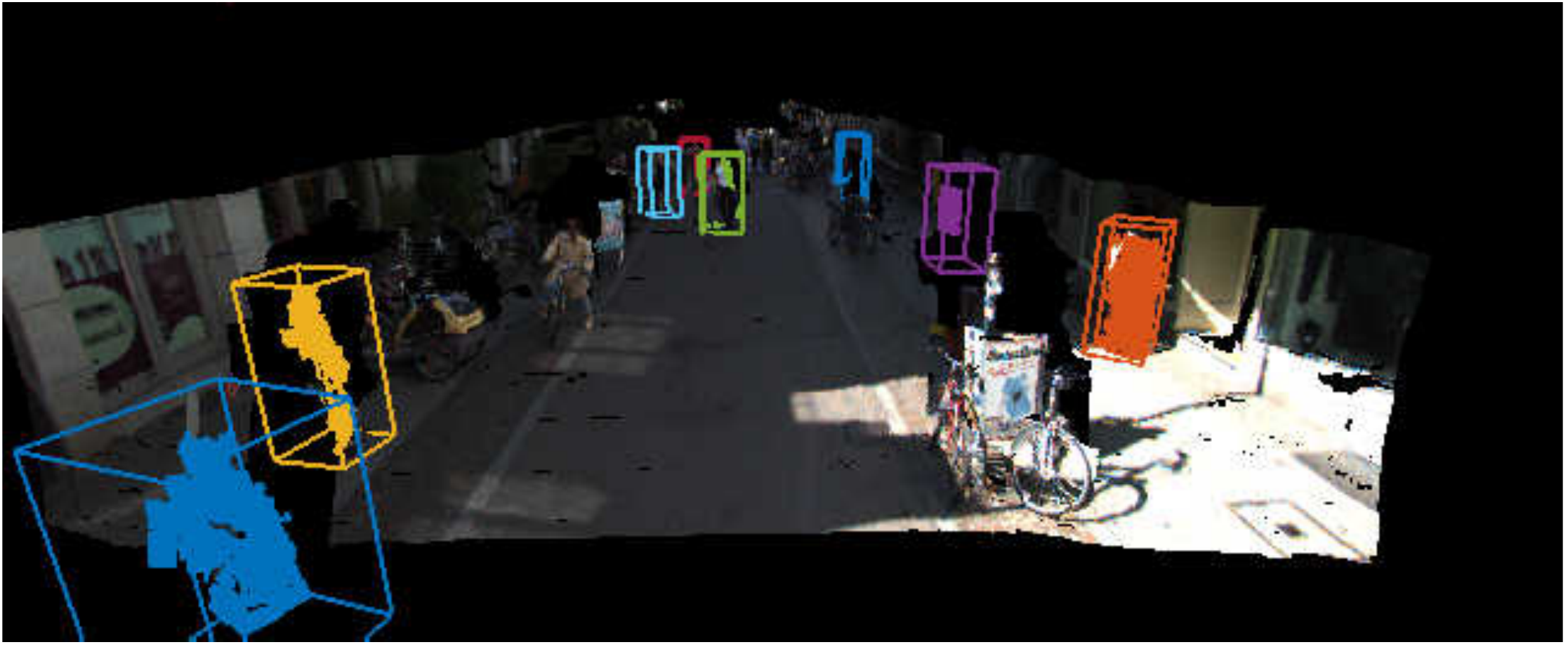}}&
			\raisebox{-0.0\height}{\includegraphics[width=0.315\linewidth,trim = 5mm 0mm 0mm 0mm, clip]{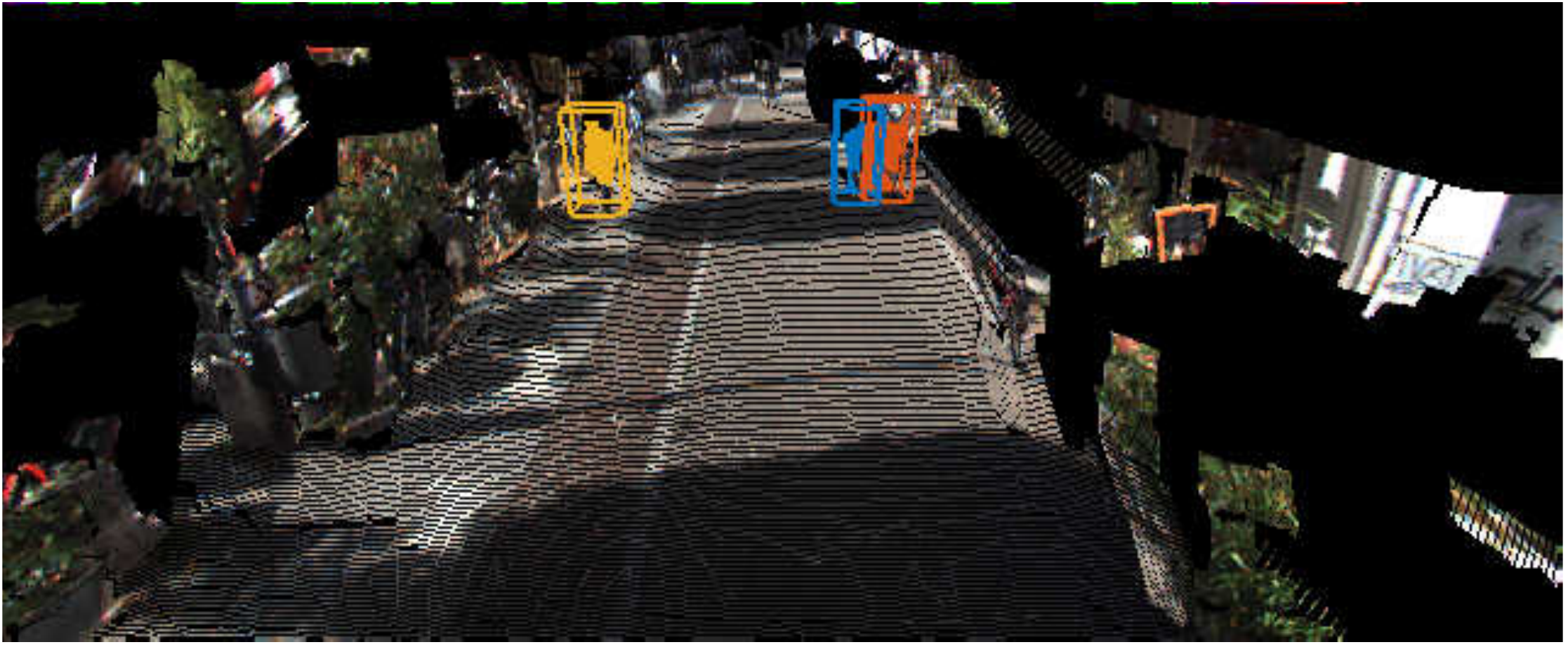}} &
			\raisebox{-0.0\height}{\includegraphics[width=0.315\linewidth,trim = 5mm 0mm 0mm 0mm, clip]{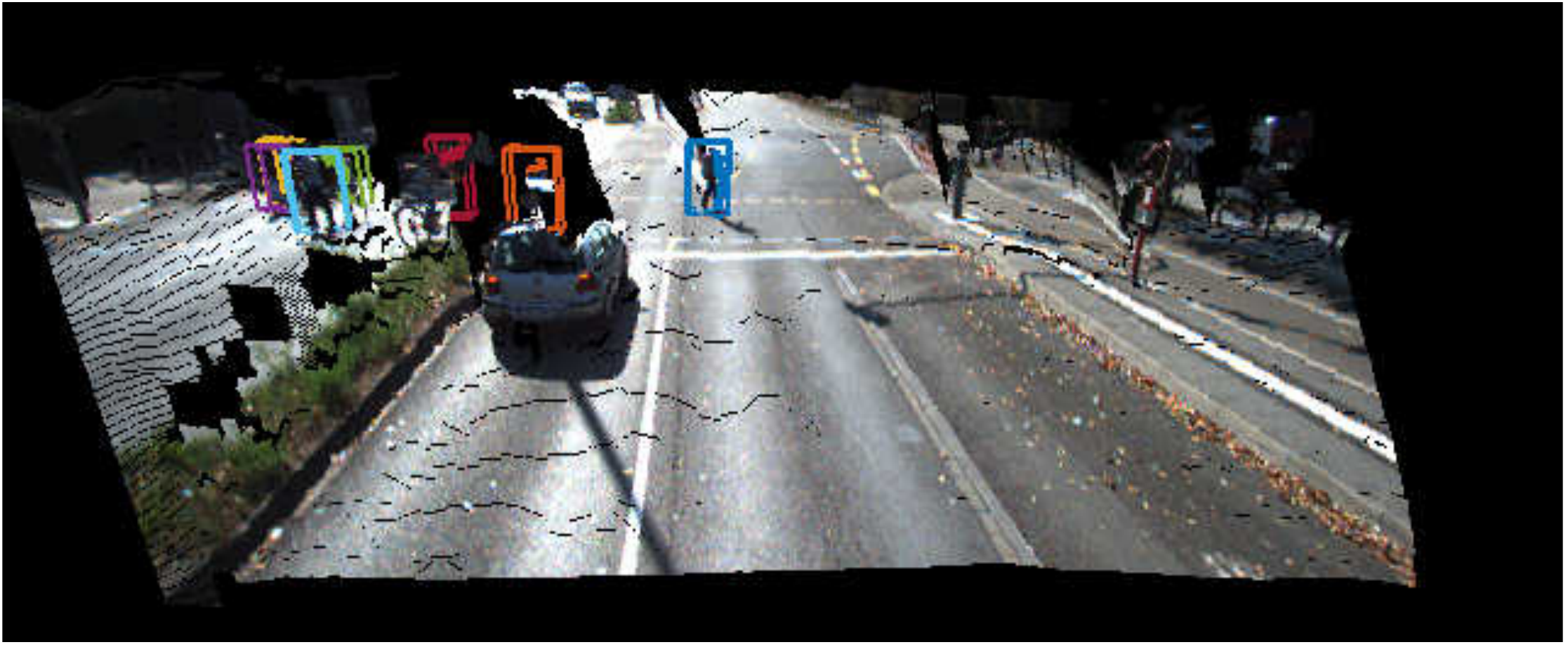}} \\[-0.7mm]
			\hspace{-0.7mm}\rotatebox{90}{\hspace{4mm}\footnotesize Best prop.} &\hspace{-0.5cm}
			\raisebox{-0.0\height}{\includegraphics[width=0.315\linewidth,trim = 5mm 0mm 0mm 0mm, clip]{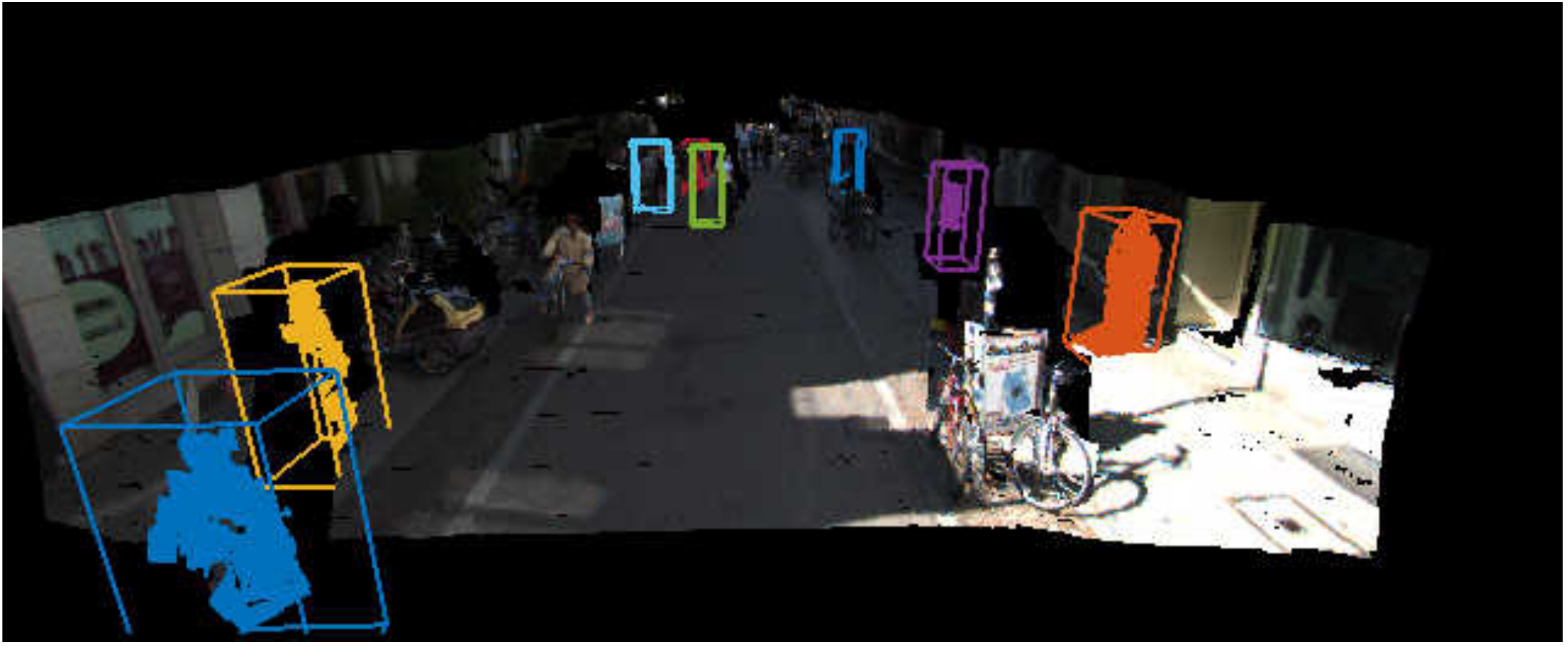}}&
			\raisebox{-0.0\height}{\includegraphics[width=0.315\linewidth,trim = 5mm 0mm 0mm 0mm, clip]{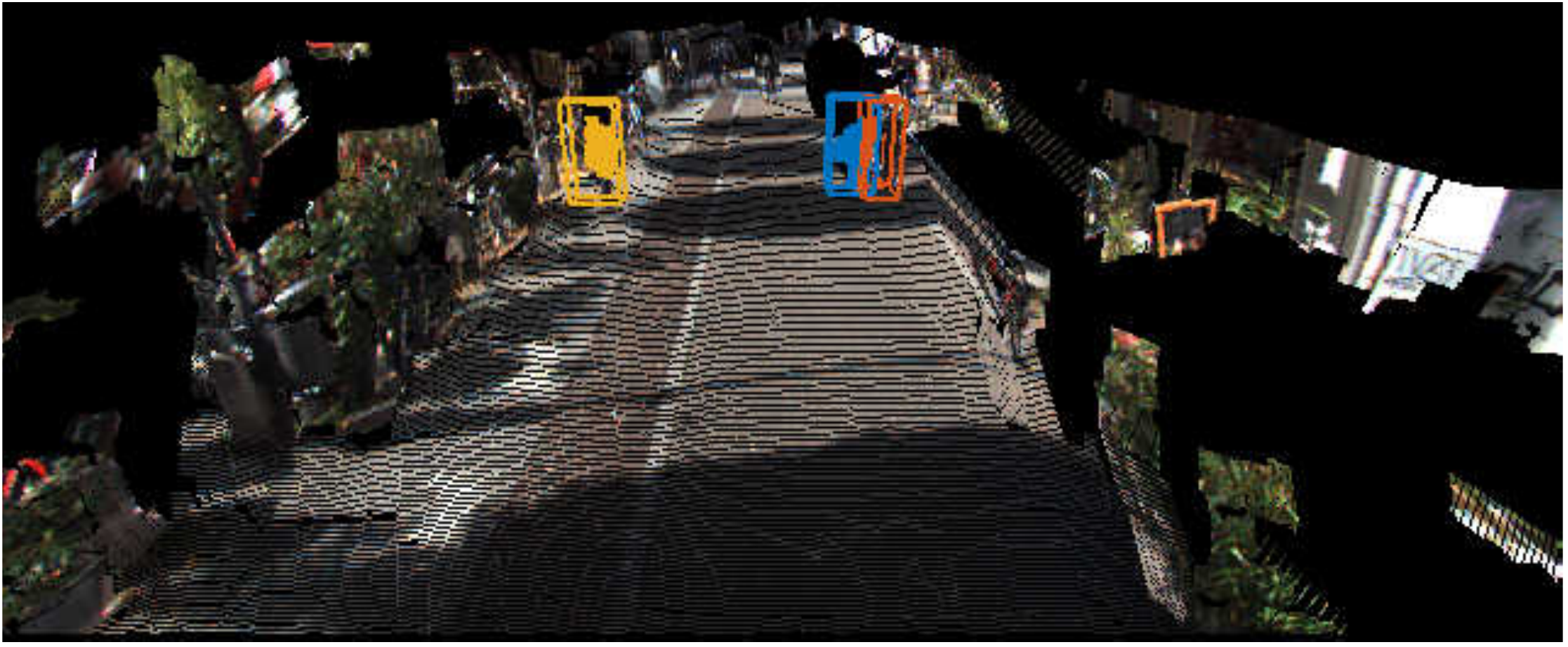}} &
			\raisebox{-0.0\height}{\includegraphics[width=0.315\linewidth,trim = 5mm 0mm 0mm 0mm, clip]{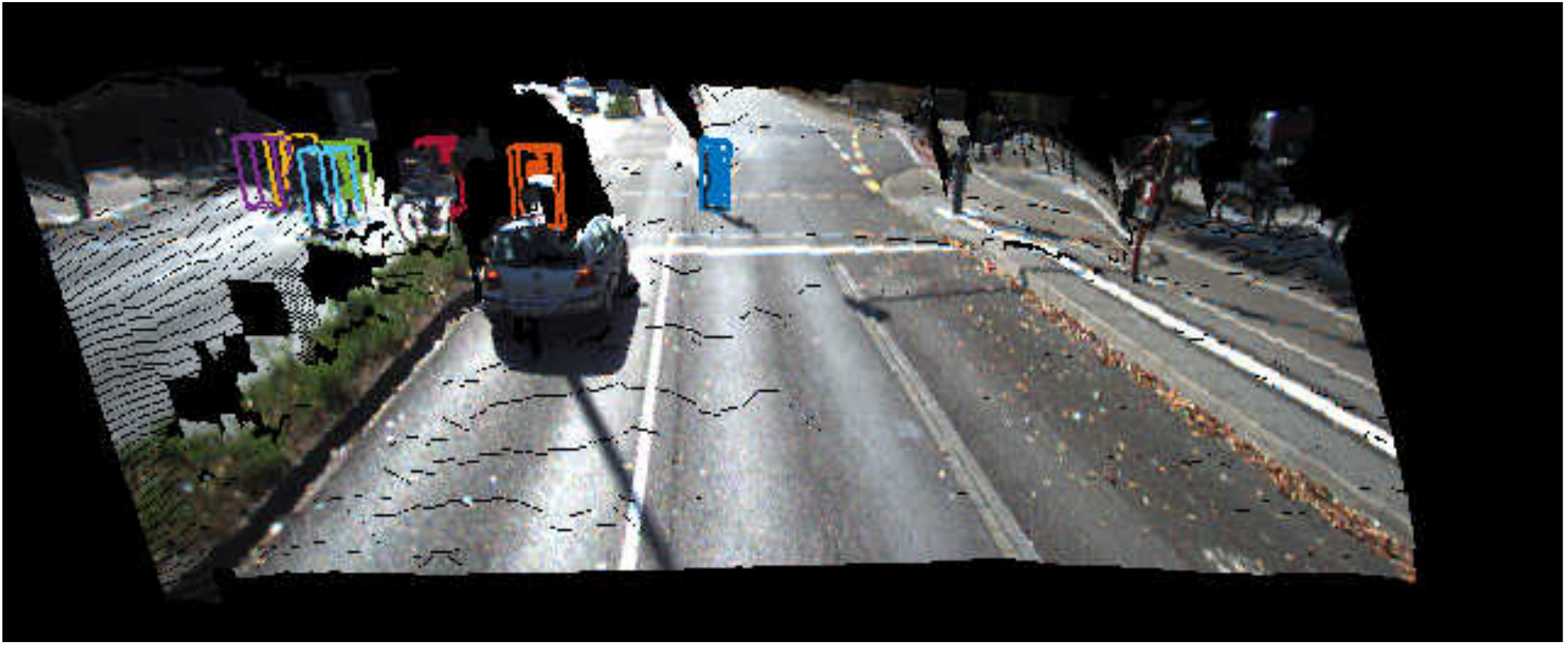}} \\[-0.7mm]
		\end{tabular}
		\vspace{-3mm}
		\caption{\small Qualitative examples for the {\it Pedestrian} class.}
		\label{fig:pcl-vis-ped}
	\end{center}
	\vspace{-5mm}
\end{figure*}

\section{Conclusion}
We  presented a novel approach to 3D object detection in the context of autonomous driving. In contrast to most existing work, we take advantage of stereo imagery and generate a set of 3D object proposals that are then run through a convolutional neural network to obtain high-quality 3D object detections. We generate 3D proposals by minimizing an energy function that encodes object size priors, ground plane context, and some depth informed features. For CNN scoring, we exploit appearance as well as depth and context information in our neural net and jointly predict 3D bounding box coordinates and object pose.

Our approach significantly outperforms existing state-of-the-art object proposal methods on the KITTI benchmark. Particularly, our approach achieves $25\%$ higher recall than the state-of-the-art RGB-D method MCG-D~\cite{guptaECCV14} for 2K proposals. We have evaluated our full 3D detection pipeline on the tasks of 2D object detection, joint 2D object detection and orientation estimation, as well as 3D object detection on KITTI. Our method significantly outperforms all previous published object detection methods for all three object classes on the challenging KITTI benchmark~\cite{kitti}.


\section*{Acknowledgments}
The work was partially supported by NSFC 61171113, NBRPC 2016YFB0100900, NSERC, ONR-N00014-14-1-0232 and Toyota Motor Corporation. We would like to thank NVIDIA for supporting our research by donating GPUs.

\bibliographystyle{IEEEtran}
\bibliography{IEEEabrv,egbib}

\begin{thebibliography}{10}
\providecommand{\url}[1]{#1}
\csname url@samestyle\endcsname
\providecommand{\newblock}{\relax}
\providecommand{\bibinfo}[2]{#2}
\providecommand{\BIBentrySTDinterwordspacing}{\spaceskip=0pt\relax}
\providecommand{\BIBentryALTinterwordstretchfactor}{4}
\providecommand{\BIBentryALTinterwordspacing}{\spaceskip=\fontdimen2\font plus
\BIBentryALTinterwordstretchfactor\fontdimen3\font minus
  \fontdimen4\font\relax}
\providecommand{\BIBforeignlanguage}[2]{{%
\expandafter\ifx\csname l@#1\endcsname\relax
\typeout{** WARNING: IEEEtran.bst: No hyphenation pattern has been}%
\typeout{** loaded for the language `#1'. Using the pattern for}%
\typeout{** the default language instead.}%
\else
\language=\csname l@#1\endcsname
\fi
#2}}
\providecommand{\BIBdecl}{\relax}
\BIBdecl

\bibitem{alexnet}
A.~Krizhevsky, I.~Sutskever, and G.~Hinton, ``Imagenet classification with deep
  convolutional neural networks,'' in \emph{NIPS}, 2012.

\bibitem{verydeep}
K.~Simonyan and A.~Zisserman, ``Very deep convolutional networks for
  large-scale image recognition,'' in \emph{arXiv:1409.1556}, 2014.

\bibitem{dpm}
P.~F. Felzenszwalb, R.~B. Girshick, D.~McAllester, and D.~Ramanan, ``Object
  detection with discriminatively trained part based models,'' \emph{PAMI},
  2010.

\bibitem{girshick2013rich}
R.~Girshick, J.~Donahue, T.~Darrell, and J.~Malik, ``Rich feature hierarchies
  for accurate object detection and semantic segmentation,'' \emph{CVPR}, 2014.

\bibitem{ZhuSegDeepM15}
Y.~Zhu, R.~Urtasun, R.~Salakhutdinov, and S.~Fidler, ``Seg{D}eep{M}: Exploiting
  segmentation and context in deep neural networks for object detection,'' in
  \emph{CVPR}, 2015.

\bibitem{pascal-voc-2010}
M.~Everingham, L.~Van~Gool, C.~K.~I. Williams, J.~Winn, and A.~Zisserman, ``The
  {PASCAL} {V}isual {O}bject {C}lasses {C}hallenge 2010 {(VOC2010)}
  {R}esults.''

\bibitem{van2011segmentation}
K.~Van~de Sande, J.~Uijlings, T.~Gevers, and A.~Smeulders, ``Segmentation as
  selective search for object recognition,'' in \emph{ICCV}, 2011.

\bibitem{ArbelaezCVPR14}
P.~Arbelaez, J.~Pont-Tusetand, J.~Barron, F.~Marques, and J.~Malik,
  ``Multiscale combinatorial grouping,'' in \emph{CVPR}, 2014.

\bibitem{AlexePAMI12}
B.~Alexe, T.~Deselares, and V.~Ferrari, ``Measuring the objectness of image
  windows,'' \emph{PAMI}, 2012.

\bibitem{BingObj2014}
M.~Cheng, Z.~Zhang, M.~Lin, and P.~Torr, ``{BING}: Binarized normed gradients
  for objectness estimation at 300fps,'' in \emph{CVPR}, 2014.

\bibitem{zitnick2014edge}
L.~Zitnick and P.~Doll{\'a}r, ``Edge boxes: Locating object proposals from
  edges,'' in \emph{ECCV}, 2014.

\bibitem{krahnbuhl15}
P.~Kr{\:a}henb{\:u}hl and V.~Koltun, ``Learning to propose objects,'' in
  \emph{CVPR}, 2015.

\bibitem{TLeeICCV15}
T.~Lee, S.~Fidler, and S.~Dickinson, ``A learning framework for generating
  region proposals with mid-level cues,'' in \emph{ICCV}, 2015.

\bibitem{deepproposal}
A.~Ghodrati, A.~Diba, M.~Pedersoli, T.~Tuytelaars, and L.~V. Gool,
  ``Deepproposal: Hunting objects by cascading deep convolutional layers,'' in
  \emph{ICCV}, 2015.

\bibitem{kitti}
A.~Geiger, P.~Lenz, and R.~Urtasun, ``Are we ready for autonomous driving? the
  kitti vision benchmark suite,'' in \emph{CVPR}, 2012.

\bibitem{girshick15fastrcnn}
R.~Girshick, ``Fast {R-CNN},'' in \emph{ICCV}, 2015.

\bibitem{ssvm}
T.~Joachims, T.~Finley, and C.-N.~J. Yu, ``Cutting-plane training of structural
  svms,'' \emph{JLMR}, 2009.

\bibitem{guptaECCV14}
S.~Gupta, R.~Girshick, P.~Arbelaez, and J.~Malik, ``Learning rich features from
  {RGB-D} images for object detection and segmentation,'' in \emph{ECCV}, 2014.

\bibitem{XiaozhiNIPS15}
X.~Chen, K.~Kundu, Y.~Zhu, A.~Berneshawi, H.~Ma, S.~Fidler, and R.~Urtasun,
  ``3d object proposals for accurate object class detection,'' in \emph{NIPS},
  2015.

\bibitem{CarreiraCpmcPAMI2012}
J.~Carreira and C.~Sminchisescu, ``Cpmc: Automatic object segmentation using
  constrained parametric min-cuts,'' \emph{PAMI}, vol.~34, no.~7, pp.
  1312--1328, 2012.

\bibitem{Banica13}
D.~Banica and C.~Sminchisescu, ``Cpmc-3d-o2p: Semantic segmentation of rgb-d
  images using cpmc and second order pooling,'' in \emph{CoRR abs/1312.7715},
  2013.

\bibitem{Lin13}
D.~Lin, S.~Fidler, and R.~Urtasun, ``Holistic scene understanding for 3d object
  detection with rgbd cameras,'' in \emph{ICCV}, 2013.

\bibitem{KarpathyICRA13}
A.~Karpathy, S.~Miller, and L.~Fei-Fei, ``Object discovery in 3d scenes via
  shape analysis,'' in \emph{ICRA}, 2013.

\bibitem{OneataECCV14}
D.~Oneata, J.~Revaud, J.~Verbeek, and C.~Schmid, ``Spatio-temporal object
  detection proposals,'' in \emph{ECCV}, 2014.

\bibitem{moving2015CVPR}
K.~Fragkiadaki, P.~Arbelaez, P.~Felsen, and J.~Malik, ``Learning to segment
  moving objects in videos,'' in \emph{CVPR}, 2015.

\bibitem{randPrim13}
S.~Manen, M.~Guillaumin, and L.~Van~Gool, ``Prime object proposals with
  randomized prim's algorithm,'' in \emph{ICCV}, 2013.

\bibitem{krahenbuhl14}
P.~Kr{\:a}henb{\:u}hl and V.~Koltun, ``Geodesic object proposals,'' in
  \emph{ECCV}, 2014.

\bibitem{s2p}
J.~Carreira, R.~Caseiro, J.~Batista, and C.~Sminchisescu, ``Semantic
  segmentation with second-order pooling,'' in \emph{ECCV}, 2012.

\bibitem{segdpm}
S.~Fidler, R.~Mottaghi, A.~Yuille, and R.~Urtasun, ``Bottom-up segmentation for
  top-down detection,'' in \emph{CVPR}, 2013.

\bibitem{bing++}
Z.~Zhang, Y.~Liu, T.~Bolukbasi, M.-M. Cheng, and V.~Saligrama, ``Bing++: A fast
  high quality object proposal generator at 100fps,'' \emph{arXiv:1511.04511},
  2015.

\bibitem{ChenCVPR15}
X.~Chen, H.~Ma, X.~Wang, and Z.~Zhao, ``Improving object proposals with
  multi-thresholding straddling expansion,'' in \emph{CVPR}, 2015.

\bibitem{Hosang2015PAMI}
J.~Hosang, R.~Benenson, P.~Doll\'ar, and B.~Schiele, ``What makes for effective
  detection proposals?'' \emph{PAMI}, 2015.

\bibitem{SongECCV14}
S.~Song and J.~Xiao, ``Sliding shapes for 3d object detection in depth
  images,'' in \emph{ECCV}, 2014.

\bibitem{PepikPAMI15}
B.~Pepik, M.~Stark, P.~Gehler, and B.~Schiele, ``Multi-view and 3d deformable
  part models,'' \emph{PAMI}, 2015.

\bibitem{ZiaIJCV15}
M.~Zia, M.~Stark, and K.~Schindler, ``Towards scene understanding with detailed
  3d object representations,'' \emph{IJCV}, 2015.

\bibitem{BarTITS15}
E.~Ohn-Bar and M.~M. Trivedi, ``Learning to detect vehicles by clustering
  appearance patterns,'' \emph{IEEE Transactions on Intelligent Transportation
  Systems}, 2015.

\bibitem{LongACCV14}
C.~Long, X.~Wang, G.~Hua, M.~Yang, and Y.~Lin, ``Accurate object detection with
  location relaxation and regionlets relocalization,'' in \emph{ACCV}, 2014.

\bibitem{WangCVPR15}
S.~Wang, S.~Fidler, and R.~Urtasun, ``Holistic 3d scene understanding from a
  single geo-tagged image,'' in \emph{CVPR}, 2015.

\bibitem{CarAOG_ECCV2014}
B.~Li, T.~Wu, and S.~Zhu, ``Integrating context and occlusion for car detection
  by hierarchical and-or model,'' in \emph{ECCV}, 2014.

\bibitem{spLBP}
Q.~Hu, S.~Paisitkriangkrai, C.~Shen, A.~van~den Hengel, and F.~Porikli, ``Fast
  detection of multiple objects in traffic scenes with a common detection
  framework,'' \emph{T-ITS}, 2015.

\bibitem{xiangcvpr15}
Y.~Xiang, W.~Choi, Y.~Lin, and S.~Savarese, ``Data-driven 3d voxel patterns for
  object category recognition,'' in \emph{CVPR}, 2015.

\bibitem{DollarPAMI14}
P.~Doll\'ar, R.~Appel, S.~Belongie, and P.~Perona, ``Fast feature pyramids for
  object detection,'' \emph{PAMI}, 2014.

\bibitem{HosangCVPR15}
J.~Hosang, M.~Omran, R.~Benenson, and B.~Schiele, ``Taking a deeper look at
  pedestrians,'' \emph{CVPR}, 2015.

\bibitem{DeepParts2015}
Y.~Tian, P.~Luo, X.~Wang, and X.~Tang, ``Deep learning strong parts for
  pedestrian detection,'' in \emph{ICCV}, 2015.

\bibitem{CompACT2015}
Z.~Cai, M.~Saberian, and N.~Vasconcelos, ``Learning complexity-aware cascades
  for deep pedestrian detection,'' in \emph{ICCV}, 2015.

\bibitem{renNIPS15fasterrcnn}
S.~Ren, K.~He, R.~Girshick, and J.~Sun, ``Faster {R-CNN}: Towards real-time
  object detection with region proposal networks,'' in \emph{NIPS}, 2015.

\bibitem{Koichiro14}
K.~Yamaguchi, D.~McAllester, and R.~Urtasun, ``Efficient joint segmentation,
  occlusion labeling, stereo and flow estimation,'' in \emph{ECCV}, 2014.

\bibitem{TsochantaridisICML2004}
I.~Tsochantaridis, T.~Hofmann, T.~Joachims, and Y.~Altun, ``{Support Vector
  Learning for Interdependent and Structured Output Spaces},'' in \emph{ICML},
  2004.

\bibitem{box}
A.~Schwing, S.~Fidler, M.~Pollefeys, and R.~Urtasun, ``Box in the box: Joint 3d
  layout and object reasoning from single images,'' in \emph{ICCV}, 2013.

\bibitem{chatfield2014return}
K.~Chatfield, K.~Simonyan, A.~Vedaldi, and A.~Zisserman, ``Return of the devil
  in the details: Delving deep into convolutional nets,'' in \emph{BMVC}, 2014.

\bibitem{behley2013iros}
J.~Behley, V.~Steinhage, and A.~B. Cremers, ``{Laser-based Segment
  Classification Using a Mixture of Bag-of-Words},'' in \emph{Proc. of the
  IEEE/RSJ Intl. Conf. on Intelligent Robots and Systems (IROS)}, 2013.

\bibitem{csor}
L.~Plotkin, ``Pydriver: Entwicklung eines frameworks für räumliche detektion
  und klassifikation von objekten in fahrzeugumgebung,'' Bachelor's Thesis,
  Karlsruhe Institute of Technology, 2015.

\bibitem{vote3d}
D.~Z. Wang and I.~Posner, ``Voting for voting in online point cloud object
  detection,'' in \emph{Proc. of Robotics: Science and Systems}, 2015.

\bibitem{velofcn}
B.~Li, T.~Zhang, and T.~Xia, ``Vehicle detection from 3d lidar using fully
  convolutional network,'' in \emph{Proc. of RSS}, 2016.

\bibitem{AGonzalez2015}
A.~Gonzalez, G.~Villalonga, J.~Xu, D.~Vazquez, J.~Amores, and A.~Lopez,
  ``Multiview random forest of local experts combining rgb and lidar data for
  pedestrian detection,'' in \emph{IV}, 2015.

\bibitem{adas2014}
J.~Xu, S.~Ramos, D.~Vozquez, and A.~Lopez, ``{Hierarchical Adaptive Structural
  SVM for Domain Adaptation},'' in \emph{arXiv:1408.5400}, 2014.

\bibitem{CPremebida_IROS2014}
C.~Premebida, J.~Carreira, J.~Batista, and U.~Nunes, ``Pedestrian detection
  combining rgb and dense lidar data,'' in \emph{IROS}, 2014.

\bibitem{Paul2014Pedestrian}
S.~Paisitkriangi, C.~Shen, and A.~van~den Hengel, ``Pedestrian detection with
  spatially pooled features and structured ensemble learning,'' in
  \emph{arXiv:1409.5209}, 2014.

\bibitem{Zhang2015FilteredICF}
S.~Zhang, R.~Benenson, and B.~Schiele, ``Filtered channel features for
  pedestrian detection,'' in \emph{CVPR}, 2015.

\bibitem{Geiger11}
A.~Geiger, C.~Wojek, and R.~Urtasun, ``Joint 3d estimation of objects and scene
  layout,'' in \emph{NIPS}, 2011.

\bibitem{bojan13cvpr}
B.~Pepik, M.~Stark, P.~Gehler, and B.~Schiele, ``Occlusion patterns for object
  class detection,'' in \emph{CVPR}, 2013.

\end{thebibliography}

\begin{IEEEbiography}[{\includegraphics[width=1in,height=1.25in,clip,keepaspectratio]{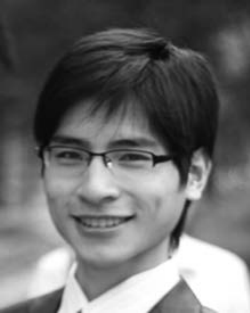}}]{Xiaozhi Chen}
received the B.S. degree in Electronic Engineering from Tsinghua University, Beijing, China in 2012, where he is currently pursuing  the Ph.D degree. His research interests include computer vision and machine learning.
\end{IEEEbiography}

\begin{IEEEbiography}[{\includegraphics[width=1in,height=1.25in,trim=45 120 35 0,clip=true,keepaspectratio]{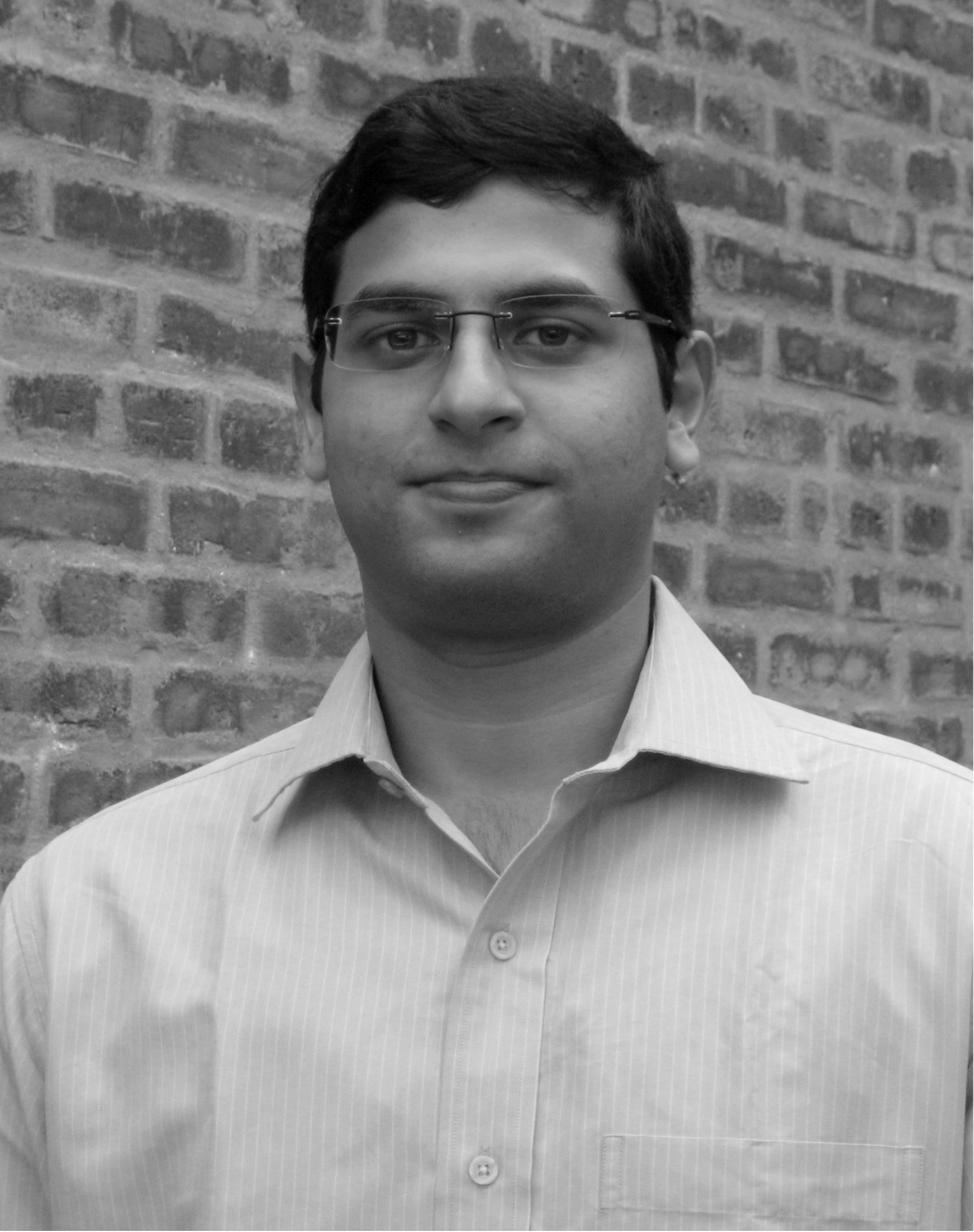}}]{Kaustav Kundu}
	received the B.Tech. (Hons.) degree in Computer Science and Engineering from IIIT Hyderabad, India in 2012. He is currently pursuing his Ph.D. degree in Computer Science from University of Toronto. His research interests include computer vision and machine learning.
\end{IEEEbiography}

\begin{IEEEbiography}[{\includegraphics[width=1in,height=1.25in,clip,keepaspectratio]{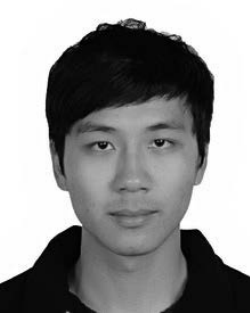}}]{Yukun Zhu}
received the BEng degree in electronic engineering from Shanghai Jiao Tong University in 2011, dual MSc degree in information \& communication engineering from Shanghai Jiao Tong University and electric \& computer engineering from Georgia Institute of Technology in 2014, the MSc degree in computer science from University of Toronto in 2016. He is currently working at Google. His research interests include computer vision and machine learning.
\end{IEEEbiography}

\begin{IEEEbiography}[{\includegraphics[width=1in,height=1.25in,clip,keepaspectratio]{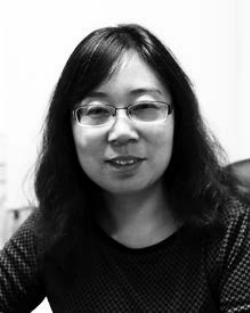}}]{Huimin Ma}
received the M.S. and Ph.D. degrees in Mechanical Electronic Engineering from Beijing Institute of Technology, Beijing, China in 1998 and 2001 respetively. She is an associate professor in the Department of Electronic Engineering of Tsinghua University, and the director of 3D Image Simulation Lab. She worked as an visiting scholar in University of Pittsburgh in 2011. She is also the executive director and the vice secretary general of China Society of Image and Graphics. Her research and teaching interests include 3D object recognition and tracking, system modeling and simulation, psychological base of visual cognition.
\end{IEEEbiography}

\begin{IEEEbiography}[{\includegraphics[width=1in,height=1.25in,trim=0 5 0 0,clip,keepaspectratio]{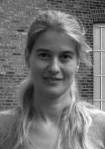}}]{Sanja Fidler}	
 is an Assistant Professor at the Department of Computer Science, University of Toronto.  Previously she was a Research Assistant Professor at TTI-Chicago. She completed her PhD in computer science at  University of Ljubljana in 2010, and was a postdoctoral fellow at University of Toronto in 2011-2012. In 2010 she visited UC Berkeley. She has served as a Program Chair of the 3DV conference, and as Area Chair of CVPR, ICCV, ACCV, EMNLP, ICLR, and NIPS. She received the NVIDIA Pioneer of AI award. Her main research interests lie in the intersection of language and vision, as well as 3D scene understanding.
\end{IEEEbiography}

\begin{IEEEbiography}[{\includegraphics[width=1in,height=1.25in,clip,keepaspectratio]{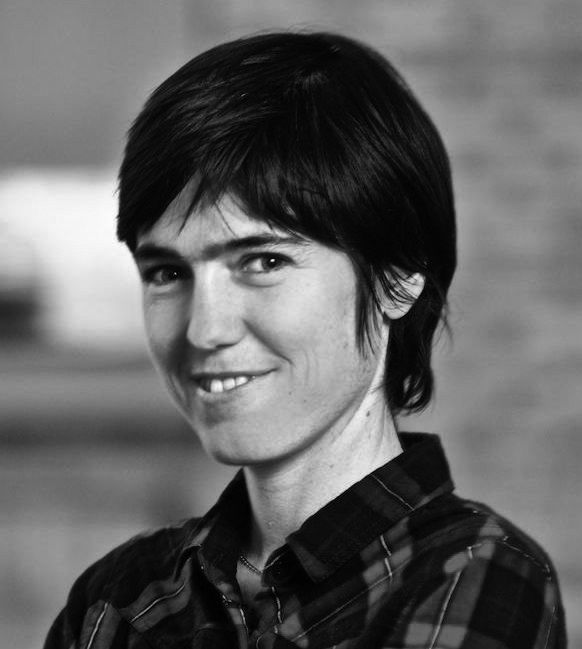}}]{Raquel Urtasun}	
Raquel Urtasun is an Associate Professor in the Department of Computer Science at the University of Toronto and a Canada Research Chair in Machine Learning and Computer Vision. Prior to this, she was an Assistant Professor at the Toyota Technological Institute at Chicago (TTIC), an academic computer science institute affiliated with the University of Chicago. She received her Ph.D. degree from the Computer Science department at Ecole Polytechnique Federal de Lausanne (EPFL) in 2006 and did her postdoc at MIT and UC Berkeley. Her research interests include machine learning, computer vision, robotics and remote sensing. Her lab was selected as an NVIDIA NVAIL lab. She is a recipient of an NSERC EWR Steacie Award, an NVIDIA Pioneers of AI Award, a Ministry of Education and Innovation Early Researcher Award, three Google Faculty Research Awards, an Amazon Faculty Research Award, a Connaught New Researcher Award and a Best Paper Runner up Prize awarded at the Conference on Computer Vision and Pattern Recognition (CVPR). She is also Program Chair of CVPR 2018, an Editor of the International Journal in Computer Vision (IJCV).
\end{IEEEbiography}
\end{document}